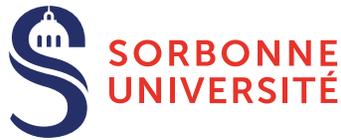

# Probabilistic Modeling for Novelty Detection with Applications to Fraud Identification

By
Rémi Domingues

A DOCTORAL DISSERTATION SUBMITTED IN PARTIAL SATISFACTION
OF THE REQUIREMENTS FOR THE DEGREE OF
DOCTOR OF PHILOSOPHY
IN THE
DOCTORAL SCHOOL N°130
COMPUTER SCIENCE, TELECOMMUNICATIONS AND ELECTRONICS OF PARIS
OF THE
SORBONNE UNIVERSITY

January 2019

COMMITTEE IN CHARGE:

| | | |
|---|---|---|
| Dr. Maurizio Filippone | EURECOM | Advisor |
| Dr. Pietro Michiardi | EURECOM | Co-advisor |
| Dr. Guido Sanguinetti | University of Edinburgh | Reviewer |
| Dr. Edwin Bonilla | CSIRO's Data61 | Reviewer |
| Dr. Marie-Jeanne Lesot | Sorbonne University | Examiner |
| Dr. Maria A. Zuluaga | Amadeus | Examiner |
| Dr. Marco Lorenzi | Inria | Examiner |

# Acknowledgments

First and foremost, I would like to thank my supervisors Maurizio Filippone and Pietro Michiardi for their advices and the time we spent discussing ideas, alternatives and reviewing my work. They always pushed me further and never ceased to suggest improvements and new research directions.

I wish to thank the Amadeus Middleware Fraud Detection team directed by Virginie Amar and Jérémie Barlet, led by the product owner Christophe Allexandre and composed of Jean-Blas Imbert, Jiang Wu, Yang Pu and Damien Fontanes for building the TRANSACTIONS, SHARED-ACCESS, PAYMENT, RIGHTS, TRANSACTIONS-FR and TRANSACTIONS-MO datasets.

I would like to express my gratitude to my colleagues and friends from the Fraud Detection and Security Administration teams of Amadeus, with whom I shared many great moments. The exchanges I had with the Innovation and Research team at Amadeus were invariably productive and buzzing. I am also grateful to my fellow Ph.D. candidates Kurt and Gia-Lac for their troubleshooting support.

A special thank goes to my dear friends Jean-Blas and Paul for the discussions we had, discovering cosmology and putting the world to rights. Last but not least, a special thank goes to my family and my partner Valiani, for their love and support during this thesis.

Thesis advisor: Dr. Maurizio Filippone             Rémi Domingues
Co-advisor: Dr. Pietro Michiardi

# Probabilistic Modeling for Novelty Detection with Applications to Fraud Identification

## Abstract

Novelty detection is the unsupervised problem of identifying anomalies in test data which significantly differ from the training set. Novelty detection is one of the classic challenges in Machine Learning and a core component of several research areas such as fraud detection, intrusion detection, medical diagnosis, data cleaning, and fault prevention. While numerous algorithms were designed to address this problem, most methods are only suitable to model continuous numerical data. Tackling datasets composed of mixed-type features, such as numerical and categorical data, or temporal datasets describing discrete event sequences is a challenging task. In addition to the supported data types, the key criteria for efficient novelty detection methods are the ability to accurately dissociate novelties from nominal samples, the interpretability, the scalability and the robustness to anomalies located in the training data.

In this thesis, we investigate novel ways to tackle these issues. In particular, we propose (i) an experimental comparison of novelty detection methods for mixed-type data (ii) an experimental comparison of novelty detection methods for sequence data, (iii) a probabilistic nonparametric novelty detection method for mixed-type data based on Dirichlet process mixtures and exponential-family distributions and (iv) an autoencoder-based novelty detection model with encoder/decoder modelled as deep Gaussian processes.

We first propose to model mixed-type features with a nonparametric mixture of exponential-family distributions. The resulting algorithm is a Dirichlet Process Mixture Model. We compare this method against a wide set of state-of-the-art machine learning algorithms in the context of unsupervised anomaly detection. The selected methods are benchmarked on publicly available datasets and novel industrial datasets from the company Amadeus. We further run extensive scalability, memory consumption and robustness tests in order to build a full overview of the algorithms' characteristics.

Thesis advisor: Dr. Maurizio Filippone                    Rémi Domingues
Co-advisor: Dr. Pietro Michiardi

We further propose a novel autoencoder based on Deep Gaussian Processes for novelty detection tasks. The learning of the proposed model is made tractable and scalable through the use of random feature approximations and stochastic variational inference. The result is a flexible model that is easy to implement and train, and can be applied to general novelty detection tasks, including large-scale problems and data with mixed-type features. The experiments indicate that the proposed model achieves competitive results with state-of-the-art novelty detection methods.

In order to tackle discrete temporal data which arise in network intrusion, genomics and user behavior analysis, we provide an experimental comparison of novelty detection methods in the context of anomalous sequence identification. The objective of this study is to identify efficient and appropriate state-of-the-art methods for specific use cases. These recommendations rely on extensive experiments based on a variety of public datasets and novel industrial datasets. We also perform thorough scalability and memory usage tests resulting in new supplementary insights of the methods' performance, key selection criterion to solve problems relying on large volumes of data and to meet the expectations of applications subject to strict response time constraints.

# Contents

# Listing of figures

# Listing of tables

# 1

# Introduction

Novelty detection is a fundamental task across numerous domains, with applications in data cleaning [Liu et al., 2004], fault detection and damage control [Dereszynski & Dietterich, 2011, Worden et al., 2000], fraud detection related to credit cards [Hodge & Austin, 2004] and network security [Pokrajac et al., 2007], along with several medical applications such as brain tumor [Prastawa et al., 2004] and breast cancer [Greensmith et al., 2006] detection. Novelty detection targets the recognition of test samples which differ significantly from the training set [Pimentel et al., 2014]. This problem is also known as "unsupervised anomaly detection". Challenges in performing novelty detection stem from the fact that labelled data identifying anomalies in the training set is usually scarce and expensive to obtain, and that very little is usually known about the distribution of such novelties. Meanwhile, the training set itself might be corrupted by outliers and this might impact the ability of novelty detection methods to accurately characterize the distribution of samples associated with a nominal behavior of the system under study. Furthermore, there are many applications, such as the ones that we study in this work, where the volume and heterogeneity of data might pose serious computational challenges to react to novelties in a timely manner and to develop flexible novelty detection algorithms. As an example, the Airline IT company Amadeus provides booking platforms handling millions of transactions per second, resulting in more than 3 million bookings per day and Petabytes of stored data. This company manages almost half of the flight bookings worldwide and is targeted by fraud attempts leading to revenue losses and indemnifications.

Detecting novelties in such large volumes of data is a daunting task for a human operator; thus, an automated and scalable approach is truly desirable. Because of the difficulty in obtaining labelled data and since the scarcity of anomalies is challenging for supervised methods [Japkowicz & Stephen, 2002], anomaly detection is often approached as an unsupervised machine learning problem [Pimentel et al., 2014], called novelty detection. Novelty detection has also been described as a semi-supervised problem [Chandola et al., 2012] when the training set is exempt of outliers. Novelty and outlier detection are two very similar tasks, and these terms are often considered interchangeable. Nonetheless outlier detection methods are by definition trained on datasets contaminated by outliers, and this term is more prevalent in the data cleaning community.

Consider an unsupervised learning problem where we are given a set of input vectors $X = [\mathrm{x}_1, \ldots, \mathrm{x}_n]^\top$. Novelty detection is the task of classifying new test points $\mathrm{x}_*$, based on the criterion that they significantly differ from the input vectors $X$, that is the data available at training time. Such data is assumed to be generated by a different generative process and called *anomalies*. Novelty detection is thus a one-class classification problem, which aims at constructing a model describing the distribution of nominal samples in a dataset. Unsupervised learning methods allow for the prediction on test data $\mathrm{x}_*$; given a model with parameters $\boldsymbol{\theta}$, predictions are defined as $h(\mathrm{x}_*|X, \boldsymbol{\theta})$. Assuming $h(\mathrm{x}_*|X, \boldsymbol{\theta})$ to be continuous, it is possible to interpret it as a means of scoring test points as novelties. The resulting scores allow for a ranking of test points $\mathrm{x}_*$ highlighting the patterns which differ the most from the training data $X$. In particular, it is possible to define a threshold $\alpha$ and flag a test point $\mathrm{x}_*$ as a novelty when $h(\mathrm{x}_*|X, \boldsymbol{\theta}) > \alpha$.

After thresholding, it is possible to assess the quality of a novelty detection algorithm using scores proposed in the literature for binary classification. Based on a labelled testing dataset, where novelties and nominal cases are defined as *positive* and *negative* samples, respectively, we can compute the *precision* and *recall* metrics given in equation 1.1. True positives (TP) are examples correctly labelled as positives, false positives (FP) refer to negative samples incorrectly labelled as positives, while false negatives (FN) are positive samples incorrectly labelled as negatives.

$$precision = \frac{TP}{TP + FP} \qquad recall = \frac{TP}{TP + FN} \qquad (1.1)$$

In the remainder of this thesis we are going to assess results of novelty detection methods by varying $\alpha$ over the range of values taken by $h(\mathrm{x}_*|X, \boldsymbol{\theta})$ over a set of test points. When we vary $\alpha$, we obtain a set of *precision* and *recall* measurements resulting in a curve. We can then compute the area under the precision-recall curve called the *average precision* (AP), which is the recommended metric to compare the performance of novelty detection methods [Davis & Goadrich, 2006]. In practical terms, $\alpha$ is chosen to strike an appropriate balance between accuracy in identifying novelties and a low level of false positives.

Novelty detection has been thoroughly investigated by theoretical studies [Pimentel et al., 2014, Hodge & Austin, 2004]. The evaluation of state-of-the-art methods was also reported in experimental papers [Emmott et al., 2016], including experiments on the resistance to the curse of dimensionality [Zimek et al., 2012]. In one of the most recent surveys on novelty detection [Pimentel et al., 2014], methods have been classified into the following categories. (i) Probabilistic approaches estimate the probability density function of $X$ defined by the model parameters $\boldsymbol{\theta}$. Novelties are scored by the likelihood function $P(\mathbf{x}_*|\boldsymbol{\theta})$, which computes the probability for a test point to be generated by the trained distribution. These approaches are generative, and provide a simple understanding of the underlying data through parameterized distributions. (ii) Distance-based methods compute the pairwise distance between samples using various similarity metrics. Patterns with a small number of neighbors within a specified radius, or distant from the center of dense clusters of points, receive a high novelty score. (iii) Domain-based methods learn the domain of the nominal class as a decision boundary. The label assigned to test points is then based on their location with respect to the boundary. (iv) Information theoretic approaches measure the increase of entropy induced by including a test point in the nominal class. As an alternative, (v) isolation methods target the isolation of outliers from the remaining samples. As such, these techniques focus on isolating anomalies instead of profiling nominal patterns. (vi) Most unsupervised neural networks suitable for novelty detection are autoencoders, i.e. networks learning a compressed representation of the training data by minimizing the error between the input data and the reconstructed output. Test points showing a high reconstruction error are labelled as novelties.

While most anomaly detection tasks target numerical datasets [Emmott et al., 2016, Breunig et al., 2000, Ramaswamy et al., 2000], novelty detection methods have been successfully applied to categorical data [Hodge & Austin, 2004], time-series [Marchi et al., 2015, Kundzewicz & Robson, 2004, Taylor & Letham, 2018], discrete sequences [Chandola et al., 2008, Warrender et al., 1999, Cohen, 1995] and mixed-type data [Domingues et al., 2018a, Domingues et al., 2018b]. The remainder of this thesis is organized as follows. Chapter 2 is dedicated to the description of state-of-the-art novelty detection methods suitable for numerical, one-hot encoded and temporal data. Chapter 3 describes a probabilistic algorithm named Dirichlet Process Mixture

Model (DPMM) which we train through variational inference. This method supports mixed-type features through a mixture of exponential-family distributions. We further perform an experimental evaluation of state-of-the-art novelty detection algorithms, including our DPMM, and compare the novelty detection abilities, scalability, robustness and sensitivity to the curse of dimensionality of the selected methods. This experimental work was published in [Domingues et al., 2018a]. A Deep Gaussian Process autoencoder is described in Chapter 4 and presented in [Domingues et al., 2018b], where we propose a nonparametric and probabilistic approach to alleviate issues related to the choice of a suitable architecture for this neural network while accounting for the uncertainty in the autoencoder mappings; crucially, we show that this can be achieved while learning the model at scale. Chapter 5 expands [Domingues et al., 2019] and extends the comparison of novelty detection algorithms to sequence-based methods, studying the methods' performance on a wide range of datasets belonging to several research areas, while providing insights on the scalability and interpretability of the selected candidates.

## Publications

### International Journal Articles

- Domingues, R., Filippone, M., Michiardi, P., & Zouaoui, J. (2018a). A comparative evaluation of outlier detection algorithms: Experiments and analyses. *Pattern Recognition*, 74, 406 -- 421

- Domingues, R., Michiardi, P., Barlet, J., & Filippone, M. (2019). A comparative evaluation of novelty detection algorithms for discrete sequences. *arXiv preprint arXiv:1902.10940* Submitted to the *Artificial Intelligence Review* journal, 2019.

### International Journal and Conference Articles

- Domingues, R., Michiardi, P., Zouaoui, J., & Filippone, M. (2018b). Deep gaussian process autoencoders for novelty detection. *Machine Learning*, 107(8), 1363--1383 Presented at *ECML-PKDD*, 2018.

### International Workshop Articles

- Domingues, R., Buonora, F., Senesi, R., & Thonnard, O. (2016). An application of unsupervised fraud detection to passenger name records. In *2016 46th Annual IEEE/IFIP International Conference on Dependable Systems and Networks Workshop (DSN-W)* (pp. 54--59)

*Through every rift of discovery some seeming anomaly drops out of the darkness, and falls, as a golden link into the great chain of order.*

Edwin Hubbell Chapin

# 2

# State-of-the-art of novelty detection methods

In this chapter, we review popular unsupervised anomaly detection methods in addition to recent developments in this field. Section 2.1 is devoted to algorithms suitable for numerical data and one-hot encoded categorical variables. We survey both probabilistic, neighbor-based, domain-based and isolation methods, in addition to state-of-the-art neural networks. Section 2.2 focuses on novelty detection methods for sequences of events, and introduces several distance metrics suitable for comparing ordered sets of events.

## 2.1 Numerical and categorical based methods

The algorithms described in this section belong to a wide range of approaches. These methods build a model representing the nominal classes, i.e. dense clusters of similar data points, during a training phase. Online or batch predictions can thereafter be applied to new data based on the trained model to assign an anomaly score to the new observations. Applying a threshold on the returned scores provides a decision boundary separating nominal samples from outliers.

We describe both parametric and nonparametric machine learning algorithms. While parametric approaches model the underlying data with a fixed number of parameters, the number of parameters of nonparametric methods is potentially infinite and can increase with the com-plexity of data. If the former are often computationally faster, they require assumptions about the data distribution, e.g. the number of clusters, and may result in a flawed model if based on erroneous assumptions. The latter make fewer assumptions about the data distribution and may thus generalize better while requiring less knowledge about the data.

### 2.1.1 Probabilistic methods

Probabilistic algorithms estimate the probability density function of a dataset $X$, by inferring the model parameters $\theta$. Data points having the smallest likelihood $P(X|\theta)$ are identified as outliers. Most probabilistic methods described in this section can be trained incrementally, i.e. an existing model can be used as prior distribution when training the model on new input data in order to consider the distribution of previous and current samples in the final model.

A popular probabilistic algorithm is the **Gaussian Mixture Model (gmm)**, which fits a given number of Gaussian distributions to a dataset. The model is trained using the Expectation-Maximization (EM) algorithm [Dempster et al., 1977] which maximizes a lower bound of the likelihood iteratively. This method has been successfully applied to identify suspicious and possibly cancerous masses in mammograms by novelty detection in [Tarassenko et al., 1995]. However, assessing the number of components of the mixture by data exploration can be complex and motivates the use of nonparametric alternatives described hereafter.

[Blei & Jordan, 2006] describe the **Dirichlet Process Gaussian Mixture Model (dpgmm)**, a nonparametric Bayesian algorithm which optimizes the model parameters and tests for convergence by monitoring a non-decreasing lower bound on the log-marginal likelihood. The result is a mixture model where each component is a product of exponential-family distributions. We detail the full derivation of this model in Chapter 3 and extend it to support mixed-type features and to include prior knowledge on the mixing proportions. In the resulting model, the number of components is inferred during the training and requires an upper bound $K$. Weights $\pi_i$ are represented by a Dirichlet Process modelled as a truncated stick-breaking process (equation 2.1). The variable $v_i$ follows a Beta distribution, where $\alpha_k$ and $\beta_k$ are variational parameters optimized during the training for each component.

$$\pi_i(\boldsymbol{v}) = v_i \prod_{j=1}^{i-1}(1-v_j) \qquad q_{\boldsymbol{\alpha},\boldsymbol{\beta}}(\boldsymbol{v}) = \prod_{k=1}^{K-1} \mathrm{Beta}(\alpha_k, \beta_k) \tag{2.1}$$

The training optimizes the parameters of the posterior, e.g. a Gaussian-Wishart posterior when using a multivariate Gaussian likelihood, through variational inference (Section 3.1).

The scoring is then made by averaging the log likelihood computed from each mixture of likelihoods sampled from the conjugate priors. A primitive version of this algorithm is applied to intrusion detection on the KDD 99 dataset in [Fan et al., 2011] and outperforms SVM and KNN algorithms. The cluster centroids provided by the model can also be valuable to an end-user as they represent the average nominal data points.

**Kernel density estimators (KDE)**, also called *Parzen windows* estimators, approximate the density function of a dataset by assigning a kernel function to each training sample, then sums the local contribution of each function to give an estimate of the density. A bandwidth parameter acts as a smoothing parameter on the density shape and can be estimated by methods such as Least-Squares Cross-Validation (LSCV). As shown in [Tarassenko et al., 1995], KDE methods are efficient when applied to novelty detection problems. However, these approaches are sensitive to outliers and struggle in finding a good estimate of the nominal data density in datasets contaminated by outliers. This issue is shown by Kim et al. in [Kim & Scott, 2012] where the authors describe a **Robust Kernel Density Estimator (RKDE)**, algorithm which uses M-estimation methods, such as the Huber loss functions, to provide a robust estimation of the maximum likelihood.

**Probabilistic principal component analysis (PPCA)** [Tipping & Bishop, 1999] is a latent variable model which estimates the principal components of the data. It allows for the projection of a $d$-dimensional observation vector $Y$ to a $k$-dimensional vector of latent variables $X$, with $k$ the number of components of the model. The relationship $\boldsymbol{Y} = \boldsymbol{WX} + \boldsymbol{\mu} + \boldsymbol{\epsilon}$ is trained by expectation-maximization and uses a Gaussian prior. The authors suggest to use the log-likelihood as a degree of novelty for new data points.

More recently, **Least-squares anomaly detection (LSA)** [Quinn & Sugiyama, 2014] developed by Quinn et al. extends the multi-class least-squares probabilistic classifier (LSPC) [Sugiyama, 2010] to a one-class problem. The approach is compared against KNN and One-class SVM using the area under the ROC curve.

### 2.1.2 Distance-based methods

This class of methods uses solely the distance space to flag outliers. As such, the **Mahalanobis distance** is suitable for anomaly detection tasks targeting multivariate datasets composed of a single Gaussian-shaped cluster [Ben-Gal, 2005]. The model parameters are the mean and inverse covariance matrix of the data, thus similar to a one-component GMM with a full covariance matrix.

### 2.1.3 Neighbor-based methods

Neighbor-based methods study the neighborhood of each data point to identify outliers. **Local outlier factor (LOF)** described in [Breunig et al., 2000] is a well-known distance based approach corresponding to this description. For a given data point $x$, LOF computes its *degree* $d_k(x)$ of being an outlier based on the Euclidean distance $d$ between $x$ and its $k^{th}$ closest neighbor $n_k$, which gives $d_k(x) = d(x, n_k)$. The scoring of $x$ also takes into account for each of its neighbors $n_i$, the maximum between $d_k(n_i)$ and $d(x, n_i)$. As shown in [Emmott et al., 2016], LOF outperforms Angle-Based Outlier Detection [Kriegel et al., 2008] and One-class SVM [Schölkopf et al., 1999] when applied on real-world datasets for outlier detection.

**Angle-Based Outlier Detection (ABOD)** [Kriegel et al., 2008] uses the radius and variance of angles measured at each input vector instead of distances to identify outliers. The motivation is here to remain efficient in high-dimensional space and to be less sensible to the curse of dimensionality. Given an input point $x$, ABOD samples several pairs of points and computes the corresponding angles at $x$ and their variance. Broad angles imply that $x$ is located inside a major cluster as it is surrounded by many data points, while small angles denote that $x$ is positioned far from most points in the dataset. Similarly, a higher variance will be observed for points inside or at the border of a cluster than for outliers. The authors show that their method outperforms LOF on synthetic datasets containing more than 50 features. According to the authors, the pairs of vectors can be built from the entire dataset, a random subset or the $k$-nearest neighbors in order to speed up the computation at the cost of lower outlier detection performance.

The **Subspace outlier detection (SOD)** [Kriegel et al., 2009] algorithm finds for each point $p$ the set of $m$ neighbors shared between $p$ and its $k$-nearest neighbors. The outlier score is then the standard deviation of $p$ from the mean of a given subspace, which is composed of a subset of dimensions. The attributes having a small variance for the set of $m$ points are selected to be part of the subspace.

### 2.1.4 Information theory

The **Kullback-Leibler (KL) divergence** was used as an information-theoretic measure for novelty detection in [Filippone & Sanguinetti, 2010]. The method first trains a Gaussian mixture model on a training set, then estimates the *information content* of new data points by measuring the KL divergence between the estimated density and the density estimated on the training set and the new point. This reduces to an $F$-test in the case of a single Gaussian.

### 2.1.5 Domain-based methods

Additional methods for outlying data identification rely on the construction of a boundary separating the nominal data from the rest of the input space, thus estimating the domain of the nominal class. Any data point falling outside of the delimited boundary is thus flagged as outlier.

**One-class SVM** [Schölkopf et al., 1999], an application of support vector machine (svm) algorithms to one-class problems, belongs to this class of algorithms. The method computes a separating hyperplane in a high dimensional space induced by kernels performing dot products between points from the input space in high-dimensional space. The boundary is fitted to the input data by maximizing the margin between the data and the origin in the high-dimensional space. The algorithm prevents overfitting by allowing a percentage $v$ of data points to fall outside the boundary. This percentage $v$ acts as regularization parameter; it is used as a lower bound on the fraction of support vectors delimiting the boundary and as an upper bound on the fraction of margin errors, i.e. training points remaining outside the boundary.

The experiment of the original paper targets mostly novelty detection, i.e. anomaly detection using a model trained on a dataset free of anomalies. This thesis uses contaminated datasets to assess the algorithm robustness with a regularization parameter significantly higher than the expected proportion of outliers.

### 2.1.6 Isolation methods

We include an isolation algorithm which focuses on separating outliers from the rest of the data points. This method differs from the previous methods as it isolates anomalies instead of profiling normal points.

The concept of **Isolation forest** was brought by Liu in [Liu et al., 2008] and uses random forests to compute an isolation score for each data point. The algorithm performs recursive random splits over the feature domain until each sample is isolated from the rest of the dataset. As a result, outliers are separated after few splits and are located in nodes close to the root of the trees. The average path length required to reach the node containing the specified point is used for scoring.

The author states that his algorithm provides linear time complexity and demonstrates outlier detection performance significantly better than lof on real-world datasets.

### 2.1.7 KOHONEN NETWORKS

[Marsland et al., 2002] propose a reconstruction-based nonparametric neural network called **Grow When Required (GWR) network**. This method is based on Kohonen networks, also called Self-Organizing maps (SOM) [Kohonen, 1998], and fits a graph of adaptive topology lying in the input space to a dataset. While training the network, nodes (also called *neurons*) and edges are added or removed in order to best fit the data, the objective being to end up with nodes positioned in all dense data regions while edges propagate the displacement of neighboring nodes.

Outliers from synthetic datasets are detected using fixed-topology SOM in an experimental work [Munoz & Muruzábal, 1998]. The paper uses two thresholds $t_1$ and $t_2$ to identify data points having their closest node further than $t_1$, or projecting on an outlying node, i.e. a neuron having a median interneuron distance (MID) higher than $t_2$. The MID of each neuron is computed by taking the median of the distance between a neuron and its 8 neighbors in a network following a 2-dimensional grid topology. Severe outliers and dense clouds of outliers are correctly identified with this technique, though some nominal samples can be mistaken as mild outliers.

### 2.1.8 AUTOENCODERS

Autoencoders are neural networks architectures composed of an encoder and a decoder. The encoder maps the input data into a low-dimensional representation called the latent variables, while the decoder maps the latent variables into the input samples. This process allows the network to learn a compressed representation of the training set, which is usually achieved by minimizing the reconstruction error, i.e. the root-mean-square error (RMSE) between the input data and the reconstructed output, or by optimizing a custom loss function. To improve readability, we append the depth of the networks used in the thesis as a suffix to the name, e.g. VAE-2 for a 2-layer variational autoencoder (VAE).

In order to compare the model that we introduce and design in Section 4.1, this thesis uses two **feedforward autoencoders (AE)** with sigmoid activation functions in the hidden layers and a *dropout* mechanism [Srivastava et al., 2014a]. Dropout is a regularization technique which consists in ignoring randomly selected neurons during the training of the network. The first network (AE-1) is a single layer autoencoder while the second one (AE-5) has a 5-layer topology. We use the reconstruction error to score novelties.

**Variational autoencoders (VAE)** [Kingma & Welling, 2014] are generative models which compress the representation of the training data into a layer of latent variables, optimized with stochastic gradient descent. The sum of the reconstruction error and the latent loss, i.e. the negative of the Kullback-Leibler divergence between the approximate posterior over the latent variables and the prior, gives the loss term optimized during the training. VAE-1 is a shallow network using one layer for latent variables representation, while VAE-2 uses a two-layer architecture with a first layer for encoding and a second one for decoding. The reconstruction error is used to detect outliers.

**VAE-DGP** [Dai et al., 2016] is a neural network performing stochastic variational inference using inducing-point approximations to train a Deep Gaussian Process (DGP) model. DGPs are detailed in Chapter 4 and involve a composition of Gaussian processes, resulting in a deep probabilistic neural network. The inducing-point approximation required by the training process of VAE-DGP-2 involves matrix factorizations such as Cholesky decompositions. This method uses an additional multilayer perceptron as a recognition model in order to constrain the variational posterior distributions of latent variables.

**Neural Autoregressive Distribution Estimator (NADE)** [Uria et al., 2016] is a neural network architecture designed for density estimation. The network uses mixtures of Gaussians to model $p(x)$. The network yields an autoregressive model, which implies that the joint distribution is modelled such that the probability for a given feature depends on the previous features fed to the network, i.e. $p(\boldsymbol{x}) = p(x_{o_d}|\boldsymbol{x}_{o_{<d}})$, where $x_{o_d}$ is the feature of index $d$ of $\boldsymbol{x}$. We use NADE-2, the two-layer deep and orderless version of NADE.

## 2.2 Sequence based methods

The current section details novelty detection methods for sequential data from the literature. In order to provide recommendations relevant to real-world use cases, only methods satisfying the following constraints were selected: (1) the method accepts *discrete sequences of events* as input, where events are represented as categorical samples; (2) the sequences fed to the method may have *variable lengths*, which implies a dedicated support or a tolerance for padding; (3) the novelty detection problem induces a distinct training and testing dataset. As such, the selected approach should be able to perform *predictions on unseen data* which was not presented to the algorithm during the training phase; (4) subject to user inputs and system changes, the set of discrete symbols in the sequences (alphabet) of the training set cannot be assumed to be complete. The algorithm should support *new symbols from the test set*; (5) in order to perform an accurate evaluation of its novelty detection capabilities and to provide practical predictions on testing data, the method should provide *continuous anomaly scores* rather than a binary decision. This last point allows for a ranking of the anomalies, and hence a meaningful manual validation of the anomalies, or the application of a user-defined threshold in the case of automatic intervention. The ranking of anomalies is also required by the performance metric used in the study and described in section 5.1.2.

### 2.2.1 HIDDEN MARKOV MODEL

**Hidden Markov Models** (HMMs) [Rabiner, 1989] are popular graphical models used to describe temporal data and generate sequences. The approach fits a probability distribution over the space of possible sequences, and is widely used in speech recognition [Gales & Young, 2008] and protein modeling [Söding, 2005]. An HMM is composed of $N$ states which are interconnected by state-transition probabilities, each state generating emissions according to its own emission probability distribution and the previous emission. To generate a sequence, an initial state is first selected based on initial probabilities. A sequence of states is then sampled according to the transition matrix of the HMM. Once the sequence of states is obtained, each state emits a symbol based on its emission distribution. The sequence of emissions is the observed data. Based on a dataset composed of emission sequences, we can achieve the inverse process, i.e. estimate the transition matrix and the emission distributions of a HMM from the emissions observed. Possible sequences of *hidden* states leading to these emissions are thus inferred during the process. Once we obtain a trained HMM $\lambda = (A, B, \pi)$ with $A$ the transition matrix, $B$ describing the emission probabilities and $\pi$ the initial state probabilities, we can compute the normalized likelihood of a sequence and use it as a score to detect novelties.

### 2.2.2 DISTANCE-BASED METHODS

Distance-based approaches rely on pairwise distance matrices computed by applying a distance function to each pair of input sequences. The resulting matrix is then used by clustering or nearest-neighbor algorithms to build a model of the data. At test time, a second distance matrix is computed to perform scoring, which contains the distance between each test sample and the training data.

#### DISTANCE METRICS

LCS is the **longest common subsequence** [Bergroth et al., 2000] shared between two sequences. A common subsequence is defined as a sequence of symbols appearing in the same order in both sequences, although they do not need to be consecutive. As an example, we have LCS(XMJYAUZ, MZJAWXU) = MJAU. Since LCS expresses a similarity between sequences, we use the negative LCS to obtain a distance.

The **Levenshtein distance** [Levenshtein, 1966], also called the *edit distance*, is a widely used metric which computes the difference between two strings or sequences of symbols. It represents the minimum number of edit operations required to transform one sequence into another, such as insertions, deletions and substitutions of individual symbols.

Both metrics are normalized by the sum of the sequence lengths (equation 2.2), which makes them suitable for sequences of different length.

$$distance(x, y) = \frac{metric(x, y)}{|x| + |y|} \tag{2.2}$$

### Algorithms

The **$k$-nearest neighbors** ($k$-NN) algorithm is often used for classification and regression. In the case of classification, $k$-NN assigns to each test sample the label the most represented among its $k$ nearest neighbors from the training set. In [Ramaswamy et al., 2000], the scoring function used to detect outliers is the distance $d(x, n_k)$ or $d_k(x)$ between a point $x$ and its $k^{th}$ nearest neighbor $n_k$. This approach was applied to sequences in [Chandola et al., 2008] using the LCS metric, and outperformed methods such as HMM and RIPPER.

**Local outlier factor** (LOF) [Breunig et al., 2000] also studies the neighborhood of test samples to identify anomalies. It compares the local density of a point $x$ to the local density of its neighbors by computing the *reachability distance* $rd_k(x, y)$ between $x$ and each of its $k$-nearest neighbors $n_i$.

$$rd_k(x, n_i) = \max(d_k(n_i), d(x, n_i)) \tag{2.3}$$

The computed distances are then aggregated into a final anomaly score detailed in [Breunig et al., 2000]. The method showed promising results when applied to intrusion detection [Lazarevic et al., 2003].

**$k$-medoids** [Park & Jun, 2009] is a clustering algorithm which uses data points from the training set, also called *medoids*, to represent the center of a cluster. The algorithm first randomly samples $k$ medoids from the input data, then cluster the remaining data points by selecting the closest medoid. The medoids of each cluster are further replaced by a data point from the same cluster which minimizes the sum of distances between the new medoid and the points in the cluster. The method uses expectation-maximization and is very similar to $k$-means, although the latter uses the arithmetic mean of a cluster as a center, called *centroid*. Since $k$-means requires numerical data and is more sensitive to outliers [Park & Jun, 2009], it was not selected for this study. We use the distance to the closest medoid to detect anomalies, which is the method used in [Budalakoti et al., 2009] and [Budalakoti et al., 2006]. Both papers used the LCS metric to preprocess the data given to $k$-MEDOIDS.

### 2.2.3 Window-based techniques

The two following methods observe subsequences of fixed length, called *windows*, within a given sequence to identify abnormal patterns. This workflow requires to preprocess the data by applying a sliding window to each sequence, shifting the window by one symbol at each iteration and resulting in a larger dataset due to overlapping subsequences.

***t*-STIDE** [Warrender et al., 1999], which stands for *threshold-based sequence time-delay embedding*, uses a dictionary or a tree to store subsequences of length $k$ observed in the training data, along with their frequency. Once this model is built, the anomaly score of a test sequence is the number of subsequences within the sequence which do not exist in the model, divided by the number of windows in the test sequence. For increased robustness, subsequences having a frequency lower than a given threshold are excluded from the model. This increases the anomaly score for uncommon patterns, and allows the algorithm to handle datasets contaminated by anomalous sequences. This scoring method is called Locality Frame Count (LFC) and was applied to intrusion detection [Warrender et al., 1999] where it performed almost as well as HMM at a reduced computational cost.

**RIPPER** [Cohen, 1995] is a supervised classifier designed for association rule learning. The training data given to the algorithm is divided into a set of sequences of length $k$, and the corresponding labels. For novelty detection, subsequences are generated by a sliding window, and the label is the symbol following each subsequence. This allows RIPPER to learn rules predicting upcoming events. This method was applied to intrusion detection in [Lee et al., 1997]. To build an anomaly score for a test sequence, the authors retrieve the predictions obtained for each subsequence, along with the confidence of the rule which triggered the prediction. Each time a prediction does not match the upcoming event, the anomaly score is increased by $confidence * 100$. The final score is then divided by the number of subsequences for normalization.

### 2.2.4 Pattern mining

Sequential Pattern Mining (SPM) consists in the unsupervised discovery of interesting and relevant subsequences in sequential databases. A recent algorithm from this field is **Interesting Sequence Miner** (ism) [Fowkes & Sutton, 2016], a probabilistic and generative method which learns a set of patterns leading to the best compression of the database. From a training set, ism learns a set of interesting subsequences ranked by probability and interestingness. To score a test sequence, we count the number of occurrences of each interesting pattern returned by ism, and multiply the number of occurrences by the corresponding probability and interestingness. This score is normalized by the length of the test sequence, a low score denoting an anomaly. While alternatives to ism exist in the literature [Gan et al., 2018], few provide both a probabilistic framework and access to their code.

### 2.2.5 Neural networks

Recurrent neural networks (rnns) are widely used algorithms for a variety of supervised tasks related to temporal data [Lipton et al., 2015]. Long Short-Term Memory (lstm) [Hochreiter & Schmidhuber, 1997], a specific topology of rnn, has the ability to model long-term dependencies and thus arbitrary long sequences of events. This network can be applied to unsupervised learning problems by using an autoencoder topology, i.e. using identical input and output layers to present the same data in input and output to the network. This allows the method to learn a compressed representation of the data. For this purpose, the following algorithms use two multilayer lstm networks, the first one encoding the data in a vector of fixed dimensionality (encoder), the second one decoding the target sequence from the vector (decoder).

The **Sequence to Sequence** (seq2seq) [Sutskever et al., 2014] network is a recent work designed for language translation. The method is based on lstm cells and uses various mechanisms such as *dropout* to prevent overfitting and *attention* [Luong et al., 2015] to focus on specific past events to establish correlations. Attention is a masking mechanism which allows a neural network to focus on a subset of its inputs. As suggested in [Sakurada & Yairi, 2014, Marchi et al., 2015], the reconstruction error is used to score anomalies. The reconstruction error is the distance between the input and the reconstructed output, computed by lcs in this study.

We also include a simpler **lstm Autoencoder** (lstm-ae) for the sake of the comparison, paired with a different scoring system. This network is also composed of two lstm networks, and both seq2seq and lstm-ae perform masking to handle padding characters appended to the end of the sequences of variable length. However, LSTM-AE does not benefit from the dropout and attention mechanisms. In addition, instead of comparing the input to the reconstructed output for scoring, we now apply a distinct novelty detection algorithm to the latent representation provided by the network. The goal of LSTM-AE is thus to learn a numerical fixed-length vector to represent each input sequence. The resulting representation of the training set is then given to the Isolation Forest algorithm. At test time, the input sequence is encoded into a vector which is scored by the trained Isolation Forest.

*The key to artificial intelligence has always been the representation.*

Jeff Hawkins

# 3

# Dirichlet Process Mixture Model for novelty detection

In this chapter, we present an algorithm suitable to model mixed-type data and perform an experimental survey of novelty detection methods. More specifically, our contributions are detailed as follows: (i) we introduce the Dirichlet Process Mixture Model (DPMM), an algorithm built upon Dirichlet Process mixtures [Blei & Jordan, 2006], trained through variational inference [Bishop, 2006] and using a Gamma prior on the Dirichlet Process [Escobar & West, 1995]; (ii) we provide new representations in the exponential family of frequently used likelihoods, conjugate priors and posteriors, making DPMM suitable to model mixed-type features through a product of exponential family distributions; (iii) we evaluate the performance of DPMM on novelty detection tasks; (iv) we perform an experimental survey on novelty detection for numerical and mixed-type data, comparing the accuracy and scalability of state-of-the-art methods from the litterature.

The Dirichlet Process Mixture Model is a probabilistic nonparametric model trained through variational inference [Jordan et al., 1999]. The algorithm is an unsupervised clustering and density estimation method in which the number of components in the mixture grows as new data are observed. The number of components, the mixing proportions and the parameters of the posterior are learnt variationally. In the Dirichlet process (DP) mixture, the observations are drawn from an exponential family distribution, which provides a flexible and accurate model suitable for mixed type features. The analytical representation in the exponential-family for suitable likelihoods, conjugate priors and posteriors is required for the derivation of the method. This work aggregates the inference of DP mixtures through variational methods presented in [Bishop, 2006], the use of a Beta prior on the Dirichlet process responsible for the mixing proportions in [Blei & Jordan, 2006] and the application of a Gamma prior on the scaling parameter of the Dirichlet process proposed by [Escobar & West, 1995].

The resulting DPMM is evaluated on a wide set of novelty detection tasks. The inherent complexity of novelty detection induced by the contamination of the training data with anomalies and the varying shape, size and density across clusters strongly motivate an experimental review of the field. In the second part of this chapter, we extend our review of novelty detection methods (Chapter 2) by performing a thorough experimental comparison bringing together numerous state-of-the-art algorithms.

This study extends previous works [Emmott et al., 2016, Zimek et al., 2012] by using 12 publicly available labelled datasets, most of which are recommended for outlier detection in [Emmott et al., 2016], in addition to 3 novel industrial datasets from the production systems of Amadeus, a major company in the travel industry. The benchmark made in [Emmott et al., 2016] used fewer datasets and methods, and solely numerical features while we benchmark and address ways to handle categorical data. While several previous works identify anomalies solely in the training set, our study tests for the generalization ability of all methods by detecting novelties in unseen testing data. The selected parametric and nonparametric algorithms belong to a variety of approaches including probabilistic algorithms, nearest-neighbor based methods, neural networks, information theoretic and isolation methods. The performance on labelled datasets are evaluated by the area under the ROC and precision-recall curves.

In order to give a full overview of these methods, we also benchmark the training time, prediction time, memory usage and robustness of each method when increasing the number of samples, features and the background noise. These measurements allow us to compare algorithms not only based on their outlier detection performance but also on their scalability and suitability for large dimensional problems.

The chapter is organized as follows: section 3.1 presents the proposed DPMM algorithm, section 3.2 details the experimental setup, the public and proprietary datasets and the generation process for the synthetic datasets; section 3.3 presents the results of the comparison and section 3.4 summarizes our conclusions.

## 3.1 Dirichlet Process Mixture Model

A Dirichlet Process Mixture Model is defined by a set of latent variables and parameters $\boldsymbol{W}$. Given a set of observations $\boldsymbol{X}$, we first consider the joint density

$$p(\boldsymbol{X}, \boldsymbol{W}) = p(\boldsymbol{W})p(\boldsymbol{X}|\boldsymbol{W}). \tag{3.1}$$

Our Bayesian model draws the latent variables from a prior $p(\boldsymbol{W})$ and aims to model the likelihood $p(\boldsymbol{X}|\boldsymbol{W})$ which is often intractable, i.e. the corresponding high-dimensional integral cannot be solved analytically to obtain a closed-form solution. The model inference consists in computing the posterior distribution $p(\boldsymbol{W}|\boldsymbol{X})$. This task is achieved by approximating $p(\boldsymbol{W}|\boldsymbol{X})$ with the variational distribution $q(\boldsymbol{W})$ through variational inference, here using mean-field approximation.

Variational methods [Blei et al., 2017] provide a learning framework for graphical models where exact inference is not feasible. These methods use simplified graphical models in order to achieve approximate inference. The underlying idea is to propose a family of densities to approximate the posterior, and then to find the member of that family which is the closest to the true posterior. The distance between the two distributions is measured by the Kullback-Leibler divergence. Markov Chain Monte Carlo (MCMC) techniques, such as Gibbs sampling [Andrieu et al., 2003], are alternative approximate inference methods for posterior estimation. While MCMC methods produce asymptotically exact samples from the target density, they are computationally much more expensive than variational inference which does not come with this guarantee. Variational inference is thus well-suited to build scalable methods able to tackle large datasets.

### 3.1.1 Model representation

The method uses a flexible mixture of exponential-family distributions to model the input data. The parameters of the corresponding likelihood are sampled from the posterior. As such, this model supports a number of distinct likelihoods, under the condition that the chosen likelihood, its conjugate prior and the corresponding posterior can be computed analytically within the exponential family. As an example, the conjugate prior of the multivariate Normal is the Normal-Wishart distribution. In the case of a Gaussian mixture likelihood, the algorithm is given the hyperparameters of the Normal-Wishart prior and learn the parameters of the corresponding posterior. The range of possible distributions and their derivation in the exponential-family are discussed in Section 3.1.2. Since the product of several exponential-family distributions is also in the exponential-family, each component of the mixture can be defined as a product of distributions. This representation allows our model to provide a general density estimation framework compatible with several probability distributions and suitable for mixed-type data.

The mixing proportions $\boldsymbol{\pi}$ of the mixture are described by a Dirichlet process (DP) described in equation 3.2 for $k = \{1, 2, \dots\}$.

$$\pi_k(\boldsymbol{v}) = v_k \prod_{j=1}^{k-1} (1 - v_j) \tag{3.2}$$

Intuitively, a DP is a stick-breaking process with an infinite number of components, where the weights $v_k$ are sampled from the following Beta distribution:

$$v_k \sim \text{Beta}(1, w) \tag{3.3}$$

In practice, we are going to put a truncation parameter $K$ on the number of components to make the inference tractable. Although $K$ can be very high, the actual number of components used in the mixture is usually much smaller, as a component $k$ vanishes when $v_k \approx 0$. This allows the number of components to grow with the complexity of the data. A truncation parameter implies that $\pi_k(\boldsymbol{v}) = 0$ for $k > K$, which is achieved by setting $v_K = 1$. While $w$ could be a defined as a hyperparameter, its value has a strong impact on the mixing proportions and on the number of components used in the approximation of the posterior. We thus integrate over $w$ and learn this parameter variationally.

We observe that the shape parameter $\alpha$ of the Beta distribution is fixed to 1. If we had used used a hyperparameter $\alpha_0$ instead, the variational distribution $q^*_{\alpha_k, \beta_k}(v_k)$ (eq. 3.29) would remain a Beta distribution of parameters $\alpha_k = \alpha_0 + N_k$ while $\beta_k$ (eq. 3.32) would be unchanged. However, we could no longer integrate out $w$ as $q^*_{g_1, g_2}(w)$ (eq. 3.37) would not be a Gamma distribution. Learning $w$ variationally comes thus with the constraint $\alpha_0 = 1$.

### 3.1.2 A general formulation in the exponential family

The exponential family of probability distributions includes all distributions for which the density can be written in the general form described in equation 3.4, where $h(\boldsymbol{x})$ is a function, $\boldsymbol{\eta^*}$ is called the natural parameter, $T(\boldsymbol{x})$ the sufficient statistics and $a(\boldsymbol{\eta^*})$ the normalization factor.

$$p(\boldsymbol{x}|\boldsymbol{\eta^*}) = h_l(\boldsymbol{x}) \exp\left(\boldsymbol{\eta^*}^T T(\boldsymbol{x}) - a_l(\boldsymbol{\eta^*})\right) \tag{3.4}$$

Given the previous exponential-family likelihood, the corresponding conjugate prior is

$$p(\boldsymbol{\eta^*}|\boldsymbol{\lambda}) = h_p(\boldsymbol{\eta^*}) \exp\left(\boldsymbol{\lambda_1}^T \boldsymbol{\eta^*} + \lambda_2(-a_l(\boldsymbol{\eta^*})) - a_p(\boldsymbol{\lambda})\right), \tag{3.5}$$

where $\boldsymbol{\lambda_1}$ and $\boldsymbol{\eta^*}$ have the same dimensionality and $\lambda_2$ is a scalar. The vector of sufficient statistics [Nielsen & Garcia, 2009] is thus $\left(\boldsymbol{\eta^*}^T, -a_l(\boldsymbol{\eta^*})\right)^T$. The natural parameter of the likelihood and the conjugate prior are $\boldsymbol{\eta^*}$ and $\boldsymbol{\lambda}$, respectively. The conjugate prior has thus one parameter more than the likelihood. The subscripts $l$ and $p$ identify the base measure $h$ and log-partition $a$ belonging to the likelihood and conjugate prior, respectively, although the parameters of these terms suffice to make the distinction.

If the conjugate prior for the chosen likelihood can be expressed analytically in exponential-family representation, the posterior can be expressed as

$$p(\boldsymbol{\eta^*}|\boldsymbol{\tau}) = h_p(\boldsymbol{\eta^*}) \exp\left(\boldsymbol{\tau_1}^T \boldsymbol{\eta^*} + \tau_2(-a_l(\boldsymbol{\eta^*})) - a_p(\boldsymbol{\tau})\right). \tag{3.6}$$

We can then compute the expectation over each term of the vector of sufficient statistics [Nielsen & Garcia, 2009]:

$$\mathbb{E}[\boldsymbol{\eta^*}] = \frac{\partial a_p(\boldsymbol{\tau_1}, \cdots)}{\partial \boldsymbol{\tau_1}} \tag{3.7}$$

$$\mathbb{E}[-a_l(\boldsymbol{\eta^*})] = \frac{\partial a_p(\cdots, \tau_2)}{\partial \tau_2} \tag{3.8}$$

More generally, given a likelihood and a conjugate prior, we can compute the following posterior:

$$\begin{aligned}
p(\boldsymbol{\eta^*}|\boldsymbol{X_{1:N}}, \lambda) &\propto p(\boldsymbol{\eta^*}|\lambda) \prod_{i=1}^{N} p(\boldsymbol{x_i}|\boldsymbol{\eta^*}) \\
&\propto h_p(\boldsymbol{\eta^*}) \exp\left(\boldsymbol{\lambda_1}^T \boldsymbol{\eta^*} + \lambda_2(-a_l(\boldsymbol{\eta^*})) - a_p(\lambda)\right) \\
&\quad \cdot \left(\prod_{i=1}^{N} h_l(\boldsymbol{x_i})\right) \exp\left(\boldsymbol{\eta^*}^T \sum_{i=1}^{N} T(\boldsymbol{x_i}) - N \cdot a_l(\boldsymbol{\eta^*})\right) \\
&\propto h_p(\boldsymbol{\eta^*}) \exp\left(\left(\boldsymbol{\lambda_1} + \sum_{i=1}^{N} T(\boldsymbol{x_i})\right)^T \boldsymbol{\eta^*} + (\lambda_2 + N)(-a_l(\boldsymbol{\eta^*}))\right)
\end{aligned} \tag{3.9}$$

This yields the posterior parameters

$$
\begin{cases}
\boldsymbol{\tau_1} = \boldsymbol{\lambda_1} + \sum_{i=1}^{N} T(\boldsymbol{x_i}), \\
\tau_2 = \lambda_2 + N.
\end{cases}
\tag{3.10}
$$

The derivations of the exponential-family representation for several probability distributions are detailed in Appendix A. In order to satisfy the requirements previously mentioned, Appendix B details the derivations of several conjugate priors in exponential family based on equation 3.5. This allows us to infer the expectation over the vector of sufficient statistics for the posterior (eq. 3.7 and 3.7) which is required in equation 3.25. Note that a few conjugate priors cannot be expressed analytically as exponential-family distributions, e.g. the conjugate prior for a Gamma likelihood. This prevents the use of such likelihoods in our model.

We report in Table 3.1 the parameter mapping from several probability distributions to their exponential-family representation. The inverse parameter mapping is used to retrieve the original parameters from the natural parameters of the exponential-family distribution.

| Distribution | Parameter(s) $\boldsymbol{\theta}$ | Natural parameters $\boldsymbol{\eta}$ | Inverse parameter mapping | Base measure $h(\boldsymbol{x})$ | Sufficient statistic $T(\boldsymbol{x})$ | Log-partition $a(\boldsymbol{\theta})$ |
|---|---|---|---|---|---|---|
| Binomial ($n$ trials) | $p$ | $\ln\frac{p}{1-p}$ | $\frac{1}{1+e^{-\eta}}$ | $\binom{n}{x}$ | $x$ | $-n\ln(1-p)$ |
| Multinomial ($n$ trials) | $p1,\cdots,p_k$ with $\sum_{i=1}^{k}p_i=1$ | $\begin{pmatrix}\ln p_1\\ \vdots\\ \ln p_k\end{pmatrix}$ | $\begin{pmatrix}e^{\eta_1}\\ \vdots\\ e^{\eta_k}\end{pmatrix}$ with $\sum_{i=1}^{k}e^{\eta_i}=1$ | $\frac{n!}{\prod_{i=1}^{n}x_i!}$ | $\begin{pmatrix}x_1\\ \vdots\\ x_k\end{pmatrix}$ | $0$ |
| Beta | $\alpha,\beta$ | $\begin{pmatrix}\alpha-1\\ \beta-1\end{pmatrix}$ | $\begin{pmatrix}\eta_1+1\\ \eta_2+1\end{pmatrix}$ | $1$ | $\begin{pmatrix}\ln x\\ \ln(1-x)\end{pmatrix}$ | $\ln\Gamma(\alpha)+\ln\Gamma(\beta)-\ln\Gamma(\alpha+\beta)$ |
| Dirichlet | $\alpha_1,\cdots,\alpha_k$ | $\begin{pmatrix}\alpha_1-1\\ \vdots\\ \alpha_k-1\end{pmatrix}$ | $\begin{pmatrix}\eta_1+1\\ \vdots\\ \eta_k+1\end{pmatrix}$ | $1$ | $\begin{pmatrix}\ln x_1\\ \vdots\\ \ln x_k\end{pmatrix}$ | $\sum_{i=1}^{k}\ln\Gamma(\alpha_i)-\ln\Gamma\left(\sum_{j=1}^{k}\alpha_j\right)$ |
| Gamma | $\alpha,\beta$ | $\begin{pmatrix}\alpha-1\\ -\beta\end{pmatrix}$ | $\begin{pmatrix}\eta_1+1\\ -\eta_2\end{pmatrix}$ | $1$ | $\begin{pmatrix}\ln x\\ x\end{pmatrix}$ | $\ln\Gamma(\alpha)-\alpha\ln\beta$ |
| Poisson | $\lambda$ | $\ln\lambda$ | $e^{\eta}$ | $\frac{1}{x!}$ | $x$ | $\lambda$ |
| Multivariate normal ($k$ dimensions) | $\boldsymbol{\mu},\boldsymbol{\Sigma}$ | $\begin{pmatrix}\boldsymbol{\Sigma}^{-1}\boldsymbol{\mu}\\ -\frac{1}{2}\boldsymbol{\Sigma}^{-1}\end{pmatrix}$ | $\begin{pmatrix}-\frac{1}{2}\boldsymbol{\eta_2}^{-1}\boldsymbol{\eta_1}\\ -\frac{1}{2}\boldsymbol{\eta_2}^{-1}\end{pmatrix}$ | $(2\pi)^{-\frac{k}{2}}$ | $\begin{pmatrix}\boldsymbol{x}\\ \boldsymbol{xx}^T\end{pmatrix}$ | $\frac{1}{2}\boldsymbol{\mu}^T\boldsymbol{\Sigma}^{-1}\boldsymbol{\mu}+\frac{1}{2}\ln|\boldsymbol{\Sigma}|$ |
| Wishart ($k$ dimensions) | $\boldsymbol{V},n$ | $\begin{pmatrix}-\frac{1}{2}\boldsymbol{V}^{-1}\\ \frac{n-d-1}{2}\end{pmatrix}$ | $\begin{pmatrix}-\frac{1}{2}\boldsymbol{\eta_1}^{-1}\\ 2\eta_2+d+1\end{pmatrix}$ | $1$ | $\begin{pmatrix}\boldsymbol{x}\\ \ln|\boldsymbol{x}|\end{pmatrix}$ | $\frac{n}{2}(d\ln 2+\ln|\boldsymbol{V}|)+\ln\Gamma_d\left(\frac{n}{2}\right)$ |
| Normal-Wishart ($k$ dimensions) | $\boldsymbol{\mu}_0,\lambda,\boldsymbol{V},n$ | $\begin{pmatrix}\frac{n-d}{2}\\ -\frac{1}{2}(\boldsymbol{\mu_0\mu_0}^T\lambda+\boldsymbol{V}^{-1})\\ \boldsymbol{\mu_0}\lambda\\ -\frac{1}{2}\lambda\end{pmatrix}$ | $\begin{pmatrix}-\frac{\boldsymbol{\eta_3}}{2\eta_4}\\ -2\eta_1\\ \left(-2\boldsymbol{\eta_2}+\frac{\boldsymbol{\eta_3\eta_3}^T}{2\eta_4}\right)^{-1}\\ 2\eta_1+d\end{pmatrix}$ | $(2\pi)^{-\frac{k}{2}}$ | $\begin{pmatrix}\ln|\boldsymbol{\Lambda}|\\ \boldsymbol{\Lambda}\\ \boldsymbol{x}^T\boldsymbol{\Lambda}\\ \boldsymbol{\Lambda xx}^T\end{pmatrix}$ | $-\frac{d}{2}\ln\lambda+\frac{nd}{2}\ln 2+\frac{n}{2}\ln|\boldsymbol{V}|+\ln\Gamma_d(\frac{n}{2})$ |
| Conjugate prior of Gamma $f(\alpha,\beta\mid p,q,r,s)\propto\frac{p^{\alpha-1}e^{-\beta q}}{\Gamma(\alpha)^r\beta^{-\alpha s}}$ | $p,q,r,s$ | $\begin{pmatrix}r\\ s\\ \ln p\\ -q\end{pmatrix}$ | $\begin{pmatrix}e^{\eta_3}\\ -\eta_4\\ \eta_1\\ \eta_2\end{pmatrix}$ | $1$ | $\begin{pmatrix}\ln\Gamma(\alpha)\\ \alpha\ln\beta\\ \alpha\\ \beta\end{pmatrix}$ | $\ln p$ |
| Conjugate prior of Beta $\pi(\alpha,\beta\mid\lambda_0,x_0,y_0)\propto\left(\frac{\Gamma(\alpha+\beta)}{\Gamma(\alpha)\Gamma(\beta)}\right)^{\lambda_0}x_0^{\alpha}y_0^{\beta}$ | $\lambda_0,x_0,y_0$ | $\begin{pmatrix}\lambda_0\\ \ln x_0\\ \ln y_0\end{pmatrix}$ | $\begin{pmatrix}\eta_1\\ e^{\eta_2}\\ e^{\eta_3}\end{pmatrix}$ | $1$ | $\begin{pmatrix}\ln\frac{\Gamma(\alpha+\beta)}{\Gamma(\alpha)\Gamma(\beta)}\\ \alpha\\ \beta\end{pmatrix}$ | $0$ |

Table 3.1: Exponential-family representation for several probability distributions.

### 3.1.3 Likelihood selection

Selecting a suitable likelihood based on the format and range of the input data has an important impact on the quality of the model representation. For this reason, we report in Table 3.2 possible choices of likelihood and conjugate prior depending on the format of a given set of features. Note that data belonging to the domain $[a, b]$, $[a, +\infty[$ or $]-\infty, a]$ can be scaled to fit in the domains $[0, 1]$, $[0, +\infty[$ and $[0, +\infty[$, respectively. In the table, rows highlighted in dark gray do not have a well-known conjugate prior in the litterature. Light gray rows denote a conjugate prior for which the normalization factor does not have an analytical form (see Appendix B). This last point prevents the evaluation of the expectation over the vector of sufficient statistics $\mathbb{E}_q[\boldsymbol{\eta}^*]$ and $\mathbb{E}_q[-a(\boldsymbol{\eta}^*)]$ for the posterior. This computation requires indeed the derivative of the unknown log-partition term, which could possibly be approximated with Markov chain Monte Carlo.

Table 3.2: Likelihood and conjugate prior per feature range

| Data description | Domain | Multivariate | Likelihood | Conjugate prior |
|---|---|---|---|---|
| Float $\in [0, 1]$ | $[0, 1]$ | No | Beta | $\propto \left(\frac{\Gamma(\alpha+\beta)}{\Gamma(\alpha)\Gamma(\beta)}\right)^{\lambda_0} x_0^{\alpha} y_0^{\beta}$ |
| Float $\in [0, 1]$ | $[0, 1]$ | Yes | Dirichlet | $\propto \frac{1}{B(\alpha)^{\eta}} e^{-\sum_{t=1}^{d} v_t \alpha_t}$ |
| Integer $\in [0, +\infty[$ | $\mathbb{N}$ | No | Poisson | Gamma |
| Integer $\in [0, +\infty[$ | $\mathbb{N}$ | Yes | Mult. Poisson | n/a |
| Float $\in [0, +\infty[$ | $\mathbb{R}^+$ | No | Gamma | $\propto \frac{p^{\alpha-1}e^{-\beta q}}{\Gamma(\alpha)^r \beta^{-\alpha s}}$ |
| Float $\in [0, +\infty[$ | $\mathbb{R}^+$ | Yes | Mult. Gamma | n/a |
| Float $\in ]-\infty, +\infty[$ | $\mathbb{R}$ | No | Normal | Normal-Gamma |
| Float $\in ]-\infty, +\infty[$ | $\mathbb{R}$ | Yes | Mult. Normal | Normal-Wishart |
| Boolean | $\{0, 1\}$ | No | Binomial | Beta |
| Boolean | $\{0, 1\}$ | Yes | Mult. Binomial | n/a |
| String, Boolean | $.^*$ | No | Multinomial | Dirichlet |

Important limitations are highlighted in Table 3.2 for the domains $[0, 1]$ and $[0, +\infty[$. While the suitable choices of likelihoods are the Beta and Gamma distributions, the lack of conjugate priors prevents us to resort to these distributions. To overcome this constraint, we propose to map the data matching the two previous ranges into the domain $]-\infty, +\infty[$. This would make the multivariate Normal likelihood suitable to model the data.

Let $\phi_p(x)$ be the cumulative distribution function (CDF) of a probability distribution $p$ and $F_p^{-1}(x)$ be the inverse cumulative distribution of this distribution, a.k.a. the quantile function. Based on the properties of these functions, for a given $x \in [0, 1]$ we obtain $F_N^{-1}(x) \in ]-\infty, +\infty[$. Similarly, the mapping of $x \in [0, +\infty[$ is achieved by $F_N^{-1}(\phi_\Gamma(x)) \in ]-\infty, +\infty[$. Note that these mappings can be reversed, resulting in the original data without loss of information. We plot the CDF and inverse CDF of $N(\mu = 0, \sigma = 1)$ in Figures 3.1 and 3.2, while Figures 3.3 and 3.4 show the CDF and inverse CDF of $\Gamma(shape = 1, scale = 2)$.

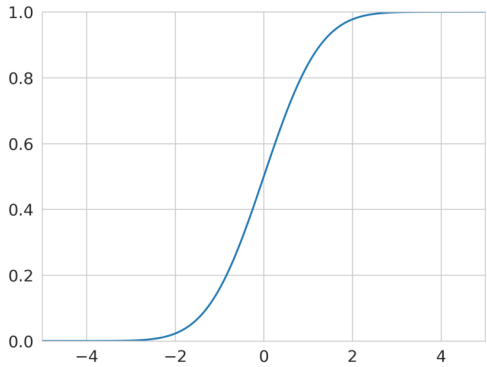

Figure 3.1: Normal CDF
$\phi_{N(\mu=0,\sigma=1)}(x)$

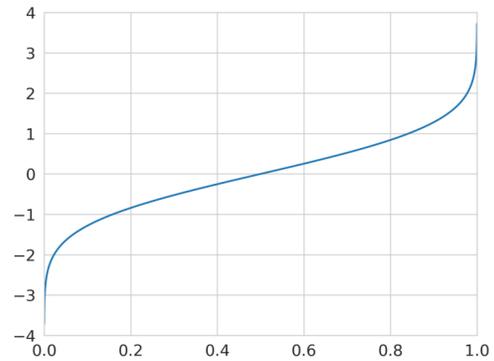

Figure 3.2: Normal inverse CDF
$F_{N(\mu=0,\sigma=1)}^{-1}(x)$

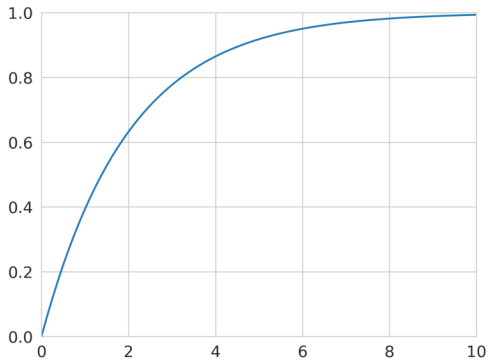

Figure 3.3: Gamma CDF
$\phi_{\Gamma(k=1,\theta=2)}(x)$

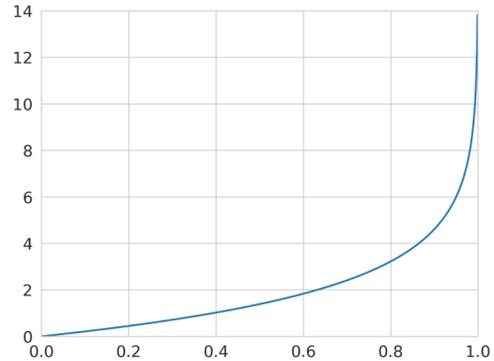

Figure 3.4: Gamma inverse CDF
$F_{\Gamma(k=1,\theta=2)}^{-1}(x)$

### 3.1.4 Lower bound optimization

We now introduce the prior distribution $p(\boldsymbol{W}|\boldsymbol{\theta})$ over the parameters $\boldsymbol{W}$, where $\boldsymbol{\theta}$ are the hyperparameters of the prior. Our model is trained by optimizing a lower bound on the log marginal likelihood $p(\boldsymbol{X}|\boldsymbol{\theta})$. We use the Kullback-Leibler divergence to perform the derivation of the marginal likelihood,

$$D_{KL}(q||p) = \int q(\boldsymbol{W}) \ln \frac{q(\boldsymbol{W})}{p(\boldsymbol{W}|\boldsymbol{X},\boldsymbol{\theta})} d\boldsymbol{W}. \tag{3.11}$$

This KL divergence is equal to 0 when $q(\boldsymbol{W})$ equals the posterior $p(\boldsymbol{W}|\boldsymbol{X},\boldsymbol{\theta})$. We aim at minimizing the divergence from $q$ to $p$ in order to learn an accurate approximation of the true posterior.

$$D_{KL}(q||p) = -\int q(\boldsymbol{W}) \ln \frac{p(\boldsymbol{W}|\boldsymbol{X},\boldsymbol{\theta})}{q(\boldsymbol{W})} d\boldsymbol{W}$$

$$D_{KL}(q||p) = \ln p(\boldsymbol{X}|\boldsymbol{\theta}) - \int q(\boldsymbol{W}) \ln \frac{p(\boldsymbol{W},\boldsymbol{X}|\boldsymbol{\theta})}{q(\boldsymbol{W})} d\boldsymbol{W}$$

$$\ln p(\boldsymbol{X}|\boldsymbol{\theta}) = \boldsymbol{\mathcal{L}}(q,\boldsymbol{\theta}) + D_{KL}(q||p) \tag{3.12}$$

Maximizing the lower bound $\boldsymbol{\mathcal{L}}$ defined in equation 3.13 is thus equivalent to minimizing $D_{KL}(q||p)$.

$$\boldsymbol{\mathcal{L}} = \int q(\boldsymbol{W}) \ln \frac{p(\boldsymbol{W},\boldsymbol{X}|\boldsymbol{\theta})}{q(\boldsymbol{W})} d\boldsymbol{W} \tag{3.13}$$

Optimizing equation 3.13 is also achieved by maximizing the log marginal likelihood defined in equation 3.14 where $\mathbb{E}_q$ is the expectation with respect to the distribution $q$. This yields a lower bound on the log marginal likelihood which can be optimized w.r.t. $q$.

$$\ln p(\boldsymbol{X}|\boldsymbol{\theta}) \geq \boldsymbol{\mathcal{L}}(q,\boldsymbol{\theta})$$

$$\ln p(\boldsymbol{X}|\boldsymbol{\theta}) \geq \int q(\boldsymbol{W}) \ln \frac{p(\boldsymbol{W},\boldsymbol{X}|\boldsymbol{\theta})}{q(\boldsymbol{W})} d\boldsymbol{W}$$

$$\ln p(\boldsymbol{X}|\boldsymbol{\theta}) \geq \mathbb{E}_q[\ln p(\boldsymbol{W},\boldsymbol{X}|\boldsymbol{\theta})] - \mathbb{E}_q[\ln q(\boldsymbol{W})] \tag{3.14}$$

### 3.1.5 Approximation of the posterior

In order to make the approximated distribution $q$ tractable, we first assume that $q(\boldsymbol{W})$ factorizes over a partitioning of the latent variables.

$$q(\boldsymbol{W}) = \prod_{i=1}^{M} q_i(\boldsymbol{W}_i) \tag{3.15}$$

We now choose to approximate the posterior with the factorized family of variational distributions reported in equation 3.16, using $\boldsymbol{W} = \{\boldsymbol{v}, \boldsymbol{\eta^*}, \boldsymbol{z}, w\}$, where $q_{\boldsymbol{\alpha},\boldsymbol{\beta}}(\boldsymbol{v})$ is a product of Beta distributions, $q_\tau$ is a product of exponential-family distributions, $q_r(\boldsymbol{z})$ is a product of multinomials on the cluster assignment variable $\boldsymbol{z}$ and $q_{g_1,g_2}(w)$ is a Gamma distribution $\Gamma$. Note that the truncation on the number of components implies $q_{\alpha_K,\beta_K}(v_K = 1) = 1$.

$$q(\boldsymbol{v}, \boldsymbol{\eta^*}, \boldsymbol{z}, w) = q_{\boldsymbol{\alpha},\boldsymbol{\beta}}(\boldsymbol{v}) \cdot q_\tau(\boldsymbol{\eta^*}) \cdot q_r(\boldsymbol{z}) \cdot q_{g_1,g_2}(w) \tag{3.16}$$

We now write in equation 3.17 the joint probability of the random variables, with the hyperparameters $\boldsymbol{\theta} = \{\boldsymbol{\lambda}, s_0, r_0\}$, where $s_0$ and $r_0$ are respectively the shape and rate parameters of the Gamma prior on $w$. The corresponding graphical model is reported in Figure 3.5.

$$p(\boldsymbol{X}, \boldsymbol{v}, \boldsymbol{\eta^*}, \boldsymbol{z}, w | \boldsymbol{\theta}) = p(\boldsymbol{X}|\boldsymbol{z}, \boldsymbol{\eta^*}) \cdot p(\boldsymbol{\eta^*}|\boldsymbol{\lambda}) \cdot p(\boldsymbol{z}|\boldsymbol{v}) \cdot p(\boldsymbol{v}|w) \cdot p(w|s_0, r_0) \tag{3.17}$$

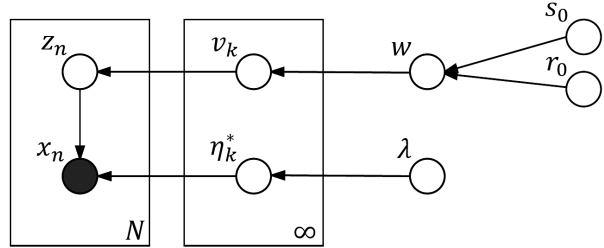

Figure 3.5: Graphical model of a Dirichlet Process Mixture Model in plate notation.

The previous distributions are defined hereafter, where $N$ denotes the number of input samples:

$$p(\boldsymbol{X}|\boldsymbol{z}, \boldsymbol{\eta^*}) = \prod_{n=1}^{N} \prod_{k=1}^{\infty} \left( h(\boldsymbol{x_n}) \exp(\boldsymbol{\eta_k^*}^T T(\boldsymbol{x_n}) - a(\boldsymbol{\eta_k^*})) \right)^{z_{nk}} \tag{3.18}$$

$$p(\boldsymbol{z}|\boldsymbol{v}) = \prod_{n=1}^{N}\prod_{k=1}^{K} Mult(\pi_k(\boldsymbol{v}))$$

$$= \prod_{n=1}^{N}\prod_{k=1}^{K} \pi_k(\boldsymbol{v})^{z_{nk}} \tag{3.19}$$

$$= \prod_{n=1}^{N}\prod_{k=1}^{K}(v_k\prod_{j=1}^{k-1}(1-v_j))^{z_{nk}}$$

$$p(\boldsymbol{\eta^*}|\boldsymbol{\lambda}) = \prod_{k=1}^{K} h(\boldsymbol{\eta_k^*})\exp(\boldsymbol{\lambda_1^T}\boldsymbol{\eta_k^*} + \lambda_2(-a(\boldsymbol{\eta_k^*})) - a(\boldsymbol{\lambda})) \tag{3.20}$$

$$p(\boldsymbol{v}|w) = \prod_{k=1}^{K} Beta(1, w) \tag{3.21}$$

$$p(w|s_0, r_0) = \Gamma(s_0, r_0) \tag{3.22}$$

### 3.1.6 COORDINATE ASCENT ALGORITHM

This section derives each term of equation 3.16 in order to provide an iterative expectation-maximization-like (EM-like) algorithm optimizing the lower bound on the log marginal likelihood. The expectation step comprises equations 3.28 and 3.41 to 3.44. The maximization step includes equations 3.31, 3.32, 3.35, 3.36, 3.39 and 3.40. In the following equations, the star in a term such as $q_r^*(\boldsymbol{z})$ denotes the optimal solution.

Based on the mean-field approximation framework, it can be shown that the optimal solution $q_j^*$ for each of the factors $q_j$ is obtained by taking the log of the joint distribution over all variables and then computing the expectation with respect to all of the other factors $q_i$ for $i \neq j$:

$$\ln q_j^*(\boldsymbol{W}_j|\boldsymbol{X}) = \mathbb{E}_{i\neq j}[\ln p(\boldsymbol{X}, \boldsymbol{W})] + const \tag{3.23}$$

We now continue the derivation of equation 3.23 for each partition of latent variables.

$$
\begin{aligned}
\ln q_r^*(\boldsymbol{z}) &= \mathbb{E}_{v,\boldsymbol{\eta^*},w}[\ln p(\boldsymbol{X}, v, \boldsymbol{\eta^*}, \boldsymbol{z}, w)] + const \\
&= \mathbb{E}_{\boldsymbol{\eta^*}}[\ln p(\boldsymbol{X}|\boldsymbol{\eta^*}, \boldsymbol{z})] + \mathbb{E}_v[\ln p(\boldsymbol{z}|v)] + const \\
&= \sum_{n=1}^{N}\sum_{k=1}^{K} z_{nk}\Big( \ln h(\boldsymbol{x_n}) + \mathbb{E}_q[\boldsymbol{\eta_k^*}]^T T(\boldsymbol{x_n}) + \mathbb{E}_q[-a(\boldsymbol{\eta_k^*})] \\
&\quad + \mathbb{E}[\ln v_k] + \sum_{i=1}^{k-1}\mathbb{E}[\ln(1-v_k)]\Big) + const
\end{aligned}
\tag{3.24}
$$

Taking the exponential of both sides, we get

$$
q_r^*(\boldsymbol{z}) \propto \prod_{n=1}^{N}\prod_{k=1}^{K}\rho_{nk}^{z_{nk}}
$$

With

$$
\begin{aligned}
\ln \rho_{nk} &= \ln h(\boldsymbol{x_n}) + \mathbb{E}_q[\boldsymbol{\eta_k^*}]^T T(\boldsymbol{x_n}) + \mathbb{E}_q[-a(\boldsymbol{\eta_k^*})] \\
&\quad + \mathbb{E}[\ln v_k] + \sum_{i=1}^{k-1}\mathbb{E}[\ln(1-v_k)]
\end{aligned}
\tag{3.25}
$$

Where $h(\boldsymbol{x_n})$ and $T(\boldsymbol{x_n})$ are respectively the base measure and sufficient statistics of the likelihood distribution. Remember that $\forall n \sum_{k=1}^{K} z_{nk} = 1$ and $z_{nk} \in \{0,1\}$. We can thus get rid of the proportionality by normalizing:

$$
q_r^*(\boldsymbol{z}) = \prod_{n=1}^{N}\prod_{k=1}^{K} r_{nk}^{z_{nk}}
\tag{3.26}
$$

$$
r_{nk} = \frac{\rho_{nk}}{\sum_{i=1}^{K}\rho_{ni}}
\tag{3.27}
$$

Thus,

$$
\mathbb{E}[z_{nk}] = r_{nk}
\tag{3.28}
$$

Note that equation 3.25 requires the derivation in analytical form for the expectation of the sufficient statistic terms of the exponential-family distributions. In order to solve these terms, the chosen likelihood must first be expressed as an exponential family distribution. We must then compute the exponential-family representation of the conjugate prior and posterior (see Appendices A and B).

$$
\begin{aligned}
\ln q^*_{\boldsymbol{\alpha},\boldsymbol{\beta}}(\boldsymbol{v}) &= \mathbb{E}_{\boldsymbol{\eta^*},\boldsymbol{z},w}[\ln p(\boldsymbol{X},\boldsymbol{v},\boldsymbol{\eta^*},\boldsymbol{z},w)] + const \\
&= \mathbb{E}_{\boldsymbol{z}}[\ln p(\boldsymbol{z}|\boldsymbol{v})] + \mathbb{E}_w[\ln p(\boldsymbol{v}|w)] + const \\
&= \sum_{n=1}^{N}\sum_{k=1}^{K}\mathbb{E}[z_{nk}]\left(\ln v_k + \sum_{i=1}^{k-1}\ln(1-v_k)\right) + \sum_{k=1}^{K-1}\Bigg((1-1)\ln v_k \\
&\quad + (\mathbb{E}[w]-1)\ln(1-v_k) - (\ln\Gamma(1)+\ln\Gamma(\mathbb{E}[w])-\ln\Gamma(1+\mathbb{E}[w]))\Bigg) + const \\
&= \sum_{n=1}^{N}\sum_{k=1}^{K}r_{nk}\left(\ln v_k + \sum_{i=1}^{k-1}\ln(1-v_k)\right) \\
&\quad + \sum_{k=1}^{K-1}\Big((\mathbb{E}[w]-1)\ln(1-v_k) - \ln B(1,\mathbb{E}[w])\Big) + const \\
&= \sum_{k=1}^{K-1}\left(\mathbb{E}[w] + \sum_{n=1}^{N}\sum_{i=k+1}^{K}r_{ni} - 1\right)\ln(1-v_k) + \sum_{n=1}^{N}r_{nk}\ln v_k \\
&\quad - \ln B(1,\mathbb{E}[w]) + const
\end{aligned}
\tag{3.29}
$$

Where $B(\alpha,\beta) = \frac{\Gamma(\alpha)\Gamma(\beta)}{\Gamma(\alpha+\beta)}$. Taking the exponential of both sides, we recognize a product of Beta distributions resulting in equation 3.30.

$$
q^*_{\boldsymbol{\alpha},\boldsymbol{\beta}}(\boldsymbol{v}) = \prod_{k=1}^{K-1} Beta(\alpha_k,\beta_k)
\tag{3.30}
$$

With

$$
\alpha_k = 1 + N_k
\tag{3.31}
$$

$$
\beta_k = \mathbb{E}[w] + \sum_{n=1}^{N}\sum_{i=k+1}^{K} r_{ni}
\tag{3.32}
$$

And where $N_k = \sum_{n=1}^{N} r_{nk}$.

$$
\begin{aligned}
\ln q_{\boldsymbol{\tau}}^*(\boldsymbol{\eta^*}) &= \mathbb{E}_{\boldsymbol{v,z},w}[\ln p(\boldsymbol{X,v,\eta^*,z},w)] + const \\
&= \mathbb{E}_{\boldsymbol{z}}[\ln p(\boldsymbol{X}|\boldsymbol{\eta^*,z})] + \ln p(\boldsymbol{\eta^*}|\boldsymbol{\lambda}) + const \\
&= \sum_{n=1}^{N}\sum_{k=1}^{K} \mathbb{E}[z_{nk}]\left(\ln h(\boldsymbol{x_n}) + \boldsymbol{\eta_k^*}^T T(\boldsymbol{x_n}) - a(\boldsymbol{\eta_k^*})\right) \\
&\quad + \sum_{k=1}^{K}\left(\ln h(\boldsymbol{\eta_k^*}) + \boldsymbol{\lambda_1}^T \boldsymbol{\eta_k^*} - \lambda_2 a(\boldsymbol{\eta_k^*}) - a(\boldsymbol{\lambda})\right) + const \\
&= \sum_{n=1}^{N}\sum_{k=1}^{K}\Bigg(\ln\left(h(\boldsymbol{x_n})^{r_{nk}} h(\boldsymbol{\eta_k^*})\right) + (r_{nk}T(\boldsymbol{x_n}) + \boldsymbol{\lambda_1})^T \boldsymbol{\eta_k^*} - (\lambda_2 + r_{nk})\, a(\boldsymbol{\eta_k^*}) \\
&\quad - a(\boldsymbol{\lambda})\Bigg) + const
\end{aligned}
\tag{3.33}
$$

The exponential of this term is an exponential-family distribution (eq. 3.34) of parameters $\boldsymbol{\tau_{k1}}$ and $\tau_{k2}$.

$$
q_{\boldsymbol{\tau}}^*(\boldsymbol{\eta^*}) = \prod_{k=1}^{K} h(\boldsymbol{\eta_k^*}) \exp(\boldsymbol{\tau_{k1}}^T \boldsymbol{\eta_k^*} + \tau_{k2}(-a(\boldsymbol{\eta_k^*})) - a(\boldsymbol{\tau_k}))
\tag{3.34}
$$

$$
\boldsymbol{\tau_{k1}} = \boldsymbol{\lambda_1} + \sum_{n=1}^{N} r_{nk} T(\boldsymbol{x_n})
\tag{3.35}
$$

$$
\tau_{k2} = \lambda_2 + \sum_{n=1}^{N} r_{nk}
\tag{3.36}
$$

The derivation of the last term gives:

$$
\begin{aligned}
\ln q^*_{g_1,g_2}(w) &= \mathbb{E}_{\boldsymbol{v},\boldsymbol{\eta^*},\boldsymbol{z}}[\ln p(\boldsymbol{X},\boldsymbol{v},\boldsymbol{\eta^*},\boldsymbol{z},w)] + const \\
&= \mathbb{E}_{\boldsymbol{v}}[\ln p(\boldsymbol{v}|w)] + \ln p(w|s_0,r_0) + const \\
&= \sum_{k=1}^{K-1} \big((w-1)\mathbb{E}_q[\ln(1-v_k)] - \ln\Gamma(w) + \ln\Gamma(w+1)\big) \\
&\quad - \ln\Gamma(s_0) + s_0\ln r_0 + (s_0-1)\ln w - r_0 w + const \\
&= (w-1)\sum_{k=1}^{K-1}\mathbb{E}_q[\ln(1-v_k)] + (K-1)\ln\frac{w\Gamma(w)}{\Gamma(w)} \\
&\quad - \ln\Gamma(s_0) + s_0\ln r_0 + (s_0-1)\ln w - r_0 w + const \\
&= (s_0 - 2 + K)\ln w - \left(r_0 - \sum_{k=1}^{K-1}\mathbb{E}_q[\ln(1-v_k)]\right)w \\
&\quad - \mathbb{E}_q[\ln(1-v_k)] - \ln\Gamma(s_0) + s_0\ln r_0 + const
\end{aligned}
\tag{3.37}
$$

We obtain a Gamma distribution of shape $g_1$ and rate $g_2$:

$$
q^*_{g_1,g_2}(w) = \Gamma(g_1,g_2) \tag{3.38}
$$

$$
g_1 = s_0 + K - 1 \tag{3.39}
$$

$$
g_2 = r_0 - \sum_{k=1}^{K-1}\mathbb{E}_q[\ln(1-v_k)] \tag{3.40}
$$

The expectations required to compute equation 3.25 are defined below, with $\psi$ the derivative of the $\Gamma$ function. These expectations are well-known moments of the Beta and Gamma distributions. Since we use a truncated stick-breaking process, $\mathbb{E}[\ln(1-v_K)] = 0$. The remaining expectations $\mathbb{E}[\boldsymbol{\eta^*_k}]$ and $\mathbb{E}[-a(\boldsymbol{\eta^*_k})]$ depend on the analytical form of the posteriors detailed in Appendix B.

$$
\mathbb{E}[\ln v_k] = \psi(\alpha_k) - \psi(\alpha_k + \beta_k) \tag{3.41}
$$

$$
\mathbb{E}[\ln(1-v_k)] = \psi(\beta_k) - \psi(\alpha_k + \beta_k) \tag{3.42}
$$

$$\mathbb{E}[w] = \frac{g_1}{g_2} \tag{3.43}$$

$$\mathbb{E}[\ln w] = \psi(g_1) - \ln g_2 \tag{3.44}$$

### 3.1.7 Lower bound convergence monitoring

This section extends the derivation of the lower bound defined in equation 3.14, using the joint probability (eq. 3.17) and approximation of the posterior (eq. 3.16) previously defined. Since the lower bound is convex [Boyd & Vandenberghe, 2004], the deterministic EM-like algorithm described in Section 3.1.6 is guaranteed to converge. While the computation of the lower bound is not required to iterate over the EM algorithm, it can be used for convergence monitoring. Instead of performing a fixed number of iterations, we can use early stopping to interrupt the training when the improvement of the lower bound does not exceed a given threshold. This allows for a significant reduction of the training time. Since, this lower bound increases at each iteration, it can also be used to check for implementation errors.

$$
\begin{aligned}
\ln p(\boldsymbol{X}|\boldsymbol{\theta}) &\geq \mathbb{E}_q[\ln p(\boldsymbol{X}, \boldsymbol{z}, \boldsymbol{\eta^*}, \boldsymbol{v}, w|\boldsymbol{\theta})] - \mathbb{E}_q[\ln q(\boldsymbol{z}, \boldsymbol{\eta^*}, \boldsymbol{v}, w)] \\
&\geq \mathbb{E}_q[\ln p(\boldsymbol{X}|\boldsymbol{z}, \boldsymbol{\eta^*})] + \mathbb{E}_q[\ln p(\boldsymbol{z}|\boldsymbol{v})] + \mathbb{E}_q[\ln p(\boldsymbol{\eta^*}|\boldsymbol{\lambda})] \\
&\quad + \mathbb{E}_q[\ln p(\boldsymbol{v}|w)] + \mathbb{E}_q[\ln p(w|s_0, r_0)] - \mathbb{E}_q[\ln q_{\boldsymbol{\alpha}, \boldsymbol{\beta}}(\boldsymbol{v})] \\
&\quad - \mathbb{E}_q[\ln q_{\boldsymbol{\tau}}(\boldsymbol{\eta^*})] - \mathbb{E}_q[\ln q_r(\boldsymbol{z})] - \mathbb{E}_q[\ln q_{g_1, g_2}(w)]
\end{aligned}
\tag{3.45}
$$

Taking the expectation of the logarithm under the specified latent variables for equations 3.18 to 3.22, we obtain:

$$\mathbb{E}_q[\ln p(\boldsymbol{X}|\boldsymbol{z}, \boldsymbol{\eta^*})] = \sum_{n=1}^{N} \sum_{k=1}^{K} r_{nk} \left( \ln h(\boldsymbol{x_n}) + \mathbb{E}[\boldsymbol{\eta_k^*}]^T T(\boldsymbol{x_n}) + \mathbb{E}[-a(\boldsymbol{\eta_k^*})] \right) \tag{3.46}$$

$$\mathbb{E}_q[\ln p(\boldsymbol{z}|\boldsymbol{v})] = \sum_{n=1}^{N}\sum_{k=1}^{\infty} r_{nk}\left(\mathbb{E}_q[\ln v_k] + \sum_{i=1}^{k-1}\mathbb{E}_q[\ln(1-v_k)]\right)$$

$$= \sum_{n=1}^{N}\sum_{k=1}^{\infty}\left(\left(\sum_{i=k+1}^{\infty} r_{ni}\right)\mathbb{E}_q[\ln(1-v_k)] + r_{nk}\mathbb{E}_q[\ln v_k]\right) \qquad (3.47)$$

$$= \sum_{n=1}^{N}\sum_{k=1}^{K}\left(q(z_n > k)\mathbb{E}_q[\ln(1-v_k)] + q(z_n = k)\mathbb{E}_q[\ln v_k]\right)$$

Where $q(z_n > k)$ and $q(z_n = k)$ are defined below. Since we truncate the sum at $K$ in equation 3.47, it induces $\mathbb{E}[\ln(1-v_K)] = 0$ and $q(z_n = k) = 0$ for $k > K$.

$$q(z_n > k) = \sum_{i=k+1}^{K} r_{ni} \qquad (3.48)$$

$$q(z_n = k) = r_{nk} \qquad (3.49)$$

$$\mathbb{E}_q[\ln p(\boldsymbol{\eta^*}|\boldsymbol{\lambda})] = \sum_{k=1}^{K}\left(\ln h(\boldsymbol{\eta_k^*}) + \boldsymbol{\lambda_1}^T\mathbb{E}[\boldsymbol{\eta_k^*}] + \lambda_2\mathbb{E}[-a(\boldsymbol{\eta_k^*})] - a(\boldsymbol{\lambda})\right) \qquad (3.50)$$

$$\mathbb{E}_q[\ln p(\boldsymbol{v}|w)] = \sum_{k=1}^{K}\left((\mathbb{E}[w]-1)\mathbb{E}[\ln(1-v_k)] - \ln\Gamma(\mathbb{E}[w]) + \ln\Gamma(\mathbb{E}[w]+1)\right) \qquad (3.51)$$

$$\mathbb{E}_q[\ln p(w|s_0, r_0)] = s_0\ln r_0 - \ln\Gamma(s_0) + (s_0-1)\mathbb{E}[\ln w] - r_0\mathbb{E}[w] \qquad (3.52)$$

Similarly, the following expectations are based on equations 3.26, 3.30, 3.34, and 3.38.

$$\mathbb{E}_q[\ln q_r(\boldsymbol{z})] = \sum_{n=1}^{N}\sum_{k=1}^{K} r_{nk}\ln r_{nk} \qquad (3.53)$$

$$\mathbb{E}_q[\ln q_{\boldsymbol{\alpha},\boldsymbol{\beta}}(\boldsymbol{v})] = \sum_{k=1}^{K} \Big( (\alpha_k - 1)\mathbb{E}[\ln(v_k)] + (\beta_k - 1)\mathbb{E}[\ln(1 - v_k)] \\ - \ln\Gamma(\alpha_k) - \ln\Gamma(\beta_k) + \ln\Gamma(\alpha_k + \beta_k) \Big) \tag{3.54}$$

$$\mathbb{E}_q[\ln q_{\boldsymbol{\tau}}(\boldsymbol{\eta^*})] = \sum_{k=1}^{K} \big( \ln h(\boldsymbol{\eta_k^*}) + \boldsymbol{\tau_{k1}}^T\mathbb{E}[\boldsymbol{\eta_k^*}] + \tau_{k2}\mathbb{E}[-a(\boldsymbol{\eta_k^*})] - a(\boldsymbol{\tau_k}) \big) \tag{3.55}$$

$$\mathbb{E}_q[\ln q_{g_1,g_2}(w)] = g_1\ln g_2 - \ln\Gamma(g_1) + (g_1 - 1)\mathbb{E}[\ln w] - g_2\mathbb{E}[w] \tag{3.56}$$

### 3.1.8 PREDICTIVE DENSITY

In order to approximate the predictive density $p(\boldsymbol{x}_{N+1}|\boldsymbol{X},\boldsymbol{\theta})$ for a new data point $\boldsymbol{x}_{N+1}$, we integrate out the posterior over the model parameters. As a result, we replace the posterior over $p$ by the posterior over $q$, which yields

$$p(\boldsymbol{x}_{N+1}|\boldsymbol{X},\boldsymbol{\theta}) = \int \sum_{k=1}^{\infty} \pi_k(\boldsymbol{v})p(\boldsymbol{x}_{N+1}|\boldsymbol{\eta_k^*})dp(\boldsymbol{v},\boldsymbol{\eta^*}|\boldsymbol{X},\boldsymbol{\theta}) \\ \approx \sum_{k=1}^{K} \mathbb{E}_q[\pi_k(\boldsymbol{v})]\mathbb{E}_q[p(\boldsymbol{x}_{N+1}|\boldsymbol{\eta_k^*})]. \tag{3.57}$$

Based on $\pi_k(\boldsymbol{v})$ and $v_k$ from equations 3.2 and 3.30, we have

$$\mathbb{E}_q[\pi_k(\boldsymbol{v})] = \frac{\alpha_k}{\alpha_k + \beta_k}\prod_{i=1}^{k-1}\left(1 - \frac{\alpha_i}{\alpha_i + \beta_i}\right) \tag{3.58}$$

We use Monte Carlo sampling to approximate the density. We draw $m$ samples $\boldsymbol{\eta_k^*}$ from the approximated posterior $q_{\boldsymbol{\tau}}^*(\boldsymbol{\eta^*})$, each sample allowing us to compute the corresponding likelihood $p(\boldsymbol{x}_{N+1}|\boldsymbol{\eta_k^*})$. The estimated likelihood for each component is obtained by averaging the resulting $m$ likelihoods.

For a multivariate Normal likelihood and a Normal-Wishart approximation of the posterior, the exact density is a mixture of Student's t-distributions St [Bishop, 2006] reported in equation 3.59, where $d$ is the dimensionality of the input data. The parameters of the Normal-Wishart distribution $\boldsymbol{\mu}_k$, $\lambda_k$, $\boldsymbol{V}_k$ and $\upsilon_k$ are obtained from the inverse parameter mapping of $\boldsymbol{\tau}_k$.

$$p(\boldsymbol{x}_{N+1}|\boldsymbol{X},\boldsymbol{\theta}) = \sum_{k=1}^{K}\left(\frac{\alpha_k}{\alpha_k+\beta_k}\prod_{i=1}^{k-1}\left(1-\frac{\alpha_i}{\alpha_i+\beta_i}\right)\mathrm{St}\left(\boldsymbol{x}_{N+1}|\boldsymbol{\mu}_k,\boldsymbol{L}_k,\upsilon_k+1-d\right)\right) \quad (3.59)$$

With

$$\boldsymbol{L}_k = \frac{(\upsilon_k+1-d)\lambda_k}{1+\lambda_k}\boldsymbol{V}_k \quad (3.60)$$

As explained above, the Student's t-distribution is then approximated by sampling $m$ parameters $\boldsymbol{\mu}$ and $\boldsymbol{\Sigma}$ for the multivariate Normal likelihood from the Normal-Wishart posterior, then averaging the resulting likelihoods.

## 3.2 EXPERIMENTAL EVALUATION

We evaluate the performance of the novelty detection algorithms based on two metrics computed on the labelled test sets which are part of the datasets described in section 3.2.1. For this purpose, we use the receiver operating characteristic (ROC) and the precision-recall (PR) metric. The comparison is based on the area under the curve (AUC) of both metrics, respectively the ROC AUC and the average precision (AP).

### 3.2.1 DATASETS

Our evaluation uses 15 datasets ranging from 723 to 20,000 samples and containing from 6 to 107 features. Of those datasets, 12 are publicly available on the UCI [Asuncion & Newman, 2007] or OpenML [Vanschoren et al., 2014] repositories while the 3 remaining datasets are novel proprietary datasets containing production data from the company Amadeus. Table 3.3 gives an overview of the datasets characteristics. Our study assesses if the models are able to generalize to future datasets, which is a novel approach in outlier detection works. This requires that algorithms support unseen testing data, and is achieved by performing a Monte Carlo cross validation of 5 iterations, using 80% of the data for the training phase and 20% for the prediction. Training and testing datasets contain the same proportion of outliers, and ROC AUC and AP are measured based on the predictions made. For 7 of the publicly available datasets, the outlier classes are selected according to the recommendations made in [Emmott et al., 2016], which are based on extensive datasets comparisons. However, the cited experiment discards all categorical data, while we keep those features and performed one-hot encoding to binarize them, keeping all information from the dataset at the cost of a higher di-mensionality. Normalization is further achieved by centering numerical features to the mean and scaling them to unit variance.

Table 3.3: UCI, OpenML and proprietary datasets benchmarked - (# categ. dims) is the average number of binarized features obtained after transformation of the categoricals.

| Dataset | Nominal class | Anomaly class | Numeric dims | Categ. dims | Samples | Anomalies |
|---|---|---|---|---|---|---|
| ABALONE | 8, 9, 10 | 3, 21 | 7 | 1 (3) | 1,920 | 29 (1.51%) |
| ANN-THYROID | 3 | 1 | 21 | 0 (0) | 3,251 | 73 (2.25%) |
| CAR | unacc, acc, good | vgood | 0 | 6 (21) | 1,728 | 65 (3.76%) |
| COVTYPE | 2 | 4 | 54 | 0 (0) | 10,000[1] | 95 (0.95%) |
| GERMAN-SUB | 1 | 2 | 7 | 13 (54) | 723 | 23 (3.18%)[2] |
| KDD-SUB | normal | u2r, probe | 34 | 7 (42) | 10,000[1] | 385 (3.85%) |
| MAGIC-GAMMA-SUB | g | h | 10 | 0 (0) | 12,332 | 408 (3.20%)[2] |
| MAMMOGRAPHY | -1 | 1 | 6 | 0 (0) | 11,183 | 260 (2.32%) |
| MUSHROOM-SUB | e | p | 0 | 22 (107) | 4,368 | 139 (3.20%)[2] |
| SHUTTLE | 1 | 2, 3, 5, 6, 7 | 9 | 0 (0) | 12,345 | 867 (7.02%) |
| WINE-QUALITY | 4, 5, 6, 7, 8 | 3, 9 | 11 | 0 (0) | 4,898 | 25 (0.51%) |
| YEAST [3] | CYT, NUC, MIT | ERL, POX, VAC | 8 | 0 (0) | 1,191 | 55 (4.62%) |
| PNR | 0 | 1, 2, 3, 4, 5 | 82 | 0 (0) | 20,000 | 121 (0.61%) |
| SHARED-ACCESS | 0 | 1 | 49 | 0 (0) | 18,722 | 37 (0.20%) |
| TRANSACTIONS | 0 | 1 | 41 | 1 (9) | 10,000[1] | 21 (0.21%) |

[1] Subsets of the original datasets are used, with the same proportion of outliers.

[2] Anomalies are sampled from the corresponding class, using the average percentage of outliers depicted in [Emmott et al., 2016].

[3] The first feature corresponding to the protein name was discarded.

The three following datasets contain production data collected by Amadeus, a Global Distribution System (GDS) providing online platforms to connect the travel industry. This company manages almost half of the flight bookings worldwide and is targeted by fraud attempts leading to revenue losses and indemnifications. The datasets do not contain information traceable to any specific individuals.

The PASSENGER NAME RECORDS (PNR) dataset contains booking records, mostly flight and train bookings, containing 5 types of frauds labelled by fraud experts. The features in this dataset describe the most important changes applied to a booking from its creation to its deletion. Features include time-based information, e.g. age of a PNR, percentage of cancelled flight segments or passengers, and means and standard deviations of counters, e.g. number of passenger modifications, frequent traveller cards, special service requests (additional luggage, special seat or meal), or forms of payment.

The TRANSACTIONS dataset is extracted from a Web application used to perform bookings. It focuses on user sessions which are compared to identify bots and malicious users. Examples of features are the number of distinct IPs and organizational offices used by a user, the session duration and means and standard deviations applied to the number of bookings and number of actions. The most critical actions are also monitored.

The SHARED-ACCESS dataset was generated by a backend application used to manage shared rights between different entities. It enables an entity to grant specific reading (e.g. booking retrieval, seat map display) or writing (e.g. cruise distribution) rights to another entity. Monitoring the actions made with this application should prevent data leaks and sensible right transfers. For each user account, features include the average number of actions performed per session and time unit, the average and standard deviation for some critical actions per session, and the targeted rights modified.

### 3.2.2 Datasets for scalability tests

As the choice of an outlier detection algorithm may not only be limited to its accuracy but is often subject to computational constraints, our experiment includes training time, prediction time, memory usage and noise resistance (through precision-recall measurements) of each algorithm on synthetic datasets.

For these scalability tests, we generate synthetic datasets of different sizes containing a fixed proportion of background noise. The datasets range from 10 samples to 10 million samples and from 2 to 10,000 numerical features. We also keep the number of features and samples fixed while increasing the proportion of background noise from 0% to 90% to perform robustness measurements. The experiment is repeated 5 times, using the same dataset for training and testing. We allow up to 24 hours for training or prediction steps to be completed and stop the computation after this period of time. Experiments reaching a timeout or requiring more than 256 GB RAM do not report memory usage nor robustness in section 3.3.

In order to avoid advantaging some algorithms over others, the datasets are generated using two Student's T distributions. The distributions are respectively centered in $(0, 0..., 0)$ and $(5, 5..., 5)$, while the covariance matrices are computed using $c_{ij} = \rho^{|i-j|}$ where $c_{ij}$ is entry $(i, j)$ of the matrix. The parameter $\rho$ follows a uniform distribution $\rho \sim U(0, 1)$ and the degrees of freedom parameter follows a Gamma distribution $v \sim \Gamma(1, 5)$. We then add outliers uniformly sampled from a hypercube 7 times bigger than the standard deviation of the nominal data.

### 3.2.3 Algorithms implementations and parameters

Most implementations used in this experiment are publicly available. Table 3.4 details the programming languages and initialization parameters selected. A majority of methods have flexible parameters and perform very well without an extensive tuning. The **Matlab Engine AP for Python** and the **rpy2** library allow us to call Matlab and R code from Python.

DPGMM is our own implementation of a Dirichlet Process Mixture Model and follows the guidelines given in [Blei & Jordan, 2006] where we place a Gamma prior on the scaling parameter $\beta$. Making our own implementation of this algorithm is motivated by its capability of handling a wide range of probability distributions, including categorical distributions. We thus benchmark DPGMM, which uses only Gaussian distributions to model data and thus uses continuous and binarized features as all other algorithms, and DPMM which uses a mixture of multivariate Gaussian / Categorical distributions, hence requiring fewer data transformations and working on a smaller number of features. This algorithm is the only one capable of using the raw string features from the datasets.

Note that DPGMM and DPMM converge to the same results when applied to non-categorical data, and that our DPGMM performs similarly to the corresponding scikit-learn implementation called **Bayesian Gaussian Mixture (BGM)**. However, we did not optimize our implementation which uses a more general exponential-family representation for probability distributions. This greatly increases the computational cost and results in a much higher training and prediction time.

The Grow When Required (GWR) network has a nonparametric topology. Identifying outlying neurons as described in [Munoz & Muruzábal, 1998] to detect outliers from a 2D grid topology may thus not be applicable to the present algorithm. The node connectivity can indeed differ significantly from one node to another, and we need a ranking of the outliers for our performance measurements more than a two-parameter binary classification. Therefore, the score assigned to each observation is here the squared distance between an observation and the closest node in the network. Note that regions of outliers sufficiently dense to attract neurons may not be detected with this technique.

Table 3.4: Implementations and parameters selected for the evaluation

| Algorithm | Language | Parameters |
|---|---|---|
| GMM [1] | Python | $components = 1$ |
| DPGMM DPMM | Python | $\boldsymbol{\Gamma_\theta} = (\alpha = 1, \beta = 0)), k_{max} = 10, \boldsymbol{\mu_\theta} = mean(data), \boldsymbol{\Sigma_\theta} = var(data)$ |
| RKDE [2,3] | Matlab | $bandwidth = LKCV, loss = Huber$ |
| PPCA | Python | $components = mle, svd = full$ |
| LSA | Python | $\sigma = 1.7, \rho = 100$ |
| MAHA | Python | n/a |
| LOF | Python | $k = \max(n * 0.1, 50)$ |
| ABOD [2] | R | $k = \max(n * 0.01, 50)$ |
| SOD [2] | R | $k = \max(n * 0.05, 50), k_{shared} = \frac{k}{2}$ |
| KL [12] | R | $components = 1$ |
| GWR [2] | Matlab | $it = 15, t_{hab} = 0.1, t_{insert} = 0.7$ |
| OCSVM | Python | $\nu = 0.5$ |
| IFOREST | Python | $contamination = 0.5$ |

[1]  Parameter tuning is required to maximize the mean average precision (MAP).
[2]  We extend these algorithms to add support for predictions on unseen data points.
[3]  We use Scott's rule-of-thumb $h = n^{-1/(d+4)}$ [Scott, 1992] to estimate the bandwidth when it cannot be computed due to a high number of features.

## 3.3  RESULTS

For each dataset, the methods described in section 2.1 are applied to the 5 training and testing subsets sampled by Monte Carlo cross validation. We report here the average and standard deviation over the runs. The programming language and optimizations applied to the implementations may affect the training time, prediction time and memory usage measured in sections 3.3.3 and 3.3.4. For this reason, our analysis focuses more on the curves slope and the algorithms complexity than on the measured values. The experiments are performed on a VMware virtual platform running Ubuntu 14.04 LTS and powered by an Intel Xeon E5-4627 v4 CPU (10 cores at 2.6GHz) and 256GB RAM. We use the Intel distribution of Python 3.5.2, R 3.3.2 and Matlab R2016b.

### 3.3.1 ROC AND PRECISION-RECALL

The area under the ROC curve is estimated using trapezoidal rule while the area under the precision-recall curve is computed by average precision (AP). Figure 3.7 shows the mean and standard deviation AP per algorithm and dataset while figure 3.8 reports the ROC AUC. For the clarity of presentation, figure 3.6 shows the global average and standard deviation average of both metrics, sorting algorithms by decreasing mean average precision (MAP).

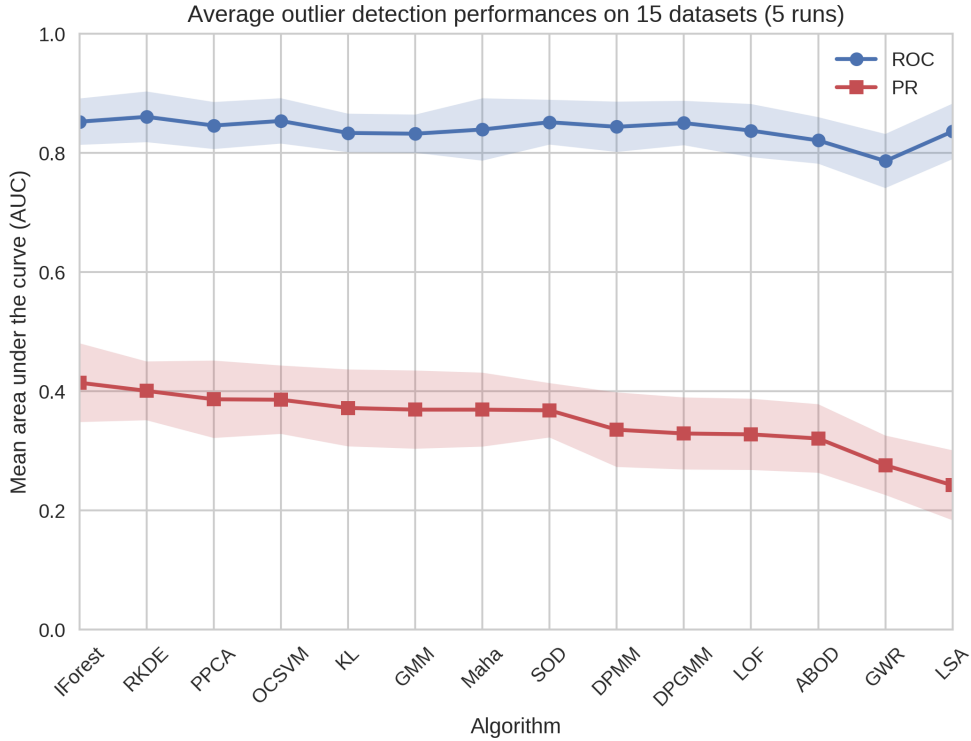

Figure 3.6: Mean area under the ROC and PR curve per algorithm (descending PR). For both metrics, the higher values, the better the results.

Since averaging these metrics may induce a bias in the final ranking caused by extreme values on some specific datasets (e.g. AP of IFOREST on KDD-SUB), we also rank the algorithms per dataset and aggregate the ranking lists without considering the measured values. The Cross-Entropy Monte Carlo algorithm and the Spearman distance are used for the aggregation [Pihur et al., 2009]. The resulted rankings presented in Table 3.5 are similar to the rankings given in figure 3.6 and confirm the previous trend observed. The rest of this paper will thus refer to the rankings introduced in figure 3.6.

Table 3.5: Rank aggregation through Cross-Entropy Monte Carlo

| Algorithm | GMM | DPGMM | DPMM | RKDE | PPCA | LSA | MAHA | LOF | ABOD | SOD | KL | GWR | OCSVM | IFOREST |
|---|---|---|---|---|---|---|---|---|---|---|---|---|---|---|
| **PR AUC** | 6 | 12 | 9 | 2 | 5 | 14 | 7 | 11 | 10 | 4 | 8 | 13 | 3 | 1 |
| **ROC AUC** | 11 | 4 | 5 | 1 | 7 | 9 | 6 | 10 | 13 | 8 | 12 | 14 | 3 | 2 |

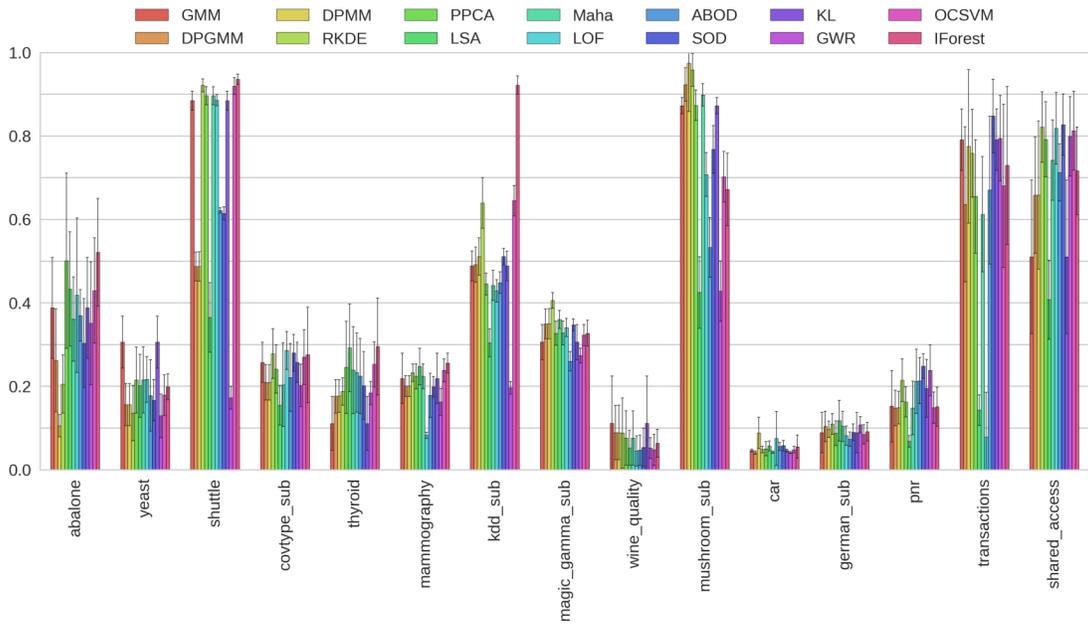

Figure 3.7: Mean and std area under the precision-recall curve per dataset and algorithm (5 runs)

Note that we are dealing with heavily imbalanced class distributions where the anomaly class is the positive class. For this kind of problems where the positive class is more interesting than the negative class though underweighted due to the high number of negative samples, precision-recall curves show to be particularly useful. Indeed, the precision metric strongly penalizes false positives, even if they only represent a small proportion of the negative class, while false positives have very little impact on the ROC [Davis & Goadrich, 2006]. The area under the ROC curve is thus reported in our experiments for the sake of completeness, but we will focus on the average precision which is better suited to novelty detection.

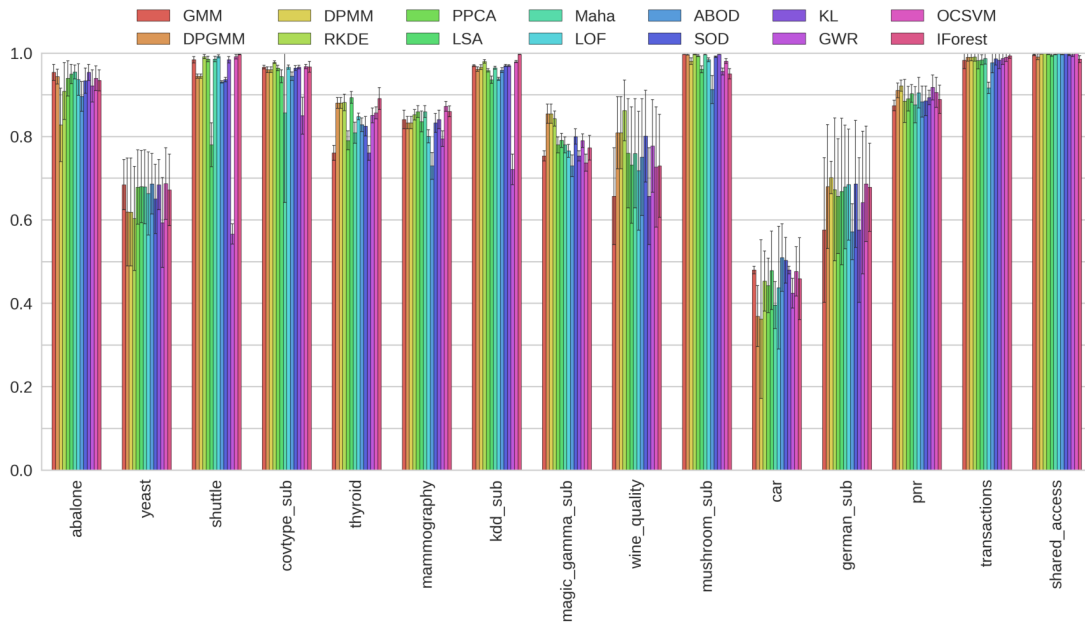

Figure 3.8: Mean and std area under the ʀᴏᴄ curve per dataset and algorithm (5 runs)

Looking at the mean average precision, ɪꜰᴏʀᴇsᴛ, ʀᴋᴅᴇ, ᴘᴘᴄᴀ and ᴏᴄsᴠᴍ show excellent performance and achieve the best outlier detection results of our benchmark. Outperforming all other algorithms on several datasets (e.g. ᴋᴅᴅ-sᴜʙ, ᴀʙᴀʟᴏɴᴇ or ᴛʜʏʀᴏɪᴅ), ɪꜰᴏʀᴇsᴛ shows good average performance which makes it a reliable choice for outlier detection. ʀᴋᴅᴇ comes in second position and also shows excellent performance on most datasets, especially when applied to high-dimensionality problems.

One-class SVM achieves good performance without requiring significant tuning. We note that the algorithm perform best on datasets containing a small proportion of outliers, which seems to confirm that the method is well-suited to novelty detection.

The parametric mixture model ɢᴍᴍ uses only a single Gaussian, and still manages to reach excellent performance. While more complex nonparametric methods may be prone to overfitting, the use of such a constrained model prevents it. We investigate these results in more details at the end of this section, where we provide an alternative ranking of the methods based on a selection of datasets.

The good ranking of ᴘᴘᴄᴀ, ᴋʟ and the Mahalanobis distance is also explained by precision-recall measurements very similar to the ones of ɢᴍᴍ. Probabilistic ᴘᴄᴀ is indeed regarded as a GMM with one component, KL is a GMM with a different scoring function, while the model of the Mahalanobis distance is closely related to the multivariate Gaussian distribution. If these simple models perform well on average, they are not suitable for more complex datasets, e.g. the proprietary datasets from Amadeus, where nonparametric methods able to handle clusters of arbitrary shape such as RKDE, SOD or even GWR prevail.

We indeed observe that SOD outperforms the other methods on the Amadeus datasets, and more generally performs very well on large datasets composed of more than 10,000 samples. This method is the best neighbor-based algorithm of our benchmark, but requires a sufficient number of features to infer suitable subspaces through feature selection.

DPMM performs better than DPGMM. As the two methods should converge identically when applied to numerical data, it is the way they handle categorical features that explains this difference. While DPGMM is outperformed by several methods when applied to numerical datasets, DPMM shows good performance on categorical and mixed-type data. Looking at the detailed average precisions highlighted in Figure 3.7 for datasets containing categorical features, we notice that DPMM outperforms DPGMM on four datasets (KDD-SUB, MUSHROOM-SUB, CAR and TRANSACTIONS), while it is outperformed on two others (ABALONE and GERMAN-SUB). This gain of performance for DPGMM is likely to be caused by categorical features strongly correlated to the true class distribution and having a high number of distinct values, resulting in several binarized features and thus a higher weight for the corresponding categorical in the final class prediction. However, when the class distribution is heavily unbalanced, the Chi-Square test based on contingency tables we performed did not allow us to confirm this hypothesis.

The results of DPMM are reached in a smaller training and prediction time due to a smaller dimensionality of the non-binarized input data. Yet, we believe that DPGMM would reach a smaller computation time if making all its computations for Normal-Wishart distributions instead of exponential-family representations. This was confirmed by measuring the computation time of the BGM implementation in *scikit-learn* on the same datasets.

LOF and ABOD do not stand out, with unexpected drop of performance observed for LOF on TRANSACTIONS and MAMMOGRAPHY that cannot be explained solely based on the dataset characteristics. Using a number of neighbors sufficiently high is important when dealing with large datasets containing a higher number of outliers. Increasing the sample size for ABOD may lead to slightly better performance at the cost of a much higher computation time, for a method which is already slow. We also benchmarked the k-nearest-neighbors approach described in the original paper which showed reduced computation time for $k = 15$ though this did not improve performance. Despite the use of angles instead of distances, this algorithm performs worse than LOF on 4 datasets among the 7 datasets containing more than 40 features. It is however one of the best outlier detection methods on the PNR and TRANSACTIONS datasets.

Although GWR achieves lower performance than other methods in our benchmark, in the case of a low proportion of anomalies, e.g. with PNR, SHARED-ACCESS and TRANSACTIONS, the algorithm reaches excellent precision-recall scores as the density of outliers is not sufficient to attract any neuron. SOM can thus be useful for novelty detection targeting datasets free of outliers, or when combined with a manual analysis of the quantization errors and MID matrix as described in [Munoz & Muruzábal, 1998]. Similarly to GWR, LSA reaches low average performance, especially for large or high-dimensional datasets, but it achieves good results for small datasets.

We have previously observed simple novelty detection methods, e.g. GMM, outperforming nonparametric alternatives such as DPMM. While GMM used a single multivariate Normal to reach such performance, DPMM and DPGMM often used between 5 and 10 components to model these complex datasets of variable density and shape. The number of components for parametric models was selected to maximize the MAP. The careful reader will notice that several novelty detection datasets described in Table 3.3 and recommended in [Emmott et al., 2016] are based on classification data. For these datasets, anomalies are not sparse background noise, but are sampled from one or more classes from the original classification task. If such classes were to form dense clusters of outliers, nonparametric methods could provide a more accurate density model capturing these clouds of outliers, remnants of former classes, while receiving a lower average precision. This is confirmed by the results observed for most one-class datasets contaminated by noisy anomalies, where DPMM and DPGMM outperform GMM, e.g. MAGIC-GAMMA-SUB, MUSHROOM-SUB and GERMAN-SUB.

In order to demonstrate our theory, we showcase the overall performance of nonparametric methods in Figure 3.9 and in Table 3.6, exluding the 5 datasets which were generated from classification data. While IFOREST, RKDE and OCSVM remain in good position, DPMM and SOD now obtain a much higher ranking. This emphasizes the need to work with a large number of established datasets of known characteristics when evaluating the performance of a given method.

In addition, the scores and ranking of IFOREST and GMM are much higher for the datasets selected in [Emmott et al., 2016] than for the other datasets where nonparametric methods prevail. Since the selection process described in their study makes use of the performance of several outlier detection methods to perform their selection, this process may benefit algorithms having a behavior similar to the chosen methods.

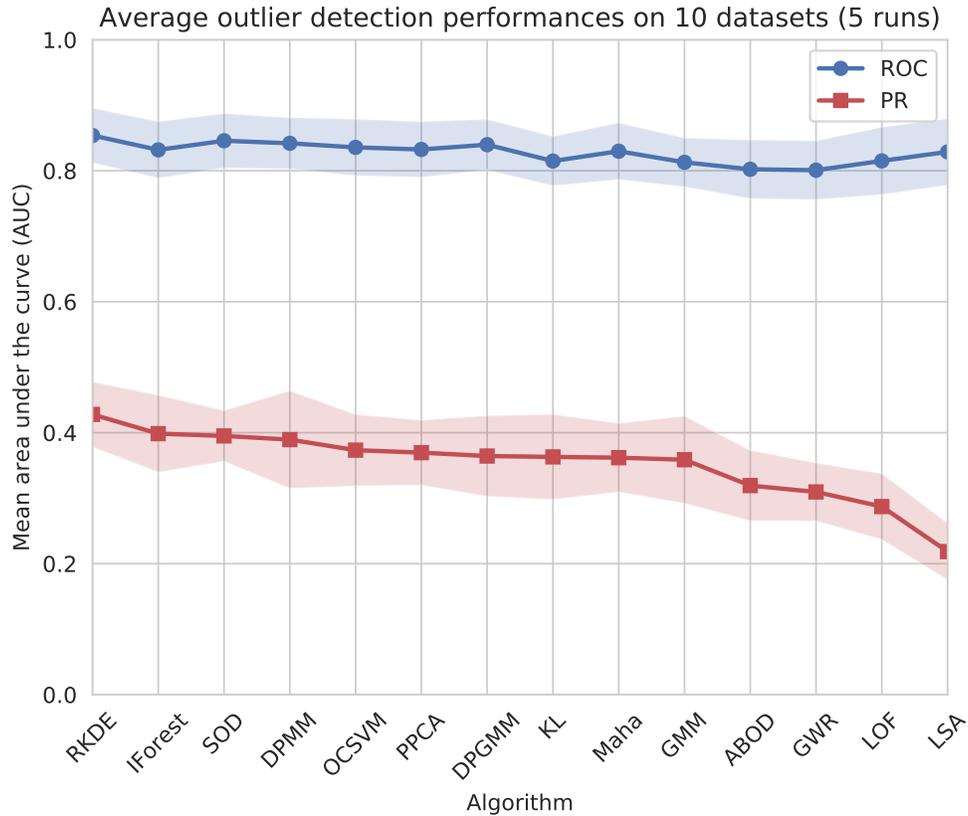

Figure 3.9: Mean area under the ROC and PR curve per algorithm (descending PR). For both metrics, the higher values, the better the results. Anomaly detection datasets generated from classification datasets have been removed.

### 3.3.2 ROBUSTNESS

As our synthetic datasets contain only numerical data and since DPMM and DPGMM are the same method when the dataset does not contain categorical features, we exclude DPMM from our results and add BGM, the *scikit-learn* optimized equivalent of our DPGMM implementation. We expect the algorithms to exhibit a similar scalability when applied to one-hot encoded data, with the exception of DPMM which consumes plain categorical features before encoding.

Table 3.6: Rank aggregation through Cross-Entropy Monte Carlo. Anomaly detection datasets generated from classification datasets have been removed.

| Algorithm | GMM | DPGMM | DPMM | RKDE | PPCA | LSA | MAHA | LOF | ABOD | SOD | KL | GWR | OCSVM | IFOREST |
|-----------|-----|-------|------|------|------|-----|------|-----|------|-----|----|----|-------|---------|
| **PR AUC** | 5 | 4 | 3 | 2 | 6 | 14 | 9 | 12 | 11 | 1 | 8 | 13 | 10 | 7 |
| **ROC AUC** | 13 | 5 | 2 | 1 | 6 | 9 | 7 | 10 | 14 | 4 | 12 | 11 | 3 | 8 |

For additional scalability insights on categorical data, the reader should thus refer to Sections 3.3.3 and 3.3.4, using the number of categorical features for DPMM and the number of one-hot encoded features for other methods.

Figures 3.10, 3.11 and 3.12 measure the area under the precision-recall curve, the positive class being the background noise for the two first figures, and the nominal samples generated by the mixture of Student's T distributions in the last figure.

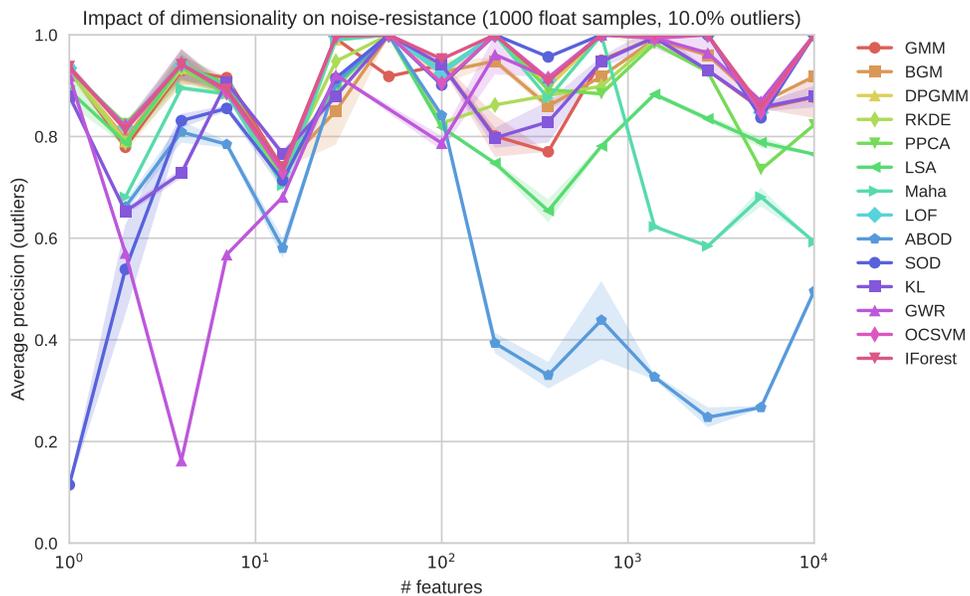

Figure 3.10: Robustness for increasing number of features

The resistance of each algorithm to the curse of dimensionality is highlighted in figure 3.10, where we keep a fixed level of background noise while increasing the dataset dimensionality. The results show good performance in average and unexpectedly good results for GWR for more than 10 features. Surprisingly, ABOD which is supposed to efficiently handle high dimensionality performs poorly here. Similarly, LSA and Mahalanobis do not perform well. The difference of results between BGM and DPGMM is likely due to a different cluster responsibility initialization, as BGM uses a K-means to assign data points to clusters centroids scattered among dense regions of the dataset.

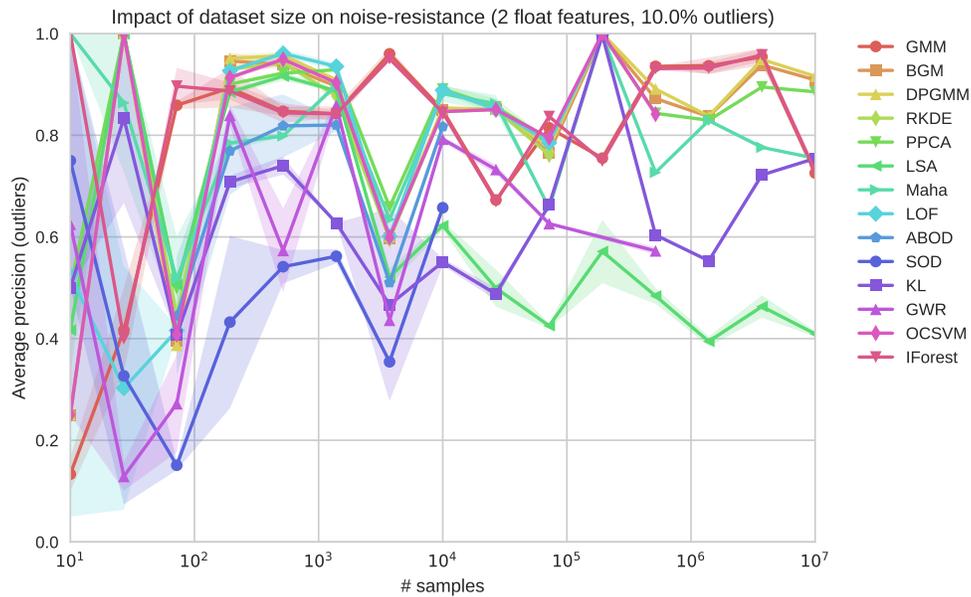

Figure 3.11: Robustness for increasing number of samples

Figure 3.11 shows decreasing outlier detection performance for LSA and GWR while increasing the number of observations in the dataset. SOD and KL do not perform well either, though their results are correlated with the variations of better methods. We thus assume that the current experiment is not well-suited for these method, as SOD applies feature selection and has demonstrated better performance in higher dimensionalities. Increasing the number of samples resulted in an overall increasing precision. The results given for less than 100 data points show a high entropy as the corresponding datasets contain very sparse data in which dense regions cannot be easily identified.

Increasing the proportion of background noise in figure 3.12 shows a lack of robustness for LSA, SOD, OCSVM, GWR and ABOD. While the two first methods are very sensible to background noise, the three others maintain good performance up to 50% of outliers. Neighbor-based method can only cope with a restricted amount of noise, though increasing the number of samples used to compute the scores could lead to better results. Similarly, most neurons of GWR were attracted by surrounding outliers. In order to avoid penalizing OCSVM, this specific evaluation uses an adaptive $\nu$ which increases with the proportion of outliers with a minimum value of 0.5. In spite of this measure, OCSVM also shows very poor results above 50% noise. Mahalanobis, KL, LOF and GMM do not perform well either in noisy environments, despite the use of 2 components by GMM and KL. For this experimental setting, the best candidates in a dataset highly contaminated by sparse outliers are IFOREST, DPGMM, BGM, RKDE and PPCA.

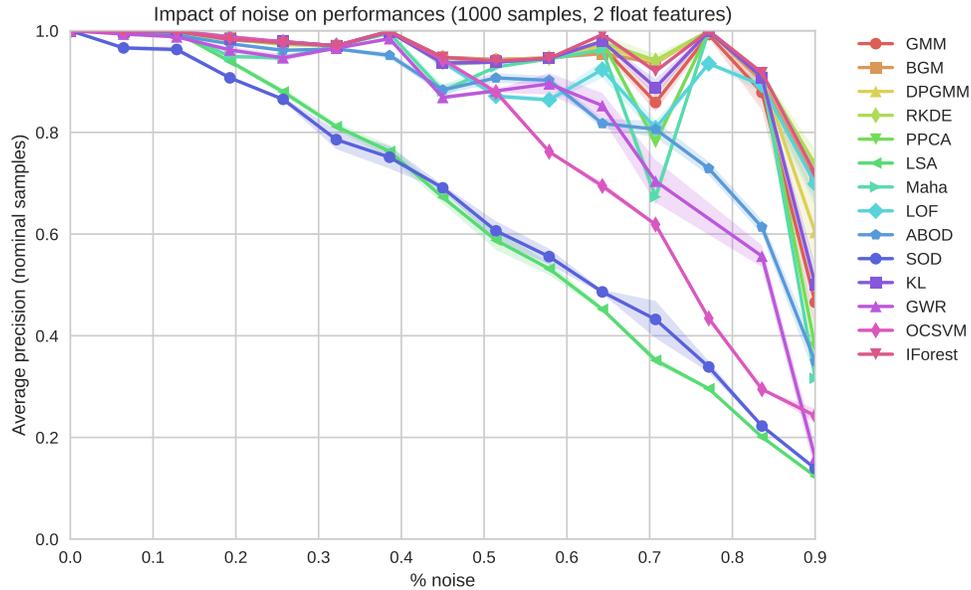

Figure 3.12: Robustness for increasing noise density

To conclude, the robustness measures on synthetic datasets confirm the poor performance of ABOD and GWR showed in our previous ranking. Good average results were observed for IFOREST, OCSVM, LOF, RKDE, DPGMM and GMM. The nearest-neighbor-based methods showed difficulties in handling datasets with a high background noise.

### 3.3.3 Complexity

We now focus on the computation and prediction time required by the different methods when increasing the dataset size and dimensionality. The running environment and the amount of optimizations applied to the implementations strongly impacts those measures. For this reason, we focus on the curves' evolutions more than on the actual value recorded. Comparing the measurements of BGM and DPGMM that implement the same algorithm is a good illustration of this statement.

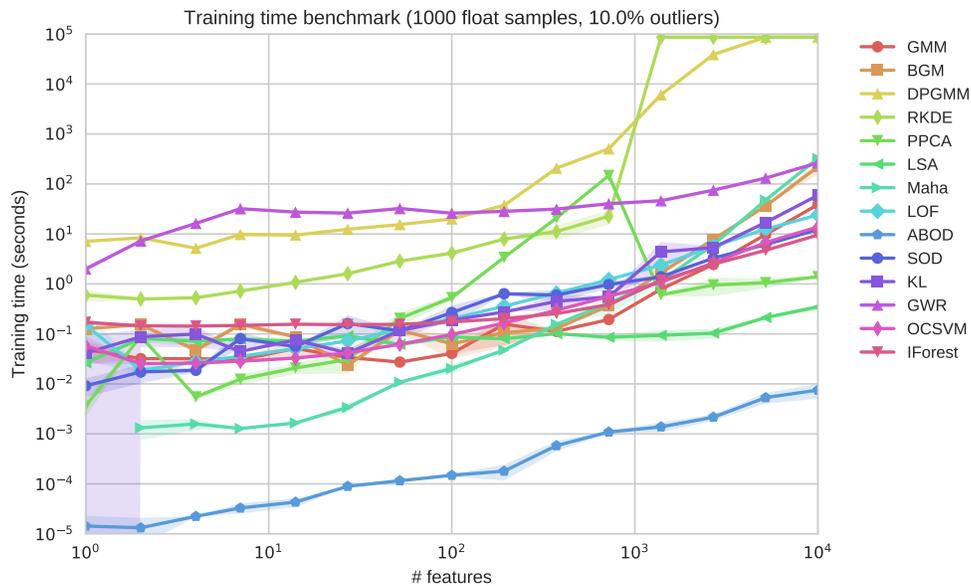

Figure 3.13: Training time for increasing number of features

Increasing the number of features in figures 3.13 and 3.14 shows an excellent scalability for LSA, GWR and IFOREST with a stable training time and a good prediction time evolution. DPGMM has here the worst training and prediction scaling, reaching the 24 hours timeout for more than 3,000 features. This scaling is confirmed by the increases observed for BGM and GMM. RKDE performs also poorly with a timeout caused by a high bandwidth when the number of features becomes higher than the number of samples. High dimensionality datasets do not strongly affect distance-based and neighbor-based methods, though probabilistic algorithms such as GMM, DPGMM, RKDE or PPCA suffer from the increasing number of dimensions. The use of computationally expensive matrix operations whose complexity depend on the data dimensionality, e.g. matrix factorizations and multiplications, is a major cause of the poor scalability observed. Maximum likelihood estimation fails to estimate the suitable number of components for ᴘᴘᴄᴀ for more than 1,000 features. We keep enough components to explain at least 90% of the variance, which explains the decrease of training time.

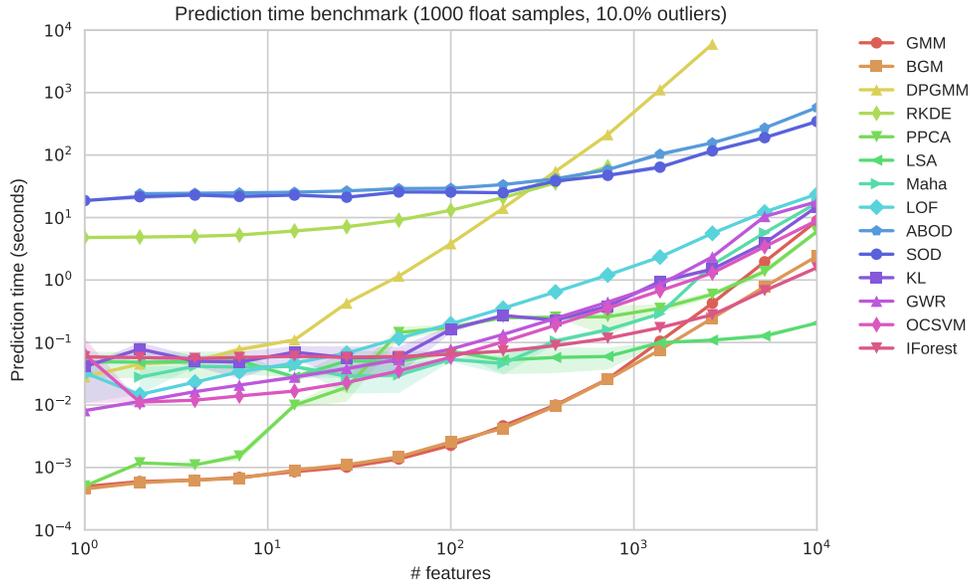

Figure 3.14: Prediction time for increasing number of features

The number of samples has a strong impact on the training and prediction time of ʀᴋᴅᴇ, ꜱᴏᴅ, ᴏᴄꜱᴠᴍ, ʟᴏꜰ and ᴀʙᴏᴅ which scale very poorly in figures 3.15 and 3.16. Those five algorithms would reach the timeout of 24 hours for less than one million samples, though ʀᴋᴅᴇ and ʟᴏꜰ exceed the available memory first (section 3.3.4). All the other algorithms show good and similar scalability, despite a higher base computation time for ɢᴡʀ and ᴅᴘɢᴍᴍ due to the lack of ᴋᴅᴇ optimizations. The additional exponential-family computations do not seem to impact the complexity of this algorithm. Training ᴀʙᴏᴅ consists only in making a copy of the training dataset, which explains the low training time reported. Its prediction time is however the least scalable, the true slope being observed for more than 5,000 samples.

In summary and looking at the overall measures, ɪꜰᴏʀᴇꜱᴛ and ʟꜱᴀ show a very good training and prediction time scaling for both increasing number of features and samples, along with a very small base computation time. DPGMM, GMM and BGM scale well on datasets with a large number of samples and thus could be suitable for systems where fast predictions matter. The base computation time of DPGMM is however an important issue when the number of features becomes higher than a hundred. RKDE, OCSVM and SOD which have good outlier detection performance on real datasets are thus computationally expensive, which adds interest to IFOREST, DPGMM and simpler models such as GMM, KL, PPCA or Mahalanobis.

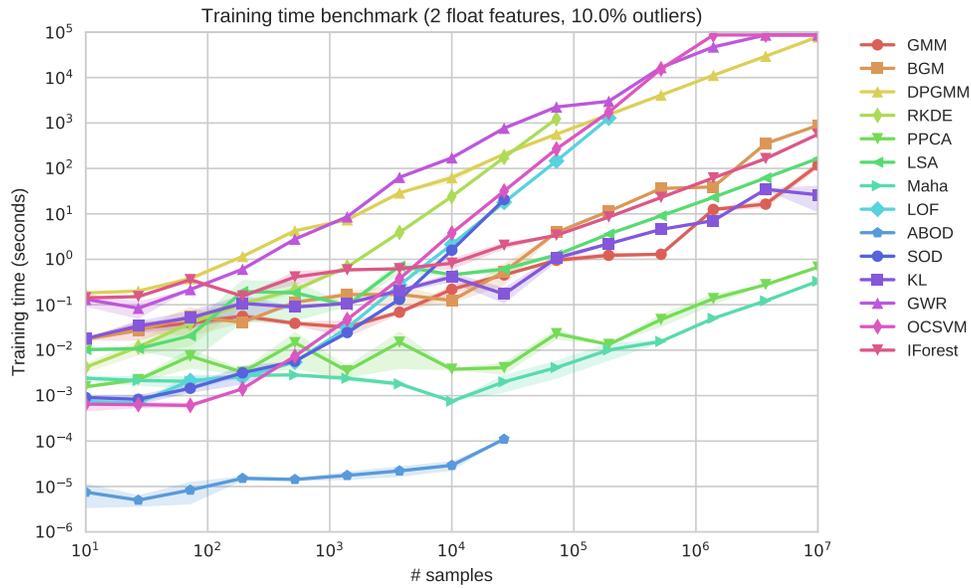

Figure 3.15: Training time for increasing number of samples

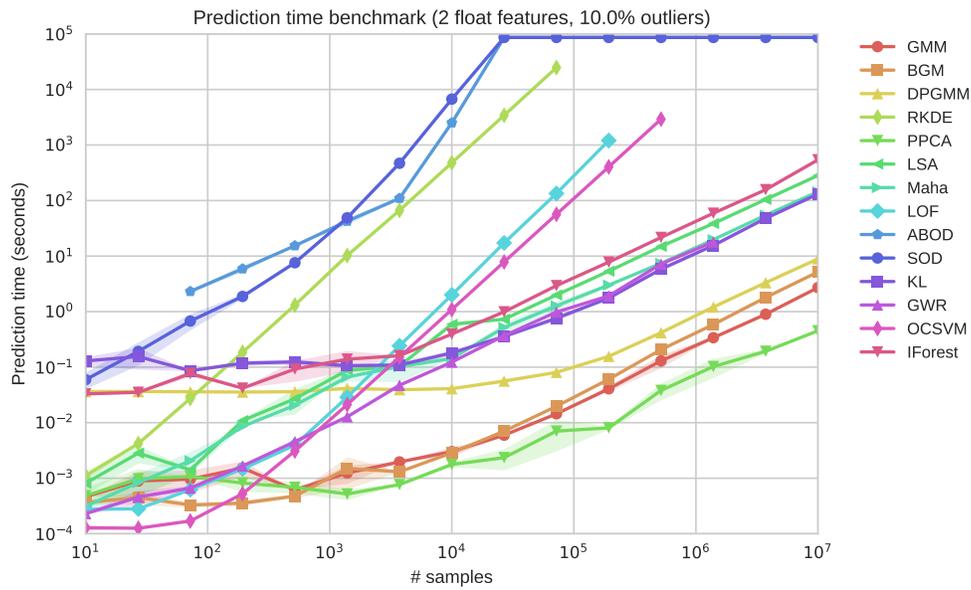

Figure 3.16: Prediction time for increasing number of samples

### 3.3.4 Memory usage

We report in figures 3.17 and 3.18 the highest amount of memory required by each algorithm when applied to our synthetic datasets during the training or prediction phase. We clear the Matlab objects and make explicit collect calls to the Python and R garbage collectors before running the algorithms. We then measure the memory used by the corresponding running process before starting the algorithm and subtract it to the memory peak observed while running it. This way, our measurements ignore the memory consumption caused by the environment and the dataset which reduces the measurement differences due the running environment, e.g. Matlab versus Python. Measures are taken at intervals of $10^{-4}$ second using the **memory_profiler** library for Python and R[*] and the UNIX *ps* command for Matlab. Small variations can be observed for measures smaller than 1 MB and are not meaningful.

As depicted in figure 3.17, most algorithms consume little memory, an amount which does not significantly increase with the number of features and should not impact the running system. IFOREST, OCSVM, LOF and LSA have a constant memory usage below 1 MB while RKDE remains near-constant. GWR, ABOD and SOD also have a good scalability. The other algorithms may require too much memory for high-dimensional problems, with Mahalanobis requiring about 4.5GB to store the mean and covariance matrices of 10,000 features. Allowing many more clusters and storing temporary data structures, DPGMM requires up to 80 GBs when applied to 2,600 features while BGM stores only 14 GBs of data for 10,000 features.

The increasing number of samples has a higher impact on the RAM consumption depicted in figure 3.18. RKDE and LOF both run out of memory before the completion of the benchmark and reach, respectively, 158 GB and 118GB memory usage for 72,000 and 193,000 samples. SOD consumes about as much memory as RKDE though reaches the timeout with fewer sam-ples. This amount of memory is mostly caused by the use of a pairwise distance matrix by these algorithms, which requires 76GB of RAM for 100,000 samples using 8 bytes per double precision distance. The other methods scale much better and do not exceed 5GB for 10 million samples, except IFOREST and LSA which allocate 60GB and 38GB RAM. For the sake of completeness, we performed the same experiment with the $k$-nearest neighbors implementation of ABOD with $k = 15$, and observed a scalability and memory usage similar to RKDE.

---

[*] *rpy2* stores R objects in the running Python process. In addition, R prevents concurrent accesses which do not allow us to use dedicated R commands to measure memory.

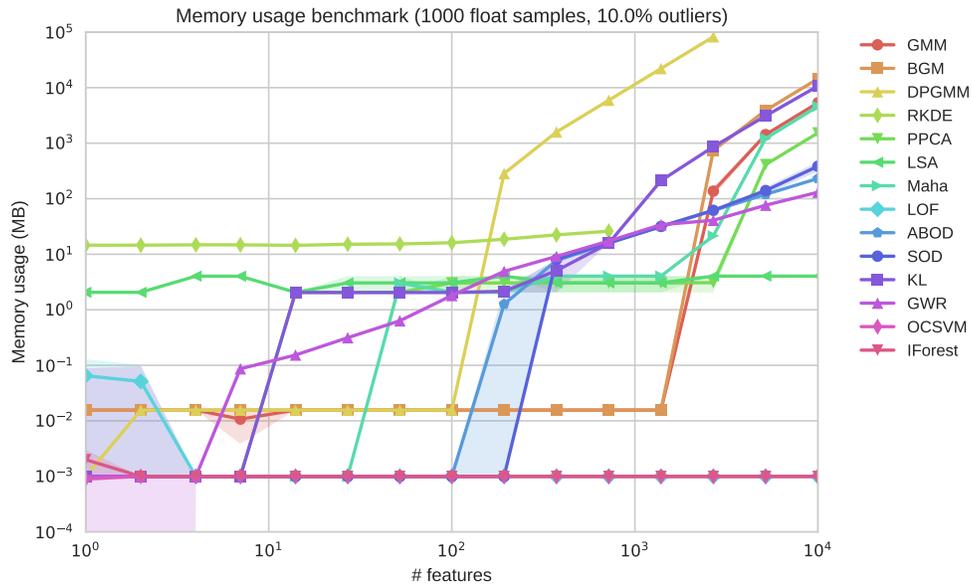

Figure 3.17: Memory usage for increasing number of features

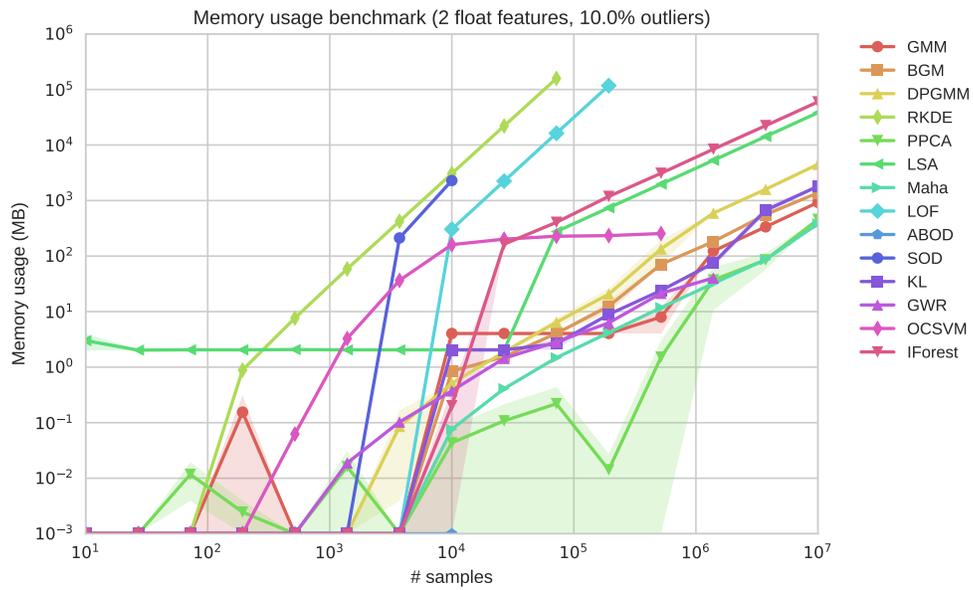

Figure 3.18: Memory usage for increasing number of samples

We have seen that several algorithms have important memory requirements which must be carefully considered depending on the available hardware. Algorithms relying on multivariate covariance matrices will be heavily impacted by the growing number of features, while methods storing a pairwise distance matrix are not suitable for a large number of samples. Our implementation of DPGMM scales as well as other mixture models though comes with a much higher memory usage on high dimensional datasets. OCSVM, GWR and ABOD have the best memory requirements and scalability and never exceed 250MB RAM, at the cost of a higher computation time since these three methods reach our 24 hours timeout.

### 3.3.5 Segmentation contours

In order to show how the choice of a given model representation affects the final density, we report in Figure 3.19 the normalized density returned by each algorithm when applied to the scaled OLD FAITHFUL dataset. Warm colors depict high anomaly scores. The density was interpolated from the predicted score of 2,500 points distributed on a 50x50 mesh grid.

The two first plots are the result of Gaussian mixture models, using 2 components for GMM and an upper bound of 10 components for DPGMM. If the plots are very similar, we denote a slightly higher density area between the two clusters for DPGMM. This difference is caused by the remaining mixture components for which the weight is close to 0 and the covariance matrix based on the entire dataset. The information theoretic algorithm based on the KL divergence predicts scores based on a GMM, which explains the similarity between the two.

The contribution of each observation to the overall density is clearly visible for RKDE which finds a density estimation tightly fitting the dataset. In contrast, the models used by PPCA and the Mahalanobis distance are much more constrained and fail at identifying the two clusters. LSA and LOF perform much better though also assign very low anomaly scores to the sparse area located between the two clusters. However, these methods should be able to handle clusters of arbitrary shape.

Table 3.7: Resistance to the curse of dimensionality, runtime scalability and memory scalability on datasets of increasing size and dimensionality. Performance on background noise detection is also reported for datasets of increasing noise proportion.

| Algorithm | Training/prediction time | | Mem. usage | | Robustness | | |
| | ↗ Samples | ↗ Features | ↗ Samples | ↗ Features | ↗ Noise | High dim. | Stability |
|---|---|---|---|---|---|---|---|
| GMM | Low/Low | Medium/Medium | Low | Medium | High | Medium | Medium |
| BGM | Low/Low | Medium/Medium | Low | Medium | High | Medium | High |
| DPGMM | Medium/Low | High/High | Low | High | High | High | High |
| RKDE | High/High | High/High | High | Low | High | High | High |
| PPCA | Low/Low | High/Low | Low | Low | High | Medium | Medium |
| LSA | Low/Medium | Low/Low | Medium | Low | Low | Low | Medium |
| MAHA | Low/Medium | Medium/Low | Low | Medium | Medium | Low | High |
| LOF | High/High | Low/Low | High | Low | Medium | High | High |
| ABOD | Low/High | Low/Medium | Low | Low | Medium | Low | Medium |
| SOD | High/High | Low/Medium | High | Low | Low | High | Medium |
| KL | Low/Medium | Low/Medium | Low | Medium | High | Medium | High |
| GWR | Medium/Medium | Medium/Low | Low | Low | Low | High | Medium |
| OCSVM | High/High | Low/Low | Low | Low | Low | High | High |
| IFOREST | Low/Medium | Low/Low | Medium | Low | High | High | Medium |

The density of ABOD is of great interest as it highlights some limitations of the method. In the case of a data distribution composed of several clusters, the lowest anomaly scores are located in the inter-cluster area instead of the cluster centroids due to the sole use of angles variance and values. This is caused by large angles measured when an angle targets two points belonging to distinct clusters, small angles when the points belong to the same cluster and thus a high overall variance for the inter-cluster area. In contrast, the dense areas surrounding the cluster centroids are assigned high anomaly scores since many angles are directed toward the other clusters, suggesting data points isolated and far from a major cluster. The same issue arises for the inter-cluster area when computing the anomaly scores with the alternative $k$-nearest neighbors approach instead of randomly sampling from the dataset.

SOD is able to estimate a very accurate density, despite some low scores between clusters. The neurons of the GWR network result in circular blue areas highlighting their position. The presence of a neuron at the center of the plot once again results in very low scores for the inter-cluster area, as low as for the theoretical cluster centroids. Using an additional threshold

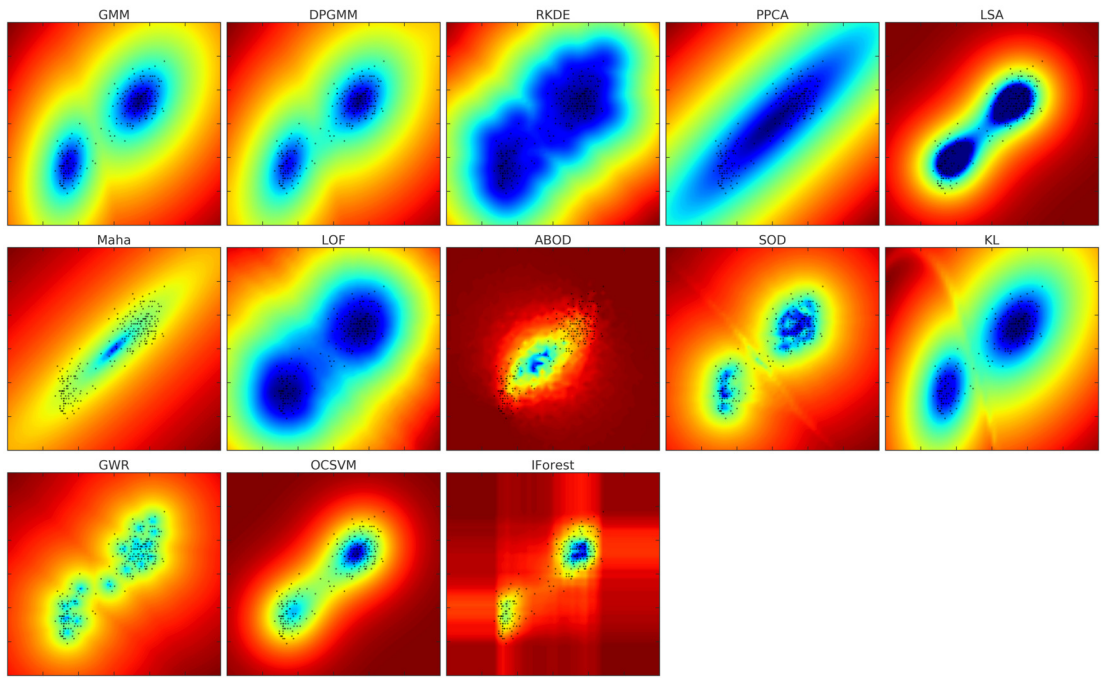

Figure 3.19: Segmentation contours per algorithm on the scaled OLD FAITHFUL dataset. Anomaly scores are normalized.

to detect outlying neurons as suggested in [Munoz & Muruzábal, 1998] would solve this issue.

A high density is also assigned by OCSVM to this region due to the mapping of the continuous decision boundary from the high-dimensional space. The segmentation made by IFOREST is much tighter than previous methods and seem less prone to overfitting than RKDE. Light trails emerging from the clusters can however be observed and may result in anomalous observations receiving a lower score than patterns slightly deviating from the mean.

## 3.4 SUMMARY

In this chapter, we described the Dirichlet Process Mixture Model, a flexible density estimation and clustering method. The method is trained through mean-field variational inference and provides a representation of the data based on a mixture of exponential-family distributions. This flexibility allows an accurate modeling of mixed-type features through a product of probabilistic distributions, which is made possible by mapping the chosen likelihood, the conjugate prior and the corresponding posterior in their exponential-family form. We pro-vided the derivation of this mapping for most standard probability distributions. As a result, a dataset composed of both categorical variables, floats and integers can be modelled through a dedicated mixture of categorical distributions, multivariate Normal distributions and Poisson distributions, while capturing the correlation between features for each trained cluster. A Dirichlet process is used to compute the mixing proportions of each component, and is regularized by a Beta prior. We also put a Gamma prior on the scaling parameter $w$ used by this Dirichlet process.

In the context of unsupervised anomaly detection, we benchmarked the average precision, robustness, computation time and memory usage of 14 algorithms on synthetic and real datasets. Our study demonstrates that IFOREST shows good novelty detection abilities while providing an excellent scalability on large datasets along with an acceptable memory usage for datasets up to one million samples. The results suggest that this algorithm is more suitable than RKDE in a production environment as the latter is much more computationally expensive and memory consuming. OCSVM is a good candidate in this benchmark, but it is not suitable either for large datasets.

Sampling a small proportion of outliers from classification datasets as suggested in [Emmott et al., 2016] resulted in dense clouds of outliers which allowed simple methods such as one-component GMM, KL, PPCA and the Mahalanobis distance to outperform several state-of-the-art density estimation algorithms. However, these scalable solutions cannot capture the complexity of most datasets where the nominal class does not follow a Gaussian distribution or is distributed across several clusters. In these cases, experiments show the superiority of nonparametric alternatives.

SOD showed good outlier detection performance and efficiently handled high-dimensional datasets at the cost of poor scalability. Exponential-family representations for DPMM revealed to be extremely time-consuming without substantially improving the detection of outliers made by Gaussian-based approaches such as DPGMM. Nonetheless, the use of categorical distributions in DPMM resulted in a reduced computation time when applied to mixed-type datasets in addition to better anomaly detection performance. The methods is thus efficient to model categorical and mixed-type data, but better alternatives exist in the litterature for numerical datasets. LOF, ABOD, GWR, KL and LSA reached the lowest performance while the three first methods also showed poor scalability. We assessed the modeling accuracy of each method and highlighted a borderline case for ABOD in the case of datasets composed of multiple clusters.

While the coverage of this study should suffice to tackle most novelty detection problems,

specific algorithms may be chosen for constrained environments. Distributed implementations, streaming or mini-batch training are a prerequisite to deal with large datasets, and several methods have been extended to support these features, e.g. DPMM, K-MEANS, SVM or GMM on Spark MLlib [Meng et al., 2016]. Other promising directions leading the research on outlier detection also focus on ensemble learning [Zimek et al., 2014] and detecting outliers from multi-view data [Iwata & Yamada, 2016]. DPMM could be improved by learning the truncation level $K$ on the number of components by variational inference. Detecting the suitable likelihood to model each input feature would also provide an additional flexibility to the model. Extending the method to support mini-batch training would eventually allow us to implement a distributed training for the algorithm.

*Variational inference is that thing you implement while waiting for your Gibbs sampler to converge.*

David Blei

# 4

# Deep Gaussian Process autoencoders for novelty detection

We have recently witnessed the rise of deep learning techniques as the preferred choice for supervised learning problems, due to their large representational power and the possibility to train these models at scale [LeCun et al., 2015]; examples of deep learning techniques achieving state-of-the-art performance on a wide variety of tasks include computer vision [Krizhevsky et al., 2012], speech recognition [Hinton et al., 2012], and natural language processing [Collobert & Weston, 2008]. A natural question is whether such impressive results can extend beyond supervised learning to unsupervised learning and further to novelty detection. Deep learning techniques for unsupervised learning are currently actively researched on [Kingma & Welling, 2014, Goodfellow et al., 2014], but it is still unclear whether these can compete with state-of-the-art novelty detection methods. We are not aware of recent surveys on neural networks for novelty detection, and the latest one we could find is almost fifteen years old [Markou & Singh, 2003] and misses the recent developments in this domain.

Key challenges with the use of deep learning methods in general learning tasks are (i) the necessity to specify a suitable architecture for the problem at hand and (ii) the necessity to control their generalization. While various forms of regularization have been proposed to mitigate the overfitting problem and improve generalization, e.g., through the use of dropout [Srivastava et al., 2014b, Gal & Ghahramani, 2016], there are still open questions on how to devise principled ways of applying deep learning methods to general learning tasks. Deep Gaussian Processes (DGPs) are ideal candidates to simultaneously tackle issues (i) and (ii) above. DGPs are deep nonparametric probabilistic models implementing a composition of probabilistic processes that implicitly allows for the use of an infinite number of neurons at each layer [Damianou & Lawrence, 2013, Duvenaud et al., 2014]. Also, their probabilistic nature induces a form of regularization that prevents overfitting, and allows for a principled way of carrying out model selection [Neal, 1996]. While DGPs are particularly appealing to tackle general deep learning problems, their training is computationally intractable. Recently, there have been contributions in the direction of making the training of these models tractable [Bui et al., 2016, Cutajar et al., 2017, Bradshaw et al., 2017], and these are currently in the position to compete with Deep Neural Networks (DNNs) in terms of scalability, accuracy, while providing superior quantification of uncertainty [Gal & Ghahramani, 2016, Cutajar et al., 2017, Gal et al., 2017].

In this chapter, we introduce an unsupervised model for novelty detection based on DGPs in autoencoder configuration. We train the proposed DGP autoencoder (DGP-AE) by approximating the DGP layers using random feature expansions, and by performing stochastic variational inference on the resulting approximate model. The key features of the proposed approach are as follows: (i) DGP-AEs are unsupervised probabilistic models that can deal with highly complex data distribution and offer a scoring method for novelty detection; (ii) DGP-AEs can model any type of data including cases with mixed-type features, such as continuous, discrete, and count data; (iii) DGP-AEs training does not require any expensive and potentially numerically troublesome matrix factorizations, but only tensor products; (iv) DGP-AEs can be trained using mini-batch learning, and could therefore exploit distributed and GPU computing; (v) DGP-AEs training using stochastic variational inference can be easily implemented taking advantage of automatic differentiation tools, making for a very practical and scalable methods for novelty detection.

We compare DGP-AEs with a number of competitors that have been proposed in the literature of deep learning to tackle large-scale unsupervised learning problems, such as Variational Autoencoders (VAE) [Kingma & Welling, 2014], Variational Auto-Encoded Deep Gaussian Process (VAE-DGP) [Dai et al., 2016] and Neural Autoregressive Distribution Estimator (NADE) [Uria et al., 2016]. Through a series of experiments, where we also compare against state-of-the-art novelty detection methods such as Isolation Forest [Liu et al., 2008] and Robust Kernel Density Estimation [Kim & Scott, 2012], we demonstrate that DGP-AEs offer flexible modeling capabilities with a practical learning algorithm, while achieving state-of-the-art performance.

The related work on the state-of-the-art was introduced in Section 2.1.7. The remainder of this chapter is organized as follows: Section 4.1 presents the proposed DGP-AE for novelty detection, while Section 4.2 and Section 4.3 report the experiments and conclusions.

## 4.1 Deep Gaussian Process Autoencoders

This section presents the proposed DGP-AE model and describes the approximation that we use to make inference tractable and scalable. Each iteration of the algorithm is linear in dimensionality of the input, batch size, dimensionality of the latent representation and number of Monte Carlo samples used in the approximation of the objective function, which highlights the tractability of the model. We also discuss the inference scheme based on stochastic variational inference, and show how predictions can be made. Finally, we present ways in which we can make the proposed DGP-AE model handle various types of data, e.g., mixing continuous and categorical features. We refer the reader to [Cutajar et al., 2017] for a detailed derivation of the random feature approximation of DGPs and variational inference of the resulting model. In this work, we extend this DGP formulation to autoencoders.

### 4.1.1 Architecture

An autoencoder is a model combining an encoder and a decoder. The encoder part takes each input $\mathbf{x}$ and maps it into a set of latent variables $\mathbf{z}$, whereas the decoder part maps latent variables $\mathbf{z}$ into the inputs $\mathbf{x}$. Because of their structure, autoencoders are able to jointly learn latent representations for a given dataset and a model to produce $\mathbf{x}$ given latent variables $\mathbf{z}$. Typically this is achieved by minimizing a reconstruction error.

Autoencoders are not generative models, and variational autoencoders have recently been proposed to enable this feature [Dai et al., 2016, Kingma & Welling, 2014]. In the context of novelty detection, the possibility to learn a generative model might be desirable but not essential, so in this work we focus in particular on autoencoders. Having said that, we believe that extending variational autoencoders using the proposed framework is possible, as well as empowering the current model to enable generative modeling; we leave these avenues of research for future work. In this work, we propose to construct the encoder and the decoder functions of autoencoders using DGPs. As a result, we aim at jointly learning a probabilistic nonlinear projection based on DGPs (the encoder) and a DGP-based latent variable model (the decoder).

The building block of DGPs are GPs, which are priors over functions; formally, a GP is a set of random variables characterized by the property that any subset of them is jointly Gaussian [Rasmussen & Williams, 2006]. The GP covariance function models the covariance between the random variables at different inputs, and it is possible to specify a parametric function for their mean.

Stacking multiple GPs into a DGP means feeding the output of GPs at each layer as the input of the GPs at the next; this construction gives rise to a composition of stochastic processes. Assume that we compose $N_L$ possible functions modelled as multivariate GPs, the resulting composition takes the form

$$\mathbf{f}(\mathbf{x}) = \left( \mathbf{f}^{(N_L)} \circ \ldots \circ \mathbf{f}^{(1)} \right) (\mathbf{x}),$$ (4.1)

Without loss of generality, we are going to assume that the GPs at each layer have zero mean, and that GP covariances at layer $(l)$ are parameterized through a set of parameters $\boldsymbol{\theta}^{(l)}$ shared across GPs in the same layer.

Denote by $F^{(i)}$ the collection of the multivariate functions $\mathbf{f}^{(i)}$ evaluated at the inputs $F^{(i-1)}$, and define $F^{(0)} := X$. The encoder part of the proposed DGP-AE model maps the inputs $X$ into a set of latent variables $Z := F^{(j)}$ through a DGP, whereas the decoder is another DGP mapping $Z$ into $X$. The DGP controlling the decoding part of the model, assumes a likelihood function that allows one to express the likelihood of the observed data $X$ as $p\left( X | F^{(N_L)}, \boldsymbol{\theta}^{(N_L)} \right)$. The likelihood reflects the choice on the mappings between latent variables and the type of data being modelled, and it can include and mix various types and dimensionality; section 3.5 discusses this in more detail.

By performing Bayesian inference on the proposed DGP-AE model we aim to integrate out latent variables at all layers, effectively integrating out the uncertainty in all the mappings in the encoder/decoder and the latent variables $Z$ themselves. Learning and making predictions with DGP-AEs, however, require being able to solve intractable integrals. To evaluate the marginal likelihood expressing the probability of observed data given model parameters, we need to solve the following

$$p(X | \boldsymbol{\theta}) = \int p\left( X | F^{(N_L)}, \boldsymbol{\theta}^{(N_L)} \right) \prod_{j=1}^{N_L} p\left( F^{(j)} | F^{(j-1)}, \boldsymbol{\theta}^{(j-1)} \right) \prod_{j=1}^{N_L} dF^{(j)}$$ (4.2)

A similar intricate integral can be derived to express the predictive probability $p(\mathbf{x}_* | X, \boldsymbol{\theta})$. For any nonlinear covariance function, these integrals are intractable. In the next section, we show how random feature expansions of the GPs at each layer expose an approximate model that

can be conveniently learned using stochastic variational inference, as described in [Cutajar et al., 2017].

### 4.1.2 RANDOM FEATURE EXPANSIONS FOR DGP-AES

To start with, consider a shallow multivariate GP and denote by $F$ the latent variables associated with the inputs. For a number of GP covariance functions, it is possible to obtain a low-rank approximation of the processes through the use of a finite set of basis functions, and transform the multivariate GP into a Bayesian linear model. For example, in the case of a radial basis covariance function (RBF) of the form

$$k_{\text{rbf}}(\mathbf{x}, \mathbf{x}') = \exp\left[-\frac{1}{2}\left\|\mathbf{x} - \mathbf{x}'\right\|^\top\right] \tag{4.3}$$

it is possible to employ standard Fourier analysis to show that $k_{\text{rbf}}$ can be expressed as an expectation under a distribution over spectral frequencies, that is:

$$k_{\text{rbf}}(\mathbf{x}, \mathbf{x}') = \int p(\boldsymbol{\omega}) \exp\left[i(\mathbf{x} - \mathbf{x}')^\top \boldsymbol{\omega}\right] d\boldsymbol{\omega}. \tag{4.4}$$

After standard manipulation, it is possible to obtain an unbiased estimate of the integral above by mean of a Monte Carlo average:

$$k_{\text{rbf}}(\mathbf{x}, \mathbf{x}') \approx \frac{1}{N_{\text{RF}}} \sum_{r=1}^{N_{\text{RF}}} \mathbf{z}(\mathbf{x}|\tilde{\boldsymbol{\omega}}_r)^\top \mathbf{z}(\mathbf{x}'|\tilde{\boldsymbol{\omega}}_r), \tag{4.5}$$

where $\mathbf{z}(\mathbf{x}|\boldsymbol{\omega}) = [\cos(\mathbf{x}^\top \boldsymbol{\omega}), \sin(\mathbf{x}^\top \boldsymbol{\omega})]^\top$ and $\tilde{\boldsymbol{\omega}}_r \sim p(\boldsymbol{\omega})$. It is possible to increase the flexibility of the RBF covariance above by scaling it by a marginal variance parameter $\sigma^2$ and by scaling the features individually with length-scale parameters $\Lambda = \text{diag}(l_1^2, \cdots, l_{D_F^{(l)}}^2)$; it is then possible to show that $p(\boldsymbol{\omega}) = N\left(\boldsymbol{\omega}|\mathbf{0}, \Lambda^{-1}\right)$ using Bochner's theorem. By stacking the samples from $p(\boldsymbol{\omega})$ by column into a matrix $\Omega$, we can define

$$\Phi_{\text{rbf}} = \sqrt{\frac{(\sigma^2)}{N_{\text{RF}}}}\left[\cos\left(F\Omega\right), \sin\left(F\Omega\right)\right], \tag{4.6}$$

where the functions $\cos()$ and $\sin()$ are applied element-wise. We can now derive a low-rank approximation of $K$ as follows:

$$K \approx \Phi\Phi^\top \tag{4.7}$$

It is straightforward to verify that the individual columns of $F$ in the original GP can be approximated by the Bayesian linear model $F_{\cdot j} = \Phi W_{\cdot j}$ with $W_{\cdot j} \sim N(\mathbf{0}, I)$, as the covariance of $F_{\cdot j}$ is indeed $\Phi \Phi^\top \approx K$.

The decomposition of the GP covariance in equation 4.3 suggests an expansion with an infinite number of basis functions, thus leading to a well-known connection with single-layered neural networks with infinite neurons [Neal, 1996]; the random feature expansion that we perform using Monte Carlo induces a truncation of the infinite expansion. Based on the expansion defined above, we can now build a cascade of approximate GPs, where the output of layer $l$ becomes the input of layer $l + 1$. The layer $\Phi^{(0)}$ first expands the input features in a high-dimensional space, followed by a linear transformation parameterized by a weight matrix $W^{(0)}$ which results in the latent variables $F^{(1)}$ in the second layer. Considering a DGP with RBF covariances obtained by stacking the hidden layers previously described, we obtain equations 4.8 and 4.9 derived from equation 4.5. These transformations are parameterized by prior parameters $(\sigma^2)^{(l)}$ which determine the marginal variance of the GPs and $\Lambda^{(l)} = \mathrm{diag}\left( \left(l_1^2\right)^{(l)}, \cdots, \left(l_{D_F^{(l)}}^2\right)^{(l)} \right)$ describing the length-scale parameters.

$$\Phi_{\mathrm{rbf}}^{(l)} = \sqrt{\frac{(\sigma^2)^{(l)}}{N_{\mathrm{RF}}^{(l)}}} \left[ \cos\left( F^{(l)} \Omega^{(l)} \right), \sin\left( F^{(l)} \Omega^{(l)} \right) \right], \tag{4.8}$$

$$F^{(l+1)} = \Phi_{\mathrm{rbf}}^{(l)} W^{(l)} \tag{4.9}$$

This leads to the proposed DGP-AE model's topology given in Figure 4.1. The resulting approximate DGP-AE model is effectively a Bayesian DNN where the priors for the spectral frequencies $\Omega^{(l)}$ are controlled by covariance parameters $\boldsymbol{\theta}^{(l)}$, and the priors for the weights $W^{(l)}$ are standard normal.

In our framework, the choice of the covariance function induces different basis functions. For example, a possible approximation of the ARC-cosine kernel [Cho & Saul, 2009] yields Rectified Linear Units (RELU) basis functions [Cutajar et al., 2017] resulting in faster computations compared to the approximation of the RBF covariance, given that derivatives of RELU basis functions are cheap to evaluate.

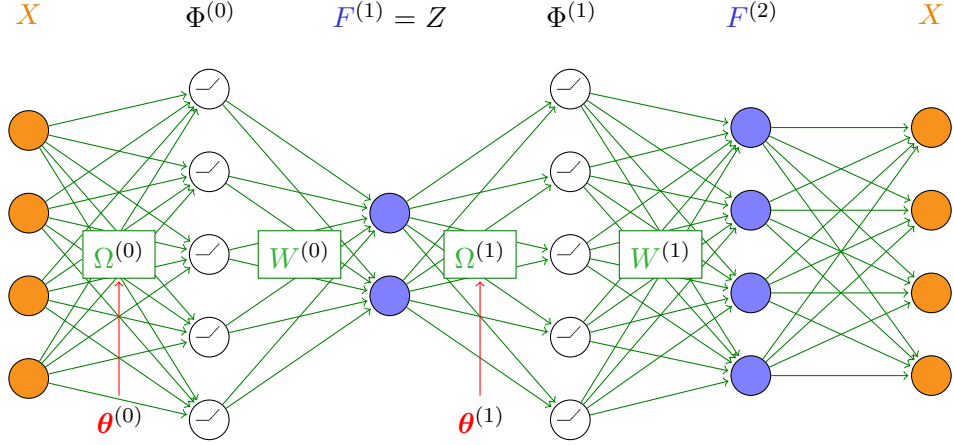

Figure 4.1: Architecture of a 2-layer DGP autoencoder. Gaussian processes are approximated by hidden layers composed of two inner layers, the first layer $\Phi^{(l)}$ performing random feature expansion followed by a linear transformation resulting in $F^{(l)}$. Covariance parameters are $\theta^{(l)} = \left((\sigma^2)^{(l)}, \Lambda^{(l)}\right)$, with prior over the weights $p\left(\Omega_{\cdot j}^{(l)}\right) = N\left(0, \left(\Lambda^{(l)}\right)^{-1}\right)$ and $p\left(W_{\cdot i}^{(l)}\right) = N(0, I)$. $Z$ is the latent variables representation.

### 4.1.3 STOCHASTIC VARIATIONAL INFERENCE FOR DGP-AES

Let $\boldsymbol{\Theta}$ be the collection of all covariance parameters $\boldsymbol{\theta}^{(l)}$ at all layers; similarly, define $\boldsymbol{\Omega}$ and $\mathbf{W}$ to be the collection of the spectral frequencies $\Omega^{(l)}$ and weight matrices $W^{(l)}$ at all layers, respectively. We are going to apply stochastic variational inference techniques to infer $\mathbf{W}$ and optimize all covariance parameters $\boldsymbol{\Theta}$; we are going to consider the case where the spectral frequencies $\boldsymbol{\Omega}$ are fixed, but these can also be learned [Cutajar et al., 2017]. The marginal likelihood $p(X|\boldsymbol{\Omega}, \boldsymbol{\Theta})$ can be bounded using standard variational inference techniques, following [Kingma & Welling, 2014] and [Graves, 2011], Defining $\mathcal{L} = \log\left[p(X|\boldsymbol{\Omega}, \boldsymbol{\Theta})\right]$, we obtain

$$\mathcal{L} \geq \mathrm{E}_{q(\mathbf{W})}\left(\log\left[p\left(X|\mathbf{W}, \boldsymbol{\Omega}, \boldsymbol{\Theta}\right)\right]\right) - \mathrm{DKL}\left[q(\mathbf{W})\|p\left(\mathbf{W}\right)\right], \qquad (4.10)$$

Here the distribution $q(\mathbf{W})$ denotes an approximation to the intractable posterior $p(\mathbf{W}|X, \boldsymbol{\Omega}, \boldsymbol{\Theta})$, whereas the prior on $\mathbf{W}$ is the product of standard normal priors resulting from the approximation of the GPs at each layer $p(\mathbf{W}) = \prod_{l=0}^{N_L - 1} p(W^{(l)})$.

We are going to assume an approximate Gaussian distribution that factorizes across layers

and weights

$$q(\mathbf{W}) = \prod_{ijl} q\left(W_{ij}^{(l)}\right) = \prod_{ijl} N\left(m_{ij}^{(l)}, (s^2)_{ij}^{(l)}\right). \tag{4.11}$$

We are interested in finding an optimal approximate distribution $q(\mathbf{W})$, so we are going to introduce the variational parameters $m_{ij}^{(l)}, (s^2)_{ij}^{(l)}$ to be the mean and the variance of each of the approximating factors. Therefore, we are going to optimize the lower bound above with respect to all variational parameters and covariance parameters $\boldsymbol{\Theta}$.

Because of the chosen Gaussian form of $q(\mathbf{W})$ and given that the prior $p(\mathbf{W})$ is also Gaussian, the DKL term in the lower bound to $\mathcal{L}$ can be computed analytically. The remaining term in the lower bound, instead, needs to be estimated. Assuming a likelihood that factorizes across observations, it is possible to perform a doubly-stochastic approximation of the expectation in the lower bound so as to enable scalable stochastic gradient-based optimization. The doubly-stochastic approximation amounts in replacing the sum over $n$ input points with a sum over a mini-batch of $m$ points selected randomly from the entire dataset:

$$\mathrm{E}_{q(\mathbf{W})}\left(\log\left[p\left(X|\mathbf{W}, \boldsymbol{\Omega}, \boldsymbol{\Theta}\right)\right]\right) \approx \frac{n}{m} \sum_{k \in \mathcal{G}_m} \mathrm{E}_{q(\mathbf{W})}(\log[p(\mathbf{x}_k|\mathbf{W}, \boldsymbol{\Omega}, \boldsymbol{\Theta})]), \tag{4.12}$$

Then, each element of the sum can itself be estimated unbiasedly using Monte Carlo sampling and averaging, with $\tilde{\mathbf{W}}_r \sim q(\mathbf{W})$:

$$\mathrm{E}_{q(\mathbf{W})}\left(\log\left[p\left(X|\mathbf{W}, \boldsymbol{\Omega}, \boldsymbol{\Theta}\right)\right]\right) \approx \frac{n}{m} \sum_{k \in \mathcal{G}_m} \frac{1}{N_{\mathrm{MC}}} \sum_{r=1}^{N_{\mathrm{MC}}} \log[p(\mathbf{x}_k|\tilde{\mathbf{W}}_r, \boldsymbol{\Omega}, \boldsymbol{\Theta})], \tag{4.13}$$

Because of the unbiasedness property of the last expression, computing its derivative with respect to the variational parameters and $\boldsymbol{\Theta}$ yields a so-called stochastic gradient that can be used for stochastic gradient-based optimization. The appeal of this optimization strategy is that it is characterized by theoretical guarantees to reach local optima of the objective function [Robbins & Monro, 1951]. Derivatives can be conveniently computed using automatic differentiation tools; we implemented our model in TensorFlow [Abadi et al., 2015] that has this feature built-in. In order to take derivatives with respect to the variational parameters we employ the so-called reparameterization trick [Kingma & Welling, 2014]

$$\left(\tilde{W}_r^{(l)}\right)_{ij} = s_{ij}^{(l)} \epsilon_{rij}^{(l)} + m_{ij}^{(l)}, \tag{4.14}$$

to fix the randomness when updating the variational parameters, and $\epsilon_{rij}^{(l)}$ are resampled after each iteration of the optimization.

### 4.1.4 Predictions with DGP-AES

The predictive distribution for the proposed DGP-AE model requires solving the following integral

$$p(\mathbf{x}_*|X, \boldsymbol{\Omega}, \boldsymbol{\Theta}) = \int p(\mathbf{x}_*|\mathbf{W}, \boldsymbol{\Omega}, \boldsymbol{\Theta})p(\mathbf{W}|X, \boldsymbol{\Omega}, \boldsymbol{\Theta})d\mathbf{W}, \qquad (4.15)$$

which is intractable due to fact that the posterior distribution over $\mathbf{W}$ is unavailable. Stochastic variational inference yields an approximation $q(\mathbf{W})$ to the posterior $p(\mathbf{W}|X, \boldsymbol{\Omega}, \boldsymbol{\Theta})$, so we can use it to approximate the predictive distribution above:

$$p(\mathbf{x}_*|X, \boldsymbol{\Omega}, \boldsymbol{\Theta}) \approx \int p(\mathbf{x}_*|\mathbf{W}, \boldsymbol{\Omega}, \boldsymbol{\Theta})q(\mathbf{W})d\mathbf{W} \approx \frac{1}{N_{\text{MC}}}\sum_{r=1}^{N_{\text{MC}}} p(\mathbf{x}_*|\tilde{\mathbf{W}}_r, \boldsymbol{\Omega}, \boldsymbol{\Theta}), \qquad (4.16)$$

where we carried out a Monte Carlo approximation by drawing $N_{\text{MC}}$ samples $\tilde{\mathbf{W}}_r \sim q(\mathbf{W})$. The overall complexity of each iteration is thus $\mathcal{O}\left(mD_F^{(l-1)}N_{RF}^{(l)}N_{MC}\right)$ to construct the random features at layer $l$ and $\mathcal{O}\left(mN_{RF}^{(l)}D_F^{(l)}N_{MC}\right)$ to compute the value of the latent functions at layer $l$, where $m$ is the batch size and $D_F^{(l)}$ is the dimensionality of $F^{(l)}$. Hence, by carrying out updates using mini-batches, the complexity of each iteration is independent of the dataset size.

For a given test set $X_*$ containing multiple test samples, it is possible to use the predictive distribution as a scoring function to identify novelties. In particular, we can rank the predictive probabilities $p(\mathbf{x}_*|X, \boldsymbol{\Omega}, \boldsymbol{\Theta})$ for all test points to identify the ones that have the lowest probability under the given DGP-AE model. In practice, for numerical stability, our implementation uses log-sum operations to compute $\log[p(\mathbf{x}_*|X, \boldsymbol{\Omega}, \boldsymbol{\Theta})]$, and we use this as the scoring function.

### 4.1.5 Likelihood functions

One of the key features of the proposed model is the possibility to model data containing a mix of types of features. In order to do this, all we need to do is to specify a suitable likelihood for the observations given the latent variables at the last layer, that is $p(\mathbf{x}|\mathbf{f}^{(N_L)})$. Imagine that the vector $\mathbf{x}$ contains continuous and categorical features that we model using Gaussian and multinomial likelihoods; extensions to other combinations of features and distributions is straightforward. Consider a single continuous feature of $\mathbf{x}$, say $x_{[G]}$; the likelihood function for this feature is:

$$p(x_{[G]}|\mathbf{f}^{(N_L)}) = N(x_{[G]}|f_{[G]}^{(N_L)}, \sigma_{[G]}^2). \qquad (4.17)$$

For any given categorical feature, instead, assuming a one-hot encoding, say $\mathbf{x}_{[C]}$, we can use a multinomial likelihood with probabilities given by the softmax transformation of the corresponding latent variables:

$$p((\mathbf{x}_{[C]})_j|\mathbf{f}^{(N_L)}) = \frac{\exp[(f_{[C]}^{(N_L)})_j]}{\sum_i \exp[(f_{[C]}^{(N_L)})_i]}. \qquad (4.18)$$

It is now possible to combine any number of these into the following likelihood function:

$$p(\mathbf{x}|\mathbf{f}^{(N_L)}) = \prod_k p(\mathbf{x}_{[k]}|\mathbf{f}^{(N_L)}) \qquad (4.19)$$

Any extra parameters in the likelihood function, such as the variances in the Gaussian likelihoods, can be included in the set of all model parameters $\mathbf{\Theta}$ and learned jointly with the rest of parameters. For count data, it is possible to use the Binomial or Poisson likelihood, whereas for positive continuous variables we can use Exponential or Gamma. It is also possible to jointly model multiple continuous features and use a full covariance matrix for multivariate Gaussian likelihoods, multivariate Student-T, and the like. The nice feature of the proposed DGP-AE model is that the training procedure is the same regardless of the choice of the likelihood function, as long as the assumption of factorization across data points holds.

## 4.2 Experiments

We evaluate the performance of our model by monitoring the convergence of the mean log-likelihood (MLL) and by measuring the area under the Precision-Recall curve, namely the mean average precision (MAP) on real-world datasets described in section 4.2.2.

### 4.2.1 Algorithms

In order to retrieve a continuous anomaly score and to compare the convergence of the likelihood for the selected models, our comparison focuses on the probabilistic neural networks introduced in Section 2.1.8. The parameters used in the experiments are detailed in Table 4.1.

Table 4.1: Parameters and implementations of the selected methods, where $i$ is the number of iterations, $b$ is the batch size, $rf$ is the number of random features, $d$ is the dimensionality of the input data, $k$ is the number of components, $N$ and $S$ are the Normal and Softmax likelihoods, respectively.

| Algorithm | Parameters |
|---|---|
| DGP-AE G-1 | $i = 1e5, b = 200, lr = 0.01, rf = 100, gp = d, q(\mathbf{\Omega})_{fixed} = 1000, \mathbf{\Theta}_{fixed} = 7000, mc_{train} = 1, mc_{test} = 100, ll = N$ |
| DGP-AE G-2 | $i = 1e5, b = 200, lr = 0.01, rf = 100, gp = \{d, 3, d\}, q(\mathbf{\Omega})_{fixed} = 1000, \mathbf{\Theta}_{fixed} = 7000, mc_{train} = 1, mc_{test} = 100, ll = N$ |
| DGP-AE GS-1 | $i = 1e5, b = 200, lr = 0.01, rf = 100, gp = d, q(\mathbf{\Omega})_{fixed} = 1000, \mathbf{\Theta}_{fixed} = 7000, mc_{train} = 1, mc_{test} = 100, ll = \{N, S\}$ |
| DGP-AE GS-2 | $i = 1e5, b = 200, lr = 0.01, rf = 100, gp = \{d, 3, d\}, q(\mathbf{\Omega})_{fixed} = 1000, \mathbf{\Theta}_{fixed} = 7000, mc_{train} = 1, mc_{test} = 100, ll = \{N, S\}$ |
| VAE-DGP-2 | $nl = 2, epoch = 1000, units = \{max(\frac{d}{2}, 5), max(\frac{d}{3}, 4)\}, kernel = \text{RBF}, inducing\_pts = 40, mlp\_units = \{300, 150\}$ |
| AE-1 | $nl = 1, i = 1e5, b = 200, lr = 0.01, units = d, activation = sigmoid, dropout = 0.5$ |
| AE-5 | $nl = 5, i = 1e5, b = 200, lr = 0.01, units = \{d, 0.8d, 0.6d, 0.8d, d\}, activation = sigmoid, dropout = 0.5$ |
| VAE-1 | $nl = 1, i = 4000, b = 1000, lr = 0.001, hiden = 50$ |
| VAE-2 | $nl = 2, i = 4000, b = 1000, lr = 0.001, hiden = \{100, 100\}$ |
| NADE-2 | $nl = 2, i = 5000, b = 200, lr = 0.005, decay = 0.02, units = \{100, 100\}, activation = \text{RELU}, k = 20$ |
| RKDE | $bandwidth = \text{LKCV}, loss = Huber$ |
| IFOREST | $contamination = 0.5$ |

Parameter selection was achieved by grid-search and maximizes the MAP averaged over the testing datasets labelled for novelty detection and described in section 4.2.2. As a result, the methods use the same parameter settings for all datasets, which may still depend on the the datasets characteristics, e.g. $units = \frac{d}{2}$, where $d$ is the dimensionality. These can be considered as recommended default parameters for future novelty detection tasks. The depth of the networks is added to the name as a suffix, e.g. VAE-2.

Our DGP-AE is benchmarked against two Variational Autoencoders [Kingma & Welling, 2014] named VAE-1 and VAE-2. We train these two networks for 4000 iterations using a batch size of 1000 samples, a learning rate of 0.001 and an architecture of 50 hidden units. We also evaluate the Neural Autoregressive Distribution Estimator (NADE-2) [Uria et al., 2016], which is trained for 5000 iterations using batches of 200 samples, a learning rate of 0.005 and a weight decay of 0.02. Training this network for more iterations increases the risk of the training to fail due to runtime errors. The network has a 2 layer-topology with 100 hidden units and a RELU activation function. The number of components for the mixture of Gaussians was set to 20, and we use Bernoulli distributions instead of Gaussians to model datasets exclusively composed of categorical data. 15% of the training data was used for validation to select the final weights.

To showcase the performance of random feature approximation, we include VAE-DGP-2

[Dai et al., 2016], a Deep Gaussian Process network trained with variational inference through inducing points approximation. The network uses two layers of dimensionality $max(\frac{d}{2}, 5)$ and $max(\frac{d}{3}, 4)$, and is trained for 1000 epochs over all training samples. All layers use a RBF kernel with 40 inducing points. We use 300 and 150 units in the two-layer MLP.

We also include standard DNN autoencoders (AE-1, AE-5) with sigmoid activation functions and dropout regularization to give a wider context to the reader. AE-1 uses a number of hidden units equal to the number of features. AE-5 uses 80% of the number of input features on the second and fourth layer, and 60% on the third layer. The two networks are trained for 100,000 iterations with a batch size of 200 samples and a learning rate of 0.01.

We initially intended to include Real NVP [Dinh et al., 2016] and Wasserstein GAN [Arjovsky et al., 2017], but we found these networks and their implementations tightly tailored to images. The one-class classification with GPs recently developed [Kemmler et al., 2013] is actually a supervised learning task where the authors regress on the labels and use heuristics to score novelties. Since this work is neither probabilistic nor a neural network, we did not include it.

To demonstrate the value of our proposal as a competitive novelty detection method, we include top performance novelty detection methods from other domains, namely Isolation Forest (IFOREST) [Liu et al., 2008] and Robust Kernel Density Estimation (RKDE) [Kim & Scott, 2012], which are recommended for outlier detection in [Emmott et al., 2016]. Isolation Forest uses a contamination rate of 5% while RKDE uses the LKCV bandwidth and the Huber loss function.

We train the proposed DGP-AE model for 100,000 iterations using 100 random features at each hidden layer. Due to the network topology, we use a number of multivariate GPs equal to the number of input features when using a single-layer configuration, but use a multivariate GP of dimension 3 for the latent variables representation when using more than one layer. In the remainder of the thesis and when referring to deep Gaussian process autoencoders, the term *layer* describes a hidden layer composed of two inner layers $\Phi^{(i)}$ and $F^{(i+1)}$. As observed in [Duvenaud et al., 2014, Neal, 1996], deep architectures require to feed forward the input to the hidden layers in order to implement the modeling of meaningful functions. In the experiments involving more than 2 layers, we follow this advice by feed-forwarding the input to the encoding layers and feed-forward the latent variables to the decoding layers. The weights are optimized using a batch size of 200 and a learning rate of 0.01. The parameters $q(\boldsymbol{\Omega})$ and $\boldsymbol{\Theta}$ are fixed for 1000 and 7000 iterations respectively. $N_{\mathrm{MC}}$ is set to 1 during the training, while we use $N_{\mathrm{MC}} = 100$ at test time to score samples with higher accuracy. DGP-AE-G uses a Gaussian likelihood for continuous and one-hot encoded categorical variables. DGP-AE-GS is a modified DGP-AE-G where categorical features are modelled by a softmax likelihood as previously described. These networks use an RBF covariance function, except when the ARC suffix is used, e.g. DGP-AE-G-1-ARC.

### 4.2.2 Datasets

Our evaluation is based on 11 datasets, including 7 datasets made publicly available by the UCI [Asuncion & Newman, 2007], while the 4 other datasets are proprietary datasets containing production data from the company Amadeus. This company provides online platforms to connect the travel industry and manages almost half of the flight bookings worldwide. Their business is targeted by fraud attempts reported as outliers in the corresponding datasets. The proprietary datasets are given thereafter; PNR describes the history of changes applied to booking records, TRANSACTIONS depicts user sessions performed on a Web application and targets the detection of bots and malicious users, SHARED-ACCESS was extracted from a backend application dedicated to shared rights management between customers, e.g. seat map display or cruise distribution, and PAYMENT-SUB reports the booking records along with the user behavior through the booking process, e.g. searches and actions performed. Table 4.2 shows the datasets characteristics. Most datasets used in this experiment are also reported in Table 3.3, although we added PAYMENT-SUB and AIRLINE and removed the classification datasets which could induce clusters of outliers (See Section 3.3.1).

### 4.2.3 Results

This section shows the outlier detection capabilities of the methods and monitors the MLL to exhibit convergence. We also study the impact of depth and dimensionality on DGP-AEs, and plot the latent representations learnt by the network.

#### Method comparison

Our experiment performs a 5-fold Monte Carlo cross-validation, using 80% of the original dataset for the training and 20% for the testing. Training and testing datasets are normalized, and we use the characteristics of the training dataset to normalize the testing data. Both datasets contain the same proportion of anomalies. Since class distribution is by nature heavily imbalanced for novelty detection problems, we use the MAP as a performance metric instead of the average ROC AUC. The detailed MAP are reported in Table 4.3. Bold results are similar to the best MAP achieved on the dataset with nonsignificant differences. We used a pairwise

Table 4.2: UCI and proprietary datasets benchmarked - (# categ. dims) is the number of binary features after one-hot encoding of the categorical features.

| Dataset | Nominal class | Anomaly class | Numeric dims | Categ. dims | Samples | Anomalies |
|---|---|---|---|---|---|---|
| MAMMOGRAPHY | -1 | 1 | 6 | 0 (0) | 11,183 | 260 (2.32%) |
| MAGIC-GAMMA-SUB | g | h | 10 | 0 (0) | 12,332 | 408 (3.20%)[1] |
| WINE-QUALITY | 4, 5, 6, 7, 8 | 3, 9 | 11 | 0 (0) | 4,898 | 25 (0.51%) |
| MUSHROOM-SUB | e | p | 0 | 22 (107) | 4,368 | 139 (3.20%)[1] |
| CAR | unacc, acc, good | vgood | 0 | 6 (21) | 1,728 | 65 (3.76%) |
| GERMAN-SUB | 1 | 2 | 7 | 13 (54) | 723 | 23 (3.18%)[1] |
| PNR | 0 | 1, 2, 3, 4, 5 | 82 | 0 (0) | 20,000 | 121 (0.61%) |
| TRANSACTIONS | 0 | 1 | 41 | 1 (9) | 10,000 | 21 (0.21%) |
| SHARED-ACCESS | 0 | 1 | 49 | 0 (0) | 18,722 | 37 (0.20%) |
| PAYMENT-SUB | 0 | 1 | 37 | 0 (0) | 73,848 | 2769 (3.75%) |
| AIRLINE | 1 | 0 | 8 | 0 (0) | 3,188,179 | 203,501 (6.00%) |

[1] Anomalies are sampled from the corresponding class, using the average percentage of outliers depicted in [Emmott et al., 2016].

Friedman test [García et al., 2010] with a threshold of 0.05 to reject the null hypothesis. The experiments are performed on an Ubuntu 14.04 LTS powered by an Intel Xeon E5-4627 v4 CPU and 256GB RAM. This amount of memory is not sufficient to train RKDE on the AIRLINE dataset, resulting in missing data in Table 4.3.

Looking at the average performance, our DGPS autoencoders achieve the best results for novelty detection. DGPS performed well on all datasets, including high dimensional cases, and outperform the other methods on WINE-QUALITY, AIRLINE and PNR. By fitting a softmax likelihood instead of a Gaussian on one-hot encoded features, DGP-AE-GS-1 achieves better performance than DGP-AE-G-1 on 3 datasets containing categorical variables out of 4, e.g. MUSHROOM-SUB, GERMAN-SUB and TRANSACTIONS, while showing similar results on the CAR dataset. This representation allows DGPS to reach the best performance on half of the datasets and to outperform state-of-the-art algorithms for novelty detection, such as RKDE and IForest. Despite the low dimensionality representation of the latent variables, DGP-AE-G-2 achieves performance comparable to DGP-AE-G-1, which suggests good dimensionality reduction abilities. The use of a softmax likelihood in DGP-AE-GS-2 resulted in better novelty detection capabilities than DGP-

Table 4.3: Mean area under the precision-recall curve (MAP) per dataset and algorithm (5 runs). Bold results imply that we cannot reject the null hypothesis of a given MAP to be identical to the best result for the dataset. DGP-AEs are different configurations of the proposed algorithm, while VAE-DGP refers to [Dai et al., 2016]. The performance of RKDE on the AIRLINE dataset is missing due to the lack of scalability of the algorithm.

| | DGP-AE G-1 | DGP-AE G-2 | DGP-AE GS-1 | DGP-AE GS-2 | VAE-DGP-2 | AE-1 | AE-5 | VAE-1 | VAE-2 | NADE-2 | RKDE | IForest |
|---|---|---|---|---|---|---|---|---|---|---|---|---|
| MAMMOGRAPHY | **0.222** | 0.183 | **0.222** | 0.183 | 0.221 | 0.118 | 0.075 | 0.119 | 0.148 | 0.193 | **0.231** | **0.244** |
| MAGIC-GAMMA-SUB | 0.260 | 0.340 | 0.260 | 0.340 | 0.235 | 0.253 | 0.125 | 0.230 | 0.305 | **0.398** | **0.402** | 0.290 |
| WINE-QUALITY | **0.224** | **0.203** | **0.224** | **0.203** | 0.075 | 0.106 | 0.042 | 0.064 | 0.124 | 0.102 | 0.051 | 0.059 |
| MUSHROOM-SUB | 0.811 | 0.677 | **0.940** | 0.892 | 0.636 | 0.725 | 0.331 | 0.758 | 0.479 | 0.596 | 0.839 | 0.546 |
| CAR | 0.050 | 0.061 | 0.043 | 0.067 | 0.045 | 0.044 | 0.032 | **0.071** | 0.050 | 0.030 | 0.034 | 0.041 |
| GERMAN-SUB | 0.066 | 0.077 | **0.106** | 0.098 | **0.113** | 0.065 | **0.103** | **0.104** | 0.062 | **0.118** | **0.109** | 0.079 |
| PNR | **0.190** | 0.172 | **0.190** | 0.172 | **0.201** | 0.059 | 0.107 | 0.100 | 0.106 | 0.006 | 0.146 | 0.124 |
| TRANSACTIONS | 0.756 | 0.752 | **0.810** | 0.835 | 0.509 | 0.563 | 0.510 | 0.532 | 0.760 | 0.373 | 0.585 | 0.564 |
| SHARED-ACCESS | 0.692 | 0.738 | 0.692 | 0.738 | 0.668 | 0.546 | **0.766** | 0.471 | 0.527 | 0.239 | **0.783** | 0.746 |
| PAYMENT-SUB | **0.173** | **0.173** | 0.168 | 0.168 | 0.137 | 0.157 | 0.129 | **0.175** | 0.143 | 0.101 | **0.180** | 0.142 |
| AIRLINE | **0.081** | **0.079** | **0.081** | **0.079** | 0.060 | 0.063 | 0.059 | 0.068 | 0.074 | 0.064 | - | 0.069 |
| AVERAGE[1] | 0.344 | 0.338 | 0.366 | 0.370 | 0.284 | 0.264 | 0.222 | 0.262 | 0.270 | 0.216 | 0.336 | 0.284 |

[1] AIRLINE was excluded from the average due to a missing value.

AE-G-2 on the 4 datasets containing categorical features. VAE-DGP-2 achieves good results but is outperformed on most small datasets.

VAE-1 also shows good outlier detection capabilities and handles binary features better than VAE-2. However, the multilayer architecture outperforms its shallow counterpart on large datasets containing more than 10,000 samples. Both algorithms perform better than NADE-2 which fails on high dimensional datasets such as MUSHROOM-SUB, PNR or TRANSACTIONS. We performed additional tests with an increased number of units for NADE-2 to cope for the large dimensionality, but we obtained similar results.

While AE-1 shows unexpected detection capabilities for a very simple model, AE-5 reaches the lowest performance. Compressing the data to a feature space 40% smaller than the input space along with dropout layers may cause loss of information resulting in an inaccurate model.

To assess the accuracy and the scalability of the selected neural networks, we measure the MAP and mean log-likelihood (MLL) on test data during the training phase to monitor their convergence. The evolution of the two metrics for the DNNs is reported in Figure 4.2.

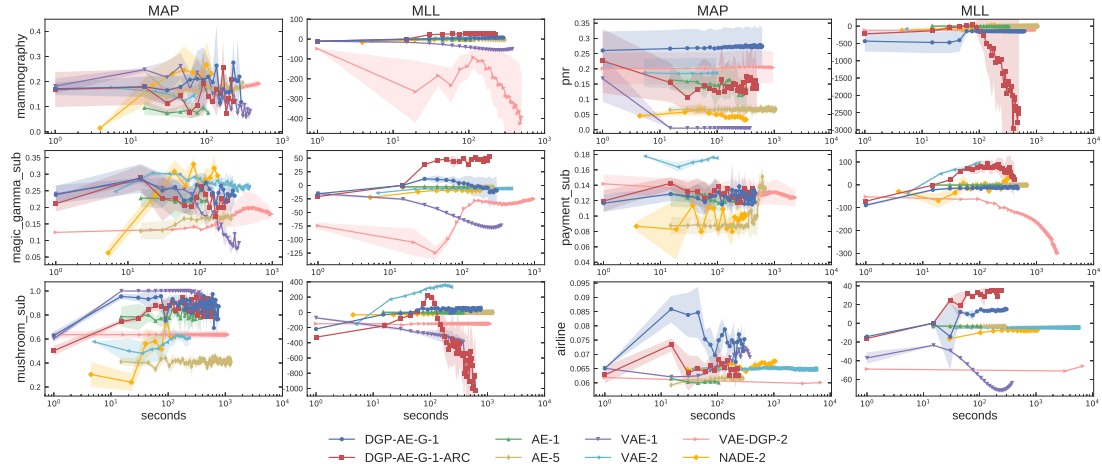

Figure 4.2: Evolution of the MAP and MLL over time for the selected networks. The metrics are computed on a 3-fold cross-validation on testing data. For both metrics, the higher values, the better the results.

While the likelihood is the objective function of most networks, the monitoring of this metric reveals occasional decreases of the MLL for all methods during the training process. If minor increases are part of the gradient optimization, the others indicate convergence issues for complex datasets. This is observed for VAE-DGP-2 and VAE-1 on MAMMOGRAPHY, or DGP-AE-G-1-ARC and VAE-1 on MUSHROOM-SUB.

Our DGPs show the best likelihood on most datasets, in particular when using the ARC kernel, with the exception of PNR and MUSHROOM-SUB where the RBF kernel is much more efficient. These results demonstrate the efficiency of regularization for DGPs and their excellent ability to generalize while fitting complex models.

On the opposite, NADE-2 barely reaches the likelihood of AE-1 and AE-5 at convergence. In addition, the network requires an extensive tuning of its parameters and has a computationally expensive prediction step. We tweaked the parameters to increase the model complexity, e.g. number of components and units, but it did not improve the optimized likelihood.

vae-dgp-2 does not reach a competitive likelihood, even with deeper architectures, and shows a computationally expensive prediction step.

Looking at the overall results of these networks, we observe that the model, depicted here by the likelihood, is refined during the entire training process, while the average precision quickly stabilizes. This behavior implies that the ordering of data points according to their outlier score converges much faster, even though small changes can still occur.

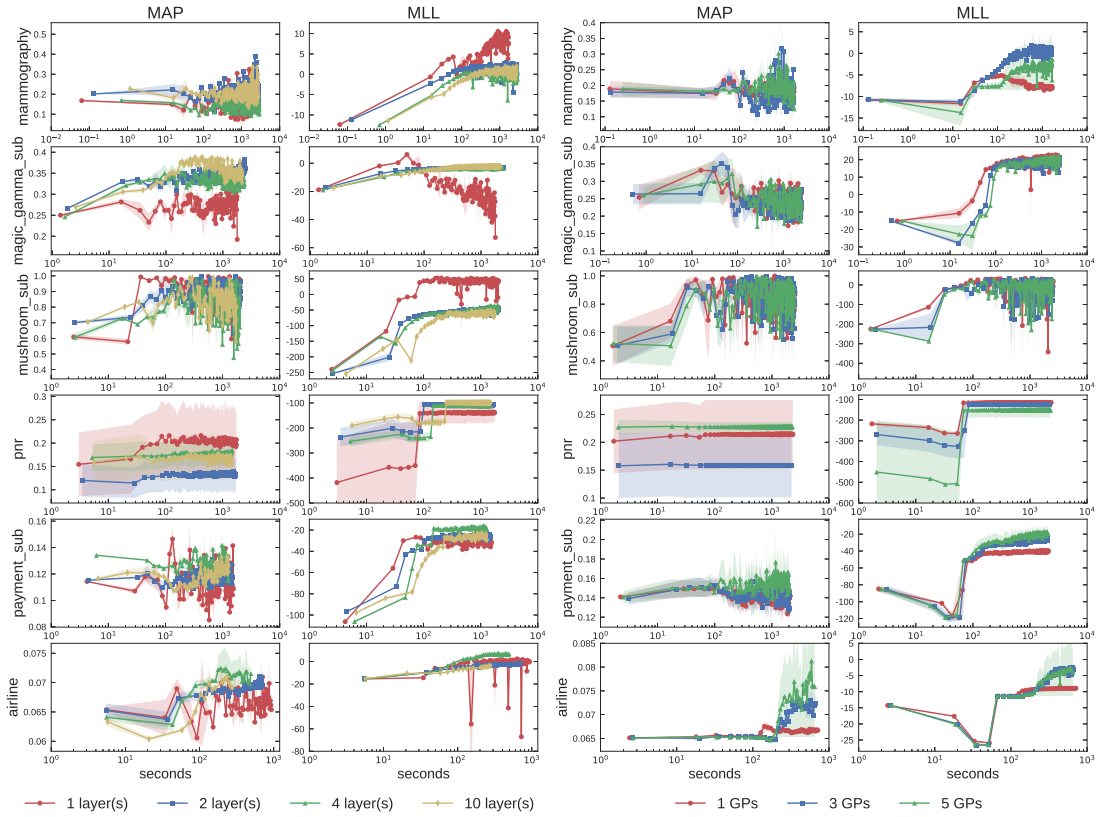

Figure 4.3: Evolution of the MAP and MLL over time on testing data based on a 3-fold cross-validation process. The left plot reports the metrics for DGP-AE-G with an increasing number of layers. For networks with more than 2 layers, we feed forward the input to the encoding layers, and feed forward the latent variables to the decoding layers. We use 3 GPs per layer and a length-scale of 1. The right plot shows the impact of an increasing number of GP nodes on a DGP-AE-G-2.

Additional convergence experiments have been performed on DGPs and are reported on Fig-ure 4.3. The left part of the figure shows the ability of DGP-AE-G to generalize while increasing the number of layers. On the right, we compare the dimensionality reduction capabilities of DGP-AE-G-2 while increasing the number of GPs on the latent variables layer.

The left part of the plot reports the convergence of DGP-AE-G for configurations ranging from one to ten layers. The plot highlights the correlation between a higher test likelihood and a higher average precision. Single-layer models show a good convergence of the MLL on most datasets, though are outperformed by deeper models, especially 4-layer networks, on MAGIC-GAMMA-SUB, PAYMENT-SUB and AIRLINE. Deep architectures result in models of higher capacity at the cost of needing larger datasets to be able to model complex representations, with a resulting slower convergence behavior. Using moderately deep networks can thus show better results on datasets where a single layer is not sufficient to capture the complexity of the data. Interestingly, the bound on the model evidence makes it possible to carry out model selection to decide on the best architecture for the model at hand [Cutajar et al., 2017].

In the right panel of Figure 4.3, we increase the dimensionality of the latent representation fixing the architecture to a DGP-AE-G-2. Both the test likelihood and the average precision show that a univariate GP is not sufficient to model accurately the input data. The limitations of this configuration is observed on MAMMOGRAPHY, PAYMENT-SUB and AIRLINE where more complex representations achieve better performance. Increasing the number of GPs results in a higher number of weights for the model, thus in a slower convergence. While configurations using 5 GPs already perform a significant dimensionality reduction, they achieve good performance and are suitable for efficient novelty detection.

### Latent representation

In this section we illustrate the capabilities of the proposed DGP-AE model to construct meaningful latent probabilistic representations of the data. We select a two-layer DGP-AE architecture with a two-dimensional latent representation $Z := F^{(1)}$. Since the mapping of the DGP-AE model is probabilistic, each input point is mapped into a cloud of latent variables. In order to obtain a generative model, we could then train a density estimation algorithm on the latent variables to construct a density $q(\mathbf{z})$ used together with the probabilistic decoder part of the DGP-AE to generate new observations.

In Figure 4.4, we draw 300 Monte Carlo samples from the approximate posterior over the weights $\mathbf{W}$ to construct a latent representation of the OLD FAITHFUL dataset. We use a GMM with two components to cluster the input data, and color the latent representation based on the resulting labels. The point highlighted on the left panel of the plot by a cross is mapped

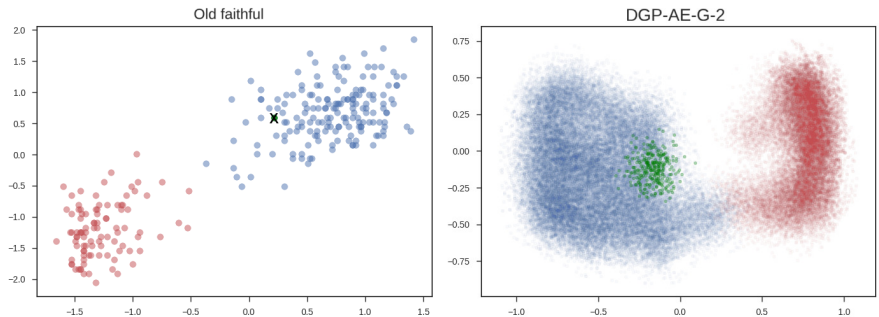

Figure 4.4: Left: normalized OLD FAITHFUL dataset. Right: latent representation of the dataset for a 2-layer DGP-AE (100,000 iterations, 300 Monte Carlo samples).

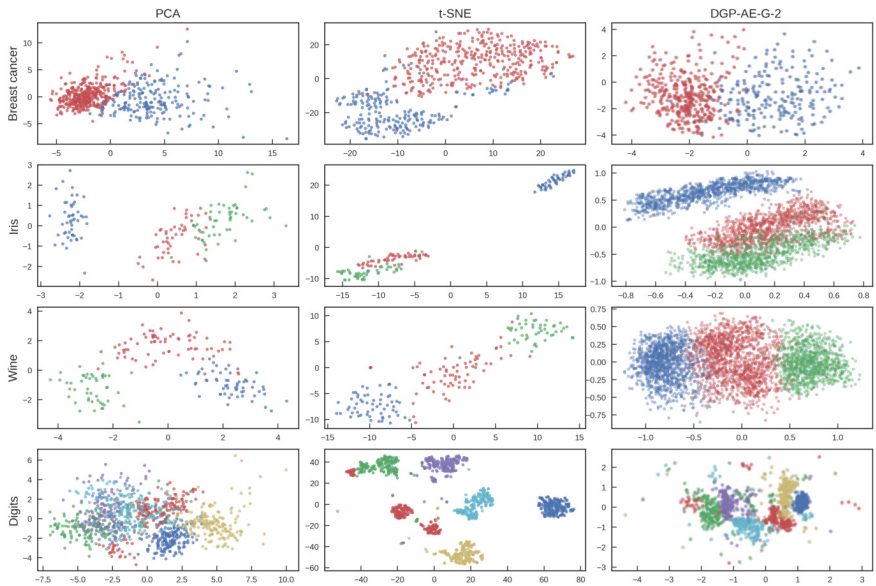

Figure 4.5: Dimensionality reduction performed on 4 classification datasets. DGP-AE-G-2 was trained for 100,000 iterations, and used 20 Monte Carlo iterations to sample the latent variables.

into the green points on the right.

We now extend our experiment to labelled datasets of higher dimensionality, using the given labels for the sole purpose of assigning a color to the points in the latent space. Figure 4.5 shows the two-dimensional representation of four datasets, BREAST CANCER (569 samples, 30 features), IRIS (150x4), WINE (178x13) and DIGITS (1797x64). For comparison, we also re-port the results of two manifold learning algorithms, namely t-sne [Maaten & Hinton, 2008] and Probabilistic pca [Tipping & Bishop, 1999]. The plot shows that our algorithm yields meaningful low-dimensional representations, comparable with state-of-the-art dimensionality reduction methods.

## 4.3 Summary

In this chapter, we introduced a novel deep probabilistic model for novelty detection. The proposed dgp-ae model is an autoencoder where the encoding and the decoding mappings are governed by dgps. We make the inference of the model tractable and scalable by approximating the dgps using random feature expansions and by inferring the resulting model through stochastic variational inference that could exploit distributed and GPU computing. The proposed dgp-ae is able to flexibly model data with mixed-types feature, which is actively investigated in the recent literature [Vergari et al., 2018]. Furthermore, the model is easy to implement using automatic differentiation tools, and is characterized by robust training given that, unlike most gp-based models [Dai et al., 2016], it only involves tensor products and no matrix factorizations.

Through a series of experiments, we demonstrated that dgp-aes achieve competitive results against state-of-the-art novelty detection methods and dnn-based novelty detection methods. Crucially, dgp-aes achieve these performance with a practical learning method, making deep probabilistic modeling as an attractive model for general novelty detection tasks. Even though we leave this for future work, dgp-aes can easily include the use of special representations based, e.g., on convolutional filters for applications involving images, and allow for end-to-end training of the model and the filters. The encoded latent representation is probabilistic and it yields uncertainty that can be used to turn the proposed autoencoder into a generative model; we also leave this investigation for future work, as well as the possibility to make use of dgps to model the mappings in variational autoencoders.

*To understand is to perceive patterns.*

Isaiah Berlin

# 5

# Comparative evaluation of novelty detection methods for discrete sequences

This chapter surveys the problem of detecting anomalies in temporal data, specifically in discrete sequences of events which have a temporal order. Such a problem can be divided into two categories. The first one is *change point detection*, where datasets are long sequences in which we seek anomalous and contiguous subsequences, denoting a sudden change of behavior in the data. Use cases relating to this problem are sensor readings [Kundzewicz & Robson, 2004] and first story detection [Petrović et al., 2010]. A second category considers datasets as sets of sequences, and targets the detection of anomalous sequences with respect to nominal samples. Our study focuses on the latter, which encompasses use cases such as protein identification for genomics [Chandola et al., 2008, Sun et al., 2006], fraud and intrusion detection [Maxion & Townsend, 2002, Warrender et al., 1999, Chandola et al., 2008] and user behavior analysis (UBA) [Sculley & Brodley, 2006].

While this is a matter of interest in the literature, most reviews addressing the issue focus on theoretical aspects [Gupta et al., 2014, Chandola et al., 2012], and as such do not assess and compare performance. Chandola et al. [Chandola et al., 2008] showcase an experimental comparison of novelty detection methods for sequential data, although this work uses a custom metric to measure the novelty detection capabilities of the algorithms and misses methods which have been recently published in the field. Our work extends previous studies by bring-ing together the following contributions: (i) comparison of the novelty detection performance for 12 algorithms, including recent developments in neural networks, on 81 datasets containing discrete sequences from a variety of research fields; (ii) assessment of the robustness for the selected methods using datasets contaminated by outliers, with contrast to previous studies which rely on clean training data; (iii) scalability measurements for each algorithm, reporting the training and prediction time, memory usage and novelty detection capabilities on synthetic datasets of increasing samples, sequence length and anomalies; (iv) discussion on the interpretability of the different approaches, in order to provide insights and motivate the predictions resulting from the trained model. To our knowledge, this study is the first to perform an evaluation of novelty detection methods for discrete sequences with so many datasets and algorithms. This work is also the first to assess the scalability of the selected methods, which is an important selection criterion for processes subject to fast response time commitments, in addition to resource-constrained systems such as embedded systems.

The state-of-the-art methods evaluated in this chapter are detailed in Section 2.2 and reminded in Section 5.1.1. The chapter is organized as follows: Section 5.1 details the real-world and synthetic datasets used for the study, in addition to the methods' parameters, Sections 5.2 and 5.3 report the results and conclusions of the work.

## 5.1 Experimental setup

### 5.1.1 Algorithms

The methods evaluated in this work satisfy the following set of constraints, inherent to most problems from the field. The methods are trained on a dataset composed of discrete sequences of events, and are compatible with sequences of variable length; The algorithms also provide a prediction step performing a continuous scoring of unseen sequences, and support new events which were not part of the training set.

HMM and ISM are probabilistic and generative approaches including a dedicated support for variable length. $k$-NN, LOF and $k$-MEDOIDS feed on a pairwise distance matrix based on the LCS or Levenshtein distance. These metrics are normalized and suitable to compare sequences of different length. As an additional functionality, $k$-MEDOIDS provides a clustering of the input data. In order to avoid dealing with padding, $t$-STIDE and RIPPER transform each input sequence into a set of subsequences having the same length. The study includes two neural networks using LSTM cells, named LSTM-AE and SEQ2SEQ, which are trained on mini-batches of padded sequences and use a masking mechanism.

Table 5.1: Parameters and implementations of the selected algorithms

| Algorithm | Language | Parameters |
|---|---|---|
| HMM [1] | Python | $components = 3, iters = 30, tol = 10^{-2}$ |
| LCS | Python | n/a |
| Levenshtein | Python | n/a |
| $k$-NN | Python | $k = \max(n * 0.1, 20)$ |
| LOF | Python | $k = \max(n * 0.1, 50)$ |
| $k$-MEDOIDS | Python | $k = 2$ |
| $t$-STIDE [2] | C | $k = 6, t = 10^{-5}$ |
| RIPPER [2] | R | $K = 9, F = 2, N = 1, O = 2$ |
| ISM | Java | $iters = 100, s = 10^5$ |
| SEQ2SEQ [3] | Python | $iters = 100, batch = 128, hidden = 40, enc\_dropout = 0.5, dec\_dropout = 0.$ |
| LSTM-AE [3] | Python | $batch = 128, iters = 50, hidden = 40, \delta = 10^{-4}$ |

[1] New symbols are not supported natively by the method.
[2] Sequences were split into sliding windows of fixed length.
[3] Padding symbols were added to the datasets to provide batches of sequences having the same length.

The implementation and configuration of the methods are detailed in Table 5.1. Parameter selection was achieved by grid-search and maximizes the MAP averaged over the testing datasets detailed in Section 5.1.2. We use rpy2 to run algorithms written in R from Python, and create dedicated subprocesses for Java and C.

### 5.1.2  PERFORMANCE TESTS

Our evaluation uses 81 datasets related to genomics, intrusion detection and user behavior analysis (UBA). The datasets are divided into 9 categories detailed in Table 5.2, and cover a total of 68,832 sequences. For a given dataset, we use 70% of the data for the training, and 30% for the testing.

The metric used to evaluate the novelty detection capabilities of the methods is the average precision (AP) computed over the testing data and detailed in Chapter 1. To ensure stability and confidence in our results, we perform 5-fold cross-validation for each method and dataset. The final performance given in Table 5.3 is thus the *mean average precision* (MAP), i.e. the AP averaged over the 5 iterations. A *robust* method is able to learn a consistent model from noisy data, i.e. a training set contaminated by anomalies. We use the same proportion of outliers in the training and testing sets to showcase the robustness of the selected methods.

Table 5.2: Datasets benchmarked, related to genomics (GEN), intrusion detection (INT) or user behavior analysis (GEN). $D$ is the number of datasets in each collection. The following characteristics are averaged over the collection of datasets: $N$ is the number of samples, $A$ and $p_A$ are the number and proportion of anomalies, respectively, $M_L$ is the length of the shortest sequence, $\mu_L$ is the average sequence length, $S_L$ is the entropy of the sequence lengths, $\sigma$ is the number of unique events, $S_\sigma$ is the entropy of the event distribution, $T_5$ (Top 5%) is the proportion of events represented by the 5% biggest events and $L_1$ (Lowest 1%) is the proportion of the smallest events representing 1% of the events.

| Category | Area | D | N | A ($p_A$) | $M_L$ | $\mu_L$ | $S_L$ | $\sigma$ | $S_\sigma$ | $T_5$ | $L_1$ |
|---|---|---|---|---|---|---|---|---|---|---|---|
| SPLICE-JUNCTIONS | GEN | 1 | 1710 | 55 (3.22%) | 60 | 60 | 0.00 | 6 | 1.39 | 25.76 | 16.67 |
| PROMOTER | GEN | 1 | 59 | 6 (10.17%) | 57 | 57 | 0.00 | 4 | 1.39 | 26.85 | 0.00 |
| PFAM | GEN | 5 | 5166 | 165 (3.19%) | 117 | 1034 | 0.15 | 45 | 1.17 | 83.97 | 40.00 |
| MASQUERADE | INT | 29 | 94 | 6 (6.29%) | 100 | 100 | 0.00 | 113 | 3.40 | 49.69 | 29.55 |
| INTRUSIONS | INT | 6 | 2834 | 202 (7.14%) | 56 | 1310 | 4.27 | 43 | 2.01 | 66.91 | 36.43 |
| UNIX | UBA | 9 | 1045 | 33 (3.20%) | 1 | 31 | 3.60 | 379 | 3.31 | 77.54 | 48.86 |
| RIGHTS | UBA | 10 | 677 | 22 (3.18%) | 1 | 15 | 3.31 | 67 | 2.19 | 70.03 | 55.95 |
| TRANSACTIONS-FR | UBA | 10 | 215 | 7 (3.21%) | 4 | 49 | 3.57 | 285 | 4.16 | 47.57 | 33.37 |
| TRANSACTIONS-MO | UBA | 10 | 386 | 12 (3.19%) | 5 | 37 | 3.88 | 416 | 4.18 | 67.08 | 33.46 |

The corpus of data described in Table 5.2 includes 6 widely used public collections of datasets, in addition to 3 new collections of industrial datasets from the company Amadeus. PFAM (v31.0) describes 5 families of proteins, namely RUB (PF00301), TET (PF00335), SNO (PF01174), NAD (PF02540) and RVP (PF08284). INTRUSIONS contains UNIX system calls for the traces LPR-MIT, LPR-UNM, SENDMAIL-CERT, SENDMAIL-UNM, STIDE and XLOCK. Concerning industrial datasets, RIGHTS details the actions performed by users in a Web application designed to manage the permissions of users and roles. The dataset shows the sessions of the 10 most active users. For each user dataset, anomalies are introduced by sampling sessions from the 9 remaining users. TRANSACTIONS-FR and TRANSACTIONS-MO are generated from a business-oriented flight booking application and covers Web traffic coming from France and Morocco. User selection and anomaly generation were performed as described previously.

### 5.1.3 SCALABILITY TESTS

Synthetic datasets are generated to measure the scalability of the selected methods. Nominal data is obtained by sampling $N$ sequences of fixed length $L$ from a Markov chain. The transition matrix used by the Markov chain is randomly generated from a uniform distribution and has dimension $\sigma$, where $\sigma$ is the size of the alphabet. Anomalies are sampled from a distinct random transition matrix of same dimension, to which we add the identity matrix. The default proportion of anomalies in the training and testing sets is 10%. Both transition matrices are normalized to provide correct categorical distributions.

We vary $N$, $L$ and the proportion of anomalies to generate datasets of increasing size and complexity. We also studied the impact of $\sigma$ on the methods, and found that it had little effect on the scalability and MAP. The training time, prediction time, memory usage and novelty detection abilities of the algorithms are measured during this process. For each configuration, we run the algorithms 3 times over distinct sampled datasets and average the metrics to increase confidence in our results. Training and testing datasets are generated from the same two transition matrices, and have the same number of samples and outliers.

The experiments are performed on a VMWare platform running Ubuntu 14.04 LTS and powered by an Intel Xeon E5-4627 v4 CPU (10 cores at 2.6 GHz) and 256GB RAM. We use the Intel distribution of Python 3.5.2, Java 8 and R 3.3.2. Due to the number of algorithms and the size of the datasets, we interrupt training and scoring steps lasting more than 12 hours. Memory usage is measured by memory_profiler for algorithms written in Python and R, and by the UNIX *ps* command for other languages. We perform a garbage collection for R and Python before starting the corresponding methods. Memory consumption is measured at intervals of $10^{-4}$ seconds, and shows the maximum usage observed during the training or scoring step. The memory required by the plain running environment and to store the dataset is subtracted to the observed memory peak.

## 5.2 RESULTS

### 5.2.1 NOVELTY DETECTION CAPABILITIES

The mean average precision (MAP) resulting from the experiment detailed in Section 5.1.2 is reported in Table 5.3 for each algorithm and dataset. When no significant difference can be observed between a given MAP and the best result achieved on the dataset, we highlight the corresponding MAP in bold. The null hypothesis is rejected based on a pairwise Friedman

Table 5.3: Mean area under the precision-recall curve (MAP) averaged per group of datasets over 5 cross-validation iterations. Results in bold indicate that we cannot reject the null hypothesis of the given MAP to be identical to the best MAP achieved for the dataset. Column *Rank* reports the aggregated rank for each method based on the Spearman footrule distance.

| | SPLICE | PROMOT. | PFAM | MASQUE. | INTRUS. | UNIX | RIGHTS | TRANS-FR | TRANS-MO | **Mean** | **Rank** |
|---|---|---|---|---|---|---|---|---|---|---|---|
| HMM | 0.027 | 0.336 | 0.387 | **0.166** | **0.580** | 0.302 | **0.246** | **0.260** | 0.164 | **0.274** | 1 |
| *k*-NN-LCS | 0.032 | **0.437** | **0.516** | 0.132 | **0.425** | 0.207 | **0.270** | 0.179 | **0.097** | **0.255** | 3 |
| *k*-NN-LEV | 0.033 | **0.412** | **0.516** | 0.129 | **0.405** | 0.120 | **0.188** | 0.185 | 0.083 | **0.230** | 5 |
| LOF-LCS | **0.042** | 0.150 | 0.029 | **0.167** | 0.141 | 0.073 | 0.042 | 0.091 | 0.041 | **0.086** | 12 |
| LOF-LEV | 0.031 | 0.226 | **0.517** | **0.156** | 0.181 | **0.132** | **0.191** | **0.192** | **0.099** | 0.192 | 4 |
| *k*-MEDOIDS-LCS | 0.027 | **0.581** | **0.510** | 0.134 | 0.318 | **0.155** | **0.218** | **0.184** | 0.092 | **0.247** | 6 |
| *k*-MEDOIDS-LEV | **0.040** | **0.692** | **0.513** | **0.148** | 0.222 | 0.086 | 0.146 | **0.189** | 0.078 | **0.235** | 7 |
| *t*-STIDE | **0.048** | **0.806** | **0.506** | 0.122 | **0.469** | 0.081 | 0.130 | 0.136 | **0.112** | **0.268** | 9 |
| RIPPER | 0.028 | **0.431** | 0.034 | **0.176** | **0.359** | 0.053 | 0.077 | 0.105 | 0.079 | **0.149** | 10 |
| ISM | 0.027 | 0.205 | 0.116 | 0.140 | **0.559** | **0.220** | **0.217** | **0.211** | **0.111** | 0.201 | 2 |
| SEQ2SEQ | **0.072** | 0.341 | 0.035 | **0.178** | 0.113 | 0.076 | 0.083 | 0.092 | 0.063 | **0.117** | 11 |
| LSTM-AE | 0.034 | **0.494** | **0.591** | **0.178** | 0.174 | 0.074 | 0.100 | **0.173** | 0.075 | **0.210** | 8 |

test [García et al., 2010] with a significance level of 0.05.

While we believe that no method outperforms all others, and that each problem may require a distinct method, we attempt to give a broad overview of how methods compare to one another. For this purpose, we extract the rank of each algorithm on each collection of datasets from Table 5.3 and aggregate them to produce an overall ranking reported in the last column. The aggregation is performed using the Cross-Entropy Monte Carlo algorithm [Pihur et al., 2009] and rely on the Spearman distance.

In order to infer the behavior of each method based on the datasets characteristics, we learn an interpretable meta-model using the features introduced in Table 5.2. While the metrics given in Table 5.2 are computed over entire datasets, then averaged over the corresponding collection, this experiment focuses on the training data and retains features for each of the 81 datasets. We use these features as input data, and fit one decision tree per algorithm in order to predict how a given method performs. The resulting models are binary classifiers where the target class is whether the average rank of the algorithm is among the top 25% performers (ranks 1 to 3), or if it reaches the lowest 25% (ranks 9 to 12). Figure 5.1 shows the trained meta-model of *k*-MEDOIDS-LEV as an example. These trees expose the strengths and weaknesses of the methods studied, and highlight the most important factors impacting the methods' performances.

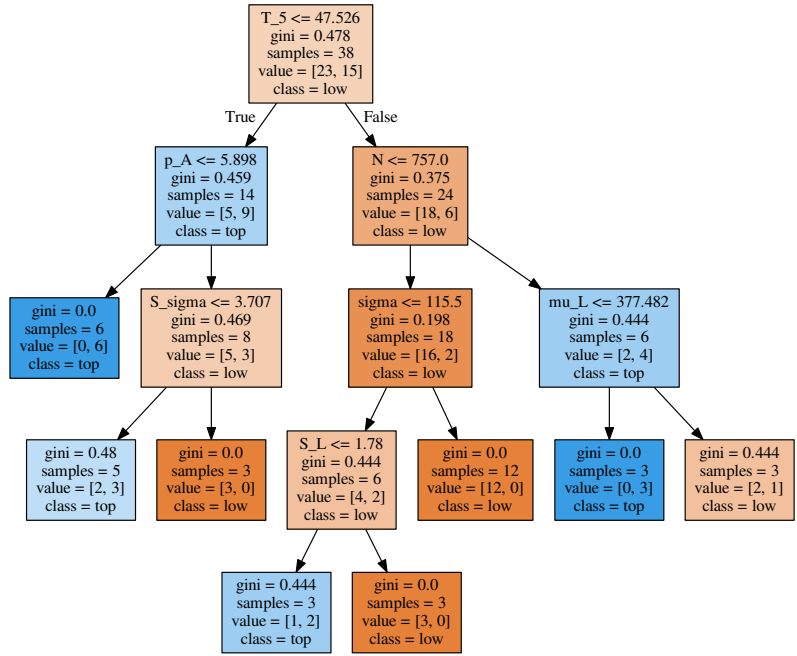

Figure 5.1: Novelty detection capability decision tree for *k*-MEDOIDS-LEV. This tree highlights position of *k*-MEDOIDS-LEV in the overall ranking based on features extracted from the datasets. Ranks have been aggregated into the *top* and *low* classes which encompass the best (1 to 3) and worst (10 to 12) 25% ranks, respectively.

In order to provide a concise visual overview of this analysis, we report in Figure 5.2 the performance of each method based on the datasets characteristics. For this purpose, we extract the rules of the nodes for which $depth < 4$ in all meta-models, then aggregate these rules per feature to identify values corresponding to the most important splits. The resulting filters are reported in the horizontal axis of the heatmap.

Our experiments show that no algorithm consistently reaches better results than the competing methods, but that HMM, *k*-NN and ISM are promising novelty detection methods. While previous comparisons [Warrender et al., 1999, Chandola et al., 2008, Budalakoti et al., 2009] use clean datasets exempt of anomalies, our study shows a good robustness for the selected methods, even for datasets with a high proportion of outliers, namely PROMOTER, MASQUERADE and INTRUSIONS.

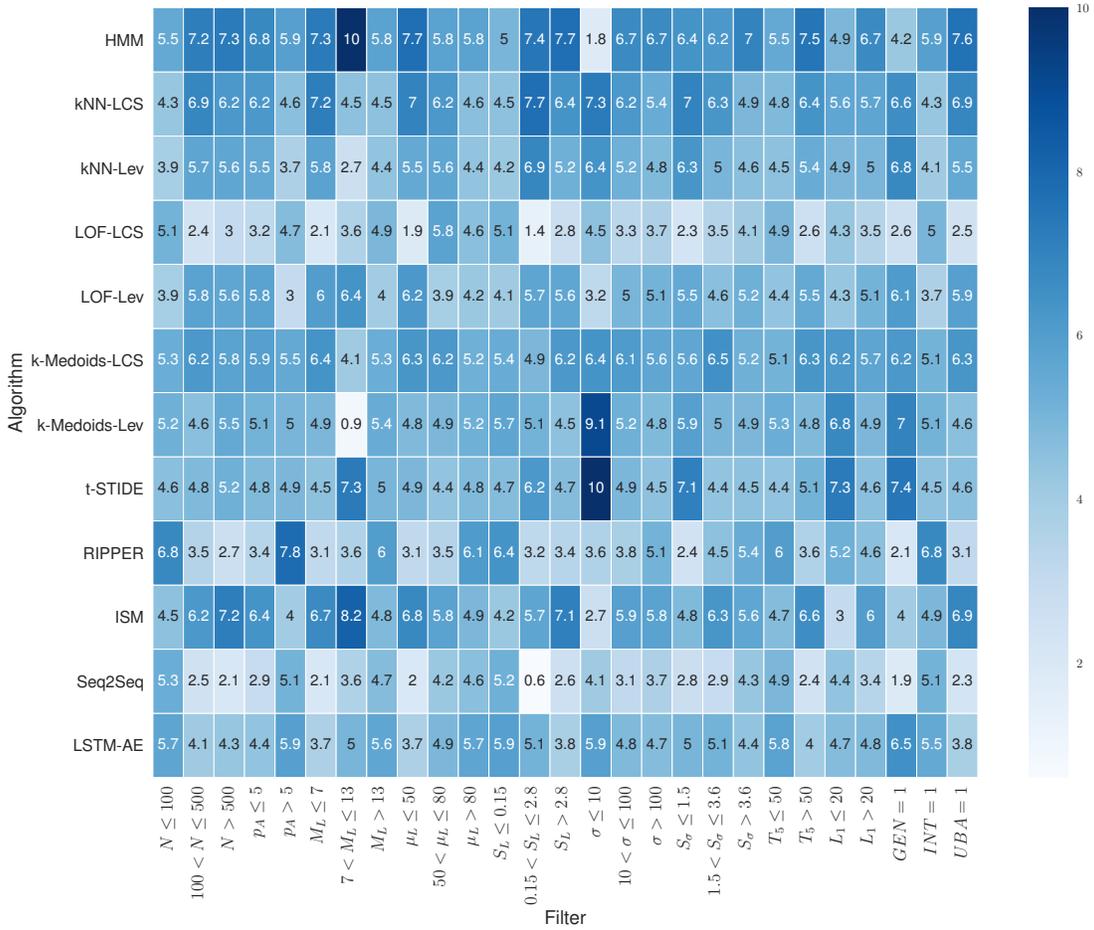

Figure 5.2: Novelty detection capability heatmap. The plot highlights the performance of each algorithm based on the datasets characteristics. Scores range from 0 to 10 and are based on the rank of the method averaged over the subset of datasets matching the corresponding filter applied to the 81 datasets. A score of 10 corresponds to an average rank of 1, while a score of 0 indicates that the method consistently ended in the last position. $N$ is the number of samples; $p_A$ is the proportion of anomalies; $M_L$, $\mu_L$ and $S_L$ are the minimum, average and entropy computed over the sequence length; $\sigma$ and $S_\sigma$ are the alphabet size and the corresponding entropy of its distribution, the entropy increasing with the number of events and the distribution uniformity; $T_5$ is the proportion of events represented by the 5% biggest events, a high value denotes important inequalities in the distribution; $L_1$ is the proportion of the smallest events representing 1% of the data, a high value indicates numerous events with rare occurrences; the genomics (GEN), intrusion detection (INT) and UBA columns target datasets related to the corresponding field of study.

Concerning the applications studied, $k$-NN, $k$-MEDOIDS, $t$-STIDE and LSTM-AE show good performance on datasets related to genomics, which are SPLICE-JUNCTIONS, PROMOTER and PFAM. $t$-STIDE apart, these methods have successfully addressed numerous supervised numerical problems, and could thus reach good performance when applied to sequence-based supervised use cases. The best methods for intrusion detection are HMM and RIPPER, while $t$-STIDE shows reduced performance compared to [Warrender et al., 1999], likely caused by the introduction of anomalies in the training sets. Our observations for genomics and intrusion detection corroborate the conclusions presented for $t$-STIDE and RIPPER in [Chandola et al., 2008]. However, our study shows much better performance for HMM, the previous study using a custom likelihood for HMM based on an aggregated sequence of binary scores. With regard to user behavior analysis, HMM, $k$-NN, $k$-MEDOIDS-LCS and ISM show the best ability to differentiate users. While the performance of $t$-STIDE on UBA are not sufficient to recommend the method, we believe that increasing the threshold of $t$-STIDE would lead to increased performance. Indeed, user actions are often based on well-defined application flows, and most of the possible subsequences are likely to exist in the training sets. The amount of supplementary information which can be provided by the models about the user behaviors will determine the most suitable methods for this field (Section 5.2.5).

Figure 5.2 shows that the performance of HMM improves significantly with the number of available samples. Both HMM and ISM achieve good performance, even when a high discrepancy is observed among the sequence lengths. HMM, ISM and RIPPER are able to handle efficiently a large alphabet of symbols. RIPPER also shows good performance for datasets containing a high proportion of outliers, while nearest neighbor methods are strongly impacted by this characteristic. Distance metrics are known to suffer from the curse of dimensionality inherent to a high number of features. Similarly, Figure 5.2 shows a decrease of performance for $k$-NN, $k$-MEDOIDS and LOF when $\sigma$ increases, these methods relying on the LCS and Levenshtein metrics for distance computations. While LCS is a metric widely used in the literature [Chandola et al., 2008, Budalakoti et al., 2009, Budalakoti et al., 2006], our experiments show that it does not perform better than the Levenshtein distance. If both metrics provide satisfactory results for novelty detection, the combination of LOF and LCS produces the lowest accuracy of our evaluation. Nonetheless, the efficiency of LOF-LEV prevents us from discarding this method, even though $k$-NN-LEV achieves a similar accuracy to LOF-LEV with a simpler scoring function. For the sake of the experiment, we evaluated the scoring function proposed for *t*-stide in [Hofmeyr et al., 1998]. For each subsequence of fixed length in a test sequence, the authors compute the hamming distance between the test window and all training windows, and return the shortest distance. This method was much slower than a binary decision based on the presence of the test window in the training set, and did not strongly improve the results. Neural networks do not stand out in this test. The reconstruction error showed good results for detecting numerical anomalies in previous studies [Sakurada & Yairi, 2014, Marchi et al., 2015], but the approach may not be appropriate for event sequences. The reconstructed sequences provided by seq2seq are often longer than the input data, and the network loops regularly for a while over a given event. Figure 5.2 show that lstm networks perform better with long sequences and a moderate alphabet size. We repeated our experiments using the Python library difflib as an alternative to lcs for seq2seq, but it did not improve the performance of the network. lstm-ae shows a correct novelty detection accuracy, which could be further improved with dropout and attention. Thanks to their moderate depth, these two networks do not require very large datasets to tune their parameters. For example, lstm-ae achieves a good map even for small datasets such as promoter and masquerade. Despite the use of masks to address padding, these methods have difficulty with datasets showing an important disparity in sequence length, such as intrusions and the four collections of uba datasets.

### 5.2.2 Robustness

Figures 5.3 to 5.5 report the mean area under the precision recall curve (map) for datasets of increasing proportion of outliers, number of samples and sequence length, respectively. The positive class represents the nominal samples in Figure 5.3, and the anomalies in Figure 5.4 and 5.5 (as in Section 5.2.1).

Figure 5.3 demonstrates a more complex test case than just identifying uniform background noise against a well-defined distribution. In this test, anomalies are sampled according to their own probability distribution, which will affect the models learnt when a sufficient proportion of anomalies is reached. The test highlights thus how algorithms deal with complex data based on multiple distributions. We observe that most algorithms focus on the major distribution as long as the proportion of corresponding samples remains higher than 60%. hmm uses 3 components and may thus learn the second distribution much earlier in the test. On the opposite, most of the distance-based methods discard the smallest distribution even if this one represents up to 40% of the data. lof-lcs shows poor performance from the very beginning,

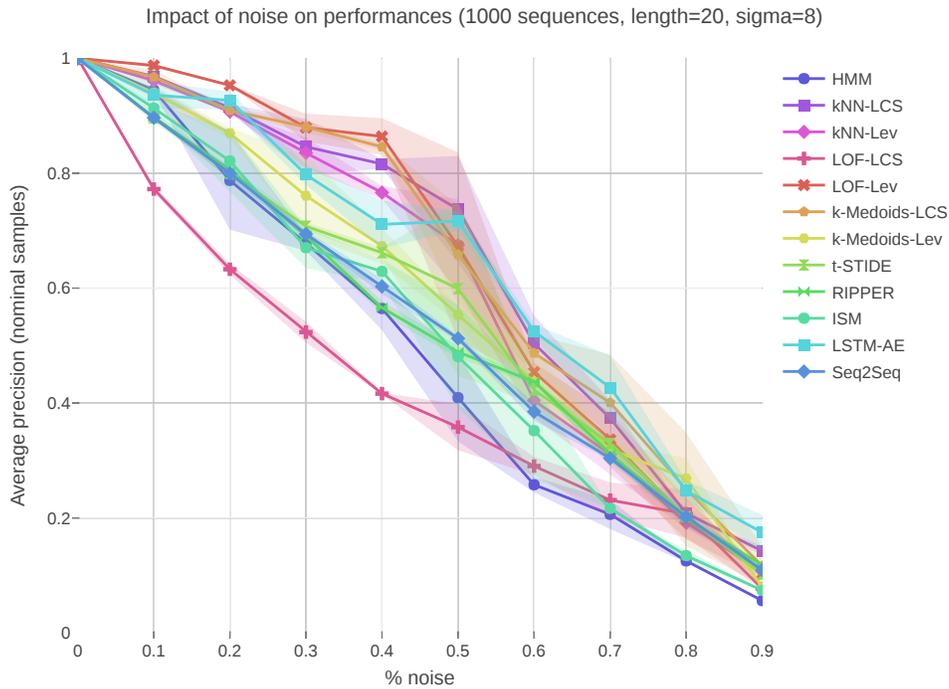

Figure 5.3: Robustness for increasing noise density

which prevents us from concluding on the behavior of this method.

Figure 5.4 shows that 200 samples are a good basis to reach stable novelty detection results. While we expected the performance of deep learning methods to improve with the number of samples, these networks did not significantly increase their detection with the size of the dataset. The best results on large datasets were achieved by distance-based methods, most of which rely on nearest-neighbor approaches particularly efficient when a high number of samples is available. Good performance were also achieved by ʜᴍᴍ, presumably due to a generation method for nominal samples and outliers based on Markov chains, which matches the internal representation of ʜᴍᴍ.

Despite the increasing volume of data over the scalability test reported in Figure 5.5, important variations can be observed for the results, possibly related to the limited number of samples available for the generated datasets. *k*-ᴍᴇᴅᴏɪᴅs achieve better performance than other distance-based methods, which suggests a better approach for small datasets. ʜᴍᴍ achieves once again good results, while ʟsᴛᴍ networks show improved novelty detection capabilities for datasets containing sequences longer than 100 events. The performance of ɪsᴍ also increases

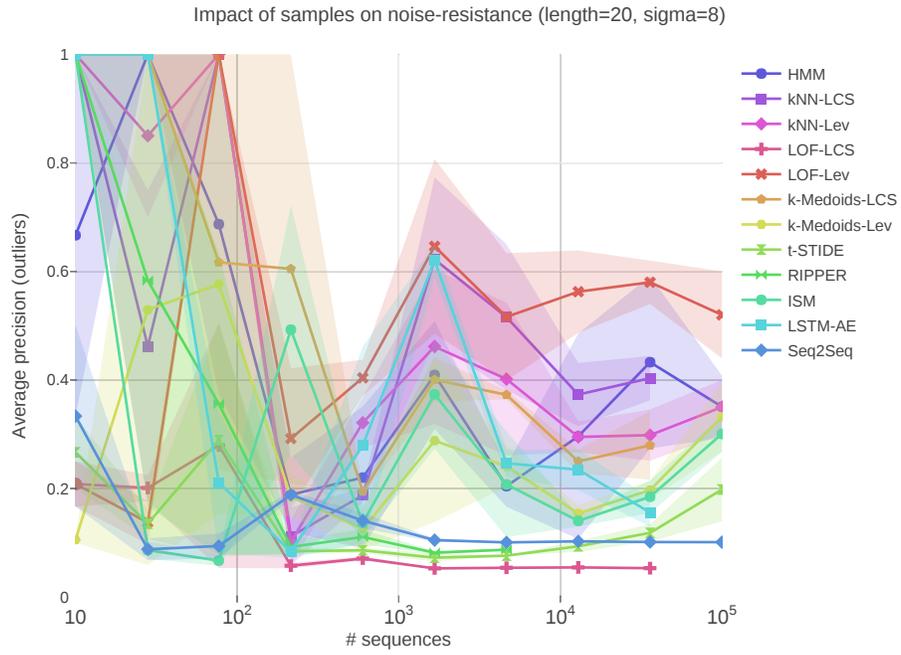

Figure 5.4: Robustness for increasing number of samples

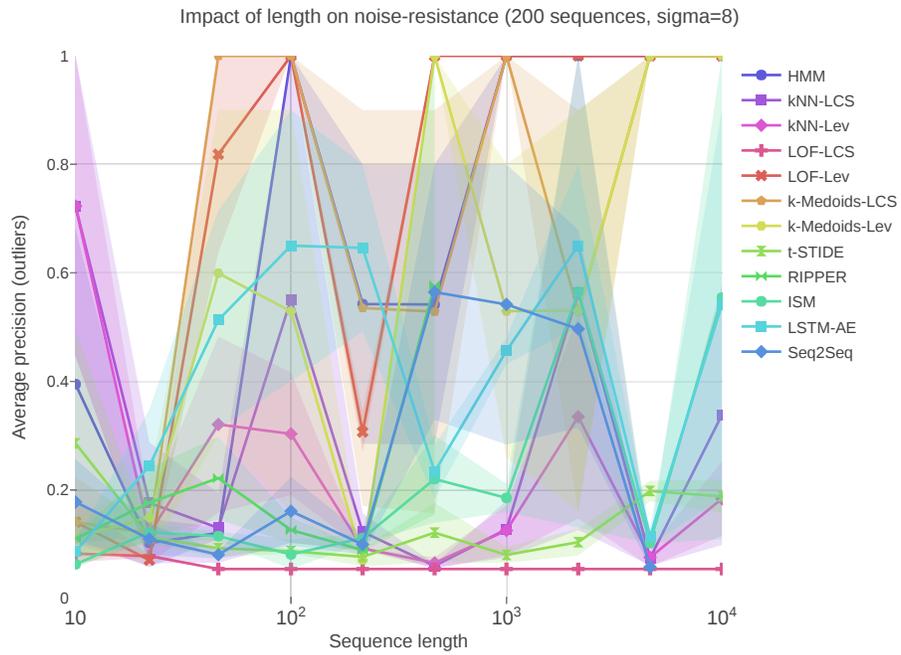

Figure 5.5: Robustness for increasing sequence length

with the volume of data, although the method require bigger datasets to reach comparable results.

In summary, our experiments show that robust models require at least 200 training samples to provide satisfactory results. LOF-LCS and $t$-STIDE do not provide satisfactory performance, even though fine-tuning $t$-STIDE by increasing the frequency threshold could lead to better results.

### 5.2.3 COMPLEXITY

The computation time for training and prediction steps is reported in figures 5.6 to 5.9. While time measurements are impacted by hardware configuration (Sec. 5.1.3), the slope of the curves and their ranking compared to other methods should remain the same for most running environments.

The measurements from Figures 5.6 and 5.7 show a poor scalability of algorithms relying on pairwise distance matrices, namely LOF, $k$-NN and $k$-MEDOIDS. Most of the training and prediction time of these methods is dedicated to the computation of the distance matrix, and thus to the LCS and Levenshtein algorithms. Since training and testing sets have the same number of samples in this test, the previous assumption is confirmed by observing a similar training and prediction time for the methods. In addition, $k$-MEDOIDS is the only distance-based algorithm with a faster prediction time, caused by a smaller number of distances to compute. The prediction step of this method requires only to compare a small number of medoids with the testing set, instead of performing a heavy pairwise comparison. Regarding distance metrics, LCS shows a much higher computation time than the Levenshtein distance. Despite a very small $\sigma$, the rule-learning algorithm RIPPER shows the highest training time, reaching our 12-hour timeout for 13,000 samples. On the opposite and as expected, the use of mini-batch learning by LSTM-AE and SEQ2SEQ allows the two methods to efficiently handle the increasing number of sequences, although we recommend to increase the batch size or the number of iterations according to the size of the training set. However, such technique is only valid for the training step, and both methods show a scoring scalability comparable to the other algorithms. The extreme simplicity of $t$-STIDE, which essentially stores subsequences in a dictionary at train time, makes this algorithm one of the fastest methods. The increasing load does not affect much ISM, since the method stops iterating over the dataset if it does not find new interesting patterns after a given number of sequences.

We now use a fixed number of samples while increasing the length of the sequences and report the computation time in Figures 5.8 and 5.9. The careful reader will notice that both scalability tests, i.e. number of sequence-based and length-based, produce datasets containing the exact same number of symbols (e.g. $10^5\ sequences * 20\ symbols = 200\ sequences * 10^4\ symbols$). This configuration reveals the true impact of samples and length on the scalability, while keeping the same volume of data. While we still observe a poor scalability for distance-based algorithms caused by a high computation time to compute distances, the training and prediction time of these methods was reduced due to a smaller number of samples to handle by the core algorithm. On the opposite, RIPPER and ISM show a much higher training time when dealing with long sequences. However, the prediction time of these two methods only depends on the volume of data, i.e. the total number of symbols in the dataset, and will be impacted similarly by the number of samples and length. Mini-batch methods are now subject to training batches of increasing volume, which reveals a poor scalability for SEQ2SEQ. LSTM-AE performs better due to an early stopping mechanism, interrupting the training when the loss does not improve sufficiently over the iterations.

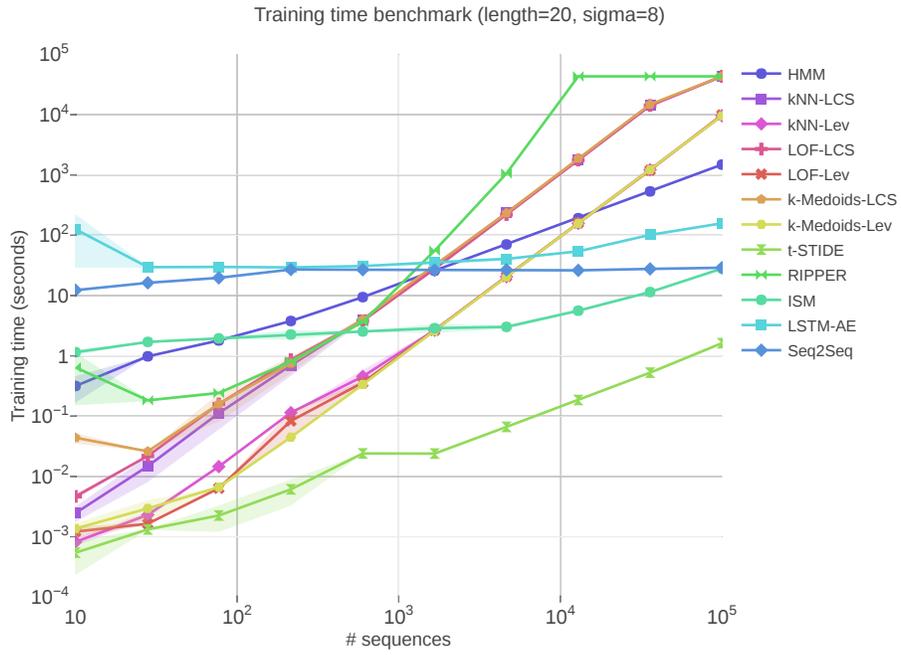

Figure 5.6: Training time for increasing number of samples

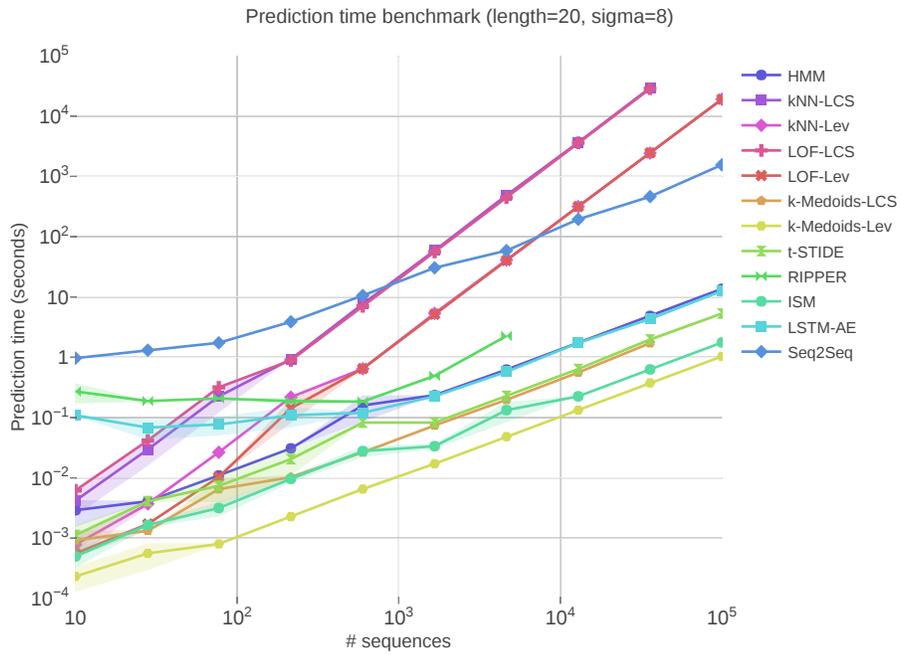

Figure 5.7: Prediction time for increasing number of test samples

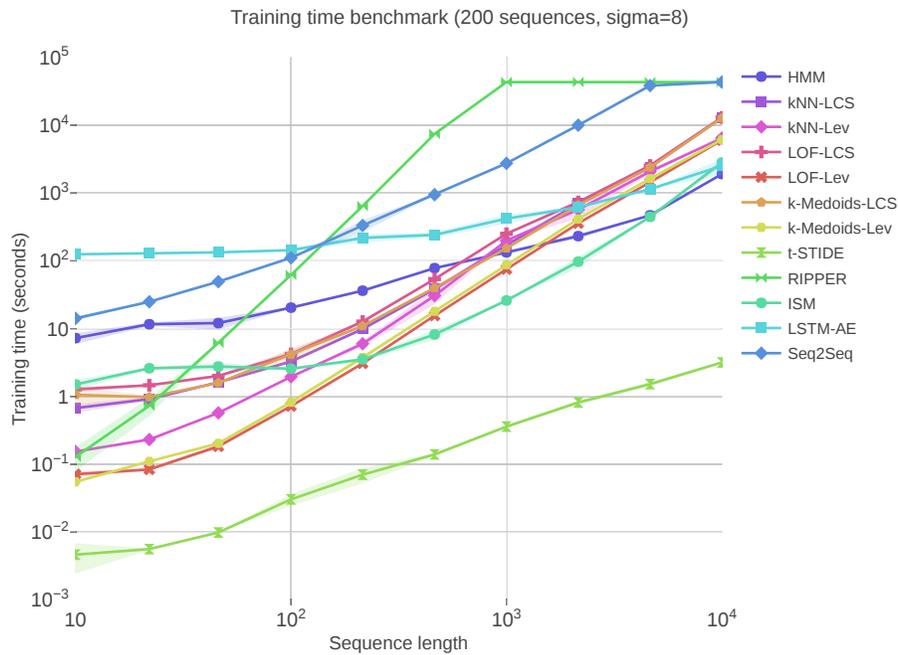

Figure 5.8: Training time for increasing sequence length

These tests show the limitations of RIPPER, which suffers from a long training step, even for datasets of reasonable size. Distance-based methods and SEQ2SEQ also show limited scalability, although $k$-MEDOIDS provide fast predictions and SEQ2SEQ easily supports datasets containing a large number of samples. ɪsm and *t*-sᴛɪᴅᴇ show the best computation time for both training and prediction steps, and could even prove useful in lightweight applications.

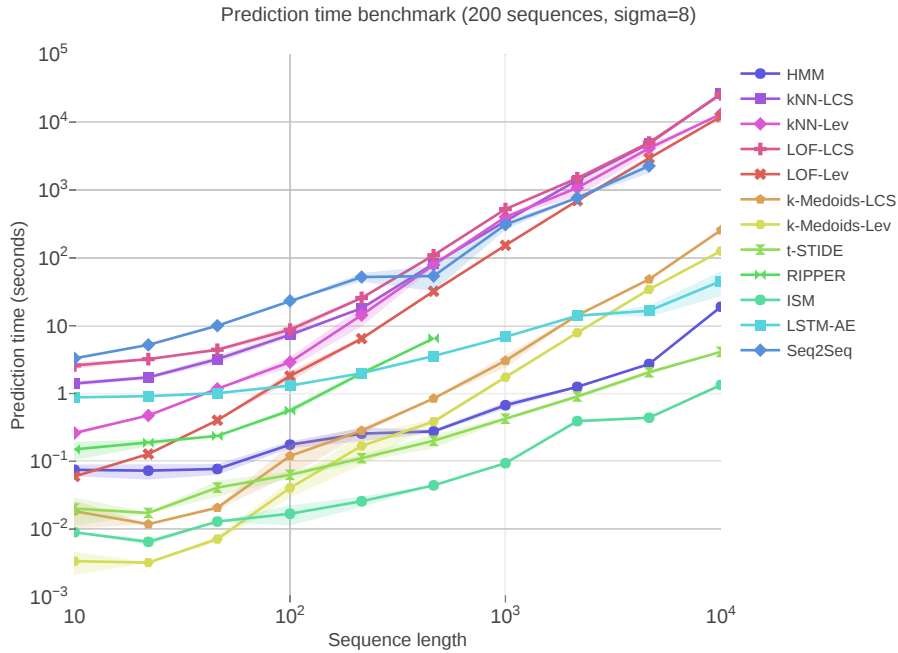

Figure 5.9: Prediction time for increasing sequence length

### 5.2.4 Mᴇᴍᴏʀʏ ᴜsᴀɢᴇ

Monitoring the memory consumption in Figures 5.10 and 5.11 highlights important scalability constraints for several algorithms.

We first observe in Figure 5.10 that memory usage for ʀɪᴘᴘᴇʀ and distance-based methods is strongly correlated with the number of input sequences. ʀɪᴘᴘᴇʀ shows a very high memory usage, although the method reaches our 12h timeout at train time before exceeding the limit of 256GB RAM. Distance-based methods are also strongly impacted by the number of samples. However, most of the memory is here consumed by the pairwise distance matrix. Despite storage optimizations, e.g. symmetric matrix, integers are stored on 24 bytes by Python, resulting in a memory usage of 114GB and 167GB for *k*-ɴɴ-ʟᴇᴠ and ʟᴏғ-ʟᴇᴠ, respectively. Interestingly, ɪsm stabilizes at 10GB after having discovered a sufficient number of patterns from the data. Mini-batch neural networks are not strongly impacted by the number of samples, and the small $\sigma$ limits the diversity of sequences, thus reducing the memory usage of *t*-STIDE.

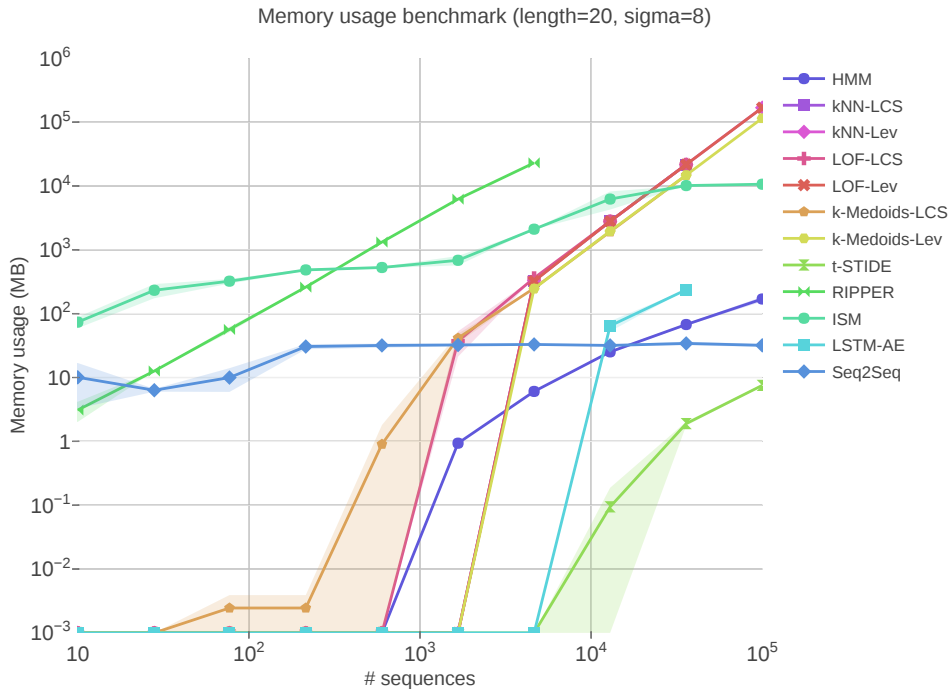

Figure 5.10: Memory usage for increasing number of samples

The metrics reported in Figure 5.11 corroborate the previous conclusions. The experiment reveals a number of rules learnt by RIPPER increasing linearly with the number of events, the final model containing in average $\frac{\#events}{50}$ rules. The size of the decision tree built by association rule learning is thus correlated with the volume of the data. To the opposite, the memory usage of ISM stabilizes again after convergence, showing a more efficient internal representation of the data than RIPPER. The memory consumption of distance-based methods is very low due to small distance matrices, although the computation of LCS shows a memory usage increasing with the length of the sequences compared. Neural networks, especially SEQ2SEQ, are more impacted by the increasing sequence length. This is caused by a network topology depending on the size of the padded sequences, in addition to matrix multiplications of dimensionalities directly impacted by the length of the sequences.

We have observed that most algorithms have a memory consumption strongly related to the volume of input data. The requirements of RIPPER are too important for most systems, and distance-based methods are not suitable to address problems pertaining to more than 20,000 sequences. Interestingly, we did not observe correlations between training or prediction time and memory usage, while one could expect fast algorithms consume more memory, performing faster computations due to a massive caching system. If this may be true when comparing similar methods, the important differences in time and memory are here caused by major discrepancies in the approaches taken by the algorithms.

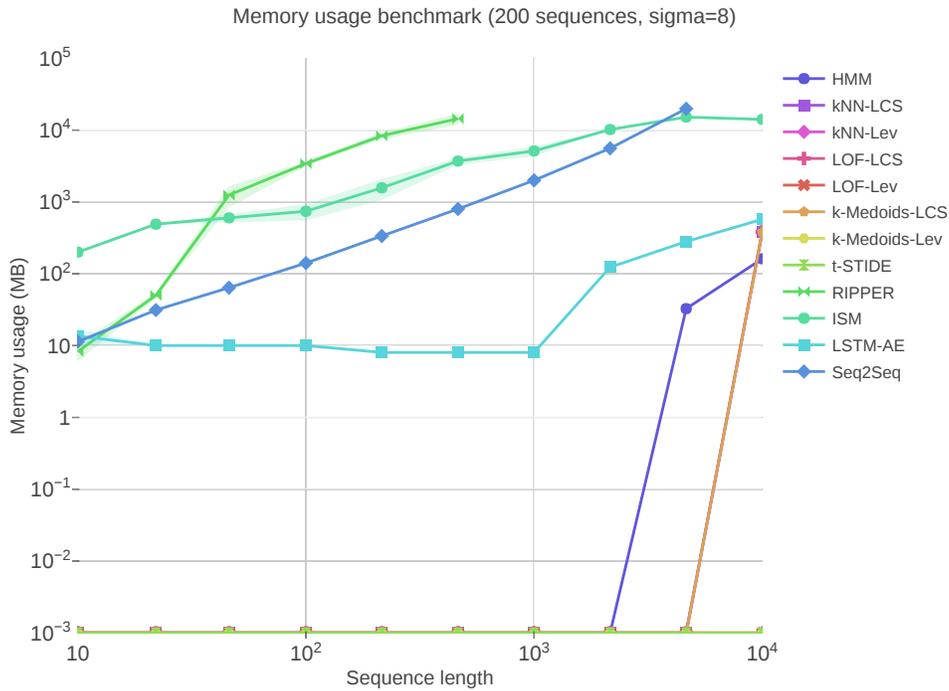

Figure 5.11: Memory usage for increasing sequence length

### 5.2.5 Interpretability

The ability for humans to understand a machine learning model and the resulting predictions is called *interpretability*. This trait allows data scientists to validate the final model and provides useful insights on the targeted dataset, e.g. discovering valuable information about user behaviors which have an important business value. While continuous scores are usually sufficient for automatic intervention modules, this information and the corresponding ranking may not be sufficient when a manual investigation of the anomalies is required. This situation arises for critical applications, where false positives could strongly impact the brand image, e.g. deny access to services for a business partner, or incur heavy costs, e.g. component replacement based on failure prediction with applications to data centers and airplanes. In this case, especially if many alerts are raised every day, the time allocated to manual investigation could be greatly reduced if we could provide the motivations behind high scores to the human expert. Transparency is thus an essential criterion for the choice of algorithms in many applications, and data analysts may accept to trade performance for model accountability. If human eyes may differentiate outlying activity from the underlying patterns in numerical time-series, this task is much harder for discrete event sequences, which emphasizes the need for model interpretability.

The internal representation of interpretable methods provides sufficient information to motivate a predicted score with respect to an input sequence. For example, HMM learns intuitive transition and emission matrices, providing an insightful weighted process flowchart. Unusual event transitions in the test sequence can be visually highlighted by putting a threshold on the emission transition probabilities. Pairwise distance matrices also convey valuable information and can be turned into intuitive visualizations. The matrices can be plotted as Voronoi diagrams, heat maps or fed into a multidimensional scaling (MDS) algorithm resulting in a scatter plot of chosen dimensionality. If additional insight on the distance computations is required, LCS is an intuitive metric and the subsequence common to two compared samples can be underlined. On the other hand, the cost matrix computed by Levenshtein is more difficult to read. Further on, the scoring performed by distance-based methods can be easily motived in the previous 2D representations of distance matrices, e.g. by highlighting the test sample and its $k^{th}$ neighbor for $k$-NN, or the corresponding medoid for $k$-MEDOIDS. The scoring function of LOF is more complex, as it studies the local density of a test sample and its neighbors. Moving back to standard sequence representations, $t$-STIDE is extremely accountable and subsequences can be underlined based on their frequency in the model, thus motivating the resulting score. Pointing out events incorrectly predicted by RIPPER should also provide some information, and interesting patterns learnt by ISM could be emphasized similarly. Neural networks are closer to black-box systems, and their interpretability has recently gained a lot of attention [Zhang & Zhu, 2018]. However, recent efforts mostly focus on numerical and convolutional networks, which leaves room for future LSTM representations. Differences between the input sequence and the reconstructed output could be highlighted for SEQ2SEQ, although it would not explain the underlying model. For LSTM-AE, we could learn and plot a low dimensional numerical representation based on the internal representation of the network, but dimensionality reduction methods will often produce an output biased towards the average sample of the dataset [On-derwater, 2015] and must be selected with care. This is the reason why the reconstruction error is used with SEQ2SEQ to identify anomalies.

Table 5.4: Scalability and interpretability summary. Runtime and memory scalability are reported for datasets of increasing number of samples and sequence length.

| Algorithm | Training/prediction time | | Mem. usage | | |
| | ↗ Samples | ↗ Length | ↗ Samples | ↗ Length | Interpretability |
| --- | --- | --- | --- | --- | --- |
| HMM | Medium/Low | Low/Low | Low | Low | High |
| *k*-NN-LCS | High/High | Medium/High | High | Low | High |
| *k*-NN-LEV | High/High | Medium/High | High | Low | Medium |
| LOF-LCS | High/High | Medium/High | High | Low | Medium |
| LOF-LEV | High/High | Medium/High | High | Low | Medium |
| *k*-MEDOIDS-LCS | High/Low | Medium/Medium | High | Low | High |
| *k*-MEDOIDS-LEV | High/Low | Medium/Medium | High | Low | Medium |
| *t*-STIDE | Low/Low | Low/Low | Low | Low | High |
| RIPPER | High/Low | High/Medium | High | High | Medium |
| ISM | Low/Low | Medium/Low | Medium | Medium | High |
| SEQ2SEQ | Low/Medium | High/High | Low | High | Low |
| LSTM-AE | Low/Low | Low/Low | Low | Medium | Low |

In order to overcome the lack of accountability of a given algorithm, an alternative approach is to infer meaningful rules based on the inputs and outputs predicted by a trained model [de Fortuny & Martens, 2015]. The rule extraction method should provide simple rules showing a transparent decision, while minimizing the prediction error. This is a popular approach used to improve the interpretability of classification models, in particular neural networks and support vector machines (SVMs). Two good rule extraction methods for classifiers are OSRE [Etchells & Lisboa, 2006] and HYPINV [Saad & Wunsch, 2007]. These methods are also compatible with novelty detection when the targeted model produces a binary output such as *fraud* and *non-fraud.* If a continuous anomaly score is required to rank anomalies, we should then resort to regression rule extraction methods which learn rules producing a continuous output, e.g. REFANN [Setiono et al., 2002], ITER [Huysmans et al., 2006] or classification and regression trees (CART) [Breiman et al., 1984]. Both regression and classification rule mining methods show good performance when applied to numerical or one-hot encoded input data. In order to feed temporal data to these algorithms (or to any standard regression or classification methods), numerical features should be extracted from the sequences during a preprocessing step. The feature selection must be performed with great care to minimize the amount of information lost, and was automated for continuous time-series in a previous work [Christ et al., 2016]. While different features should be selected for discrete event sequences, either manually or based on existing techniques [Wang et al., 2001, Saidi et al., 2010], any regression rule extraction technique can be subsequently applied for both data types. The numerical latent representation provided by LSTM autoencoders could be used as input features for rule mining, but it would only improve the interpretability of the decoder, leaving aside the data transformation performed by the encoder.

## 5.3 Summary

This chapter studied the performance and scalability of state-of-the-art novelty detection methods based on a significant collection of real and synthetic datasets. The standard metric used in the literature to compare event sequences is LCS. Given the evidence provided, we found that although LCS produced more transparent insights than the Levenshtein distance, it did not exhibit better anomalies and was computationally more expensive. Our experiments suggest that $k$-NN, $k$-MEDOIDS, $t$-STIDE and LSTM-AE are suitable choices to identify outliers in genomics, and that HMM and RIPPER are efficient algorithms to detect intrusions. HMM is a strong candidate for most novelty detection applications, and shows a good scalability and interpretability. These characteristics make HMM appropriate for user behavior analysis, along with $k$-NN, $k$-MEDOIDS and ISM which also provide a good model accountability. The fast scoring achieved by HMM, $t$-STIDE and ISM implies an excellent management of heavy loads arising in production environments. Major scalability constraints are pointed out for RIPPER and distance-based methods, namely $k$-NN, $k$-MEDOIDS and LOF. The resort to alternative approaches when tackling large volumes of data is recommended. The widely used LSTM networks show a lack of interpretability, and we believe that improving the understanding of recurrent networks as performed in [Karpathy et al., 2016] would strongly benefit to the research community.

# 6

# Conclusions

Unsupervised anomaly detection methods are widely used in a variety of research areas. These approaches are challenged by large volumes of heterogeneous data, training sets contaminated by anomalous samples and strong computational constraints. In this thesis, we studied and developed state-of-the-art algorithms for novelty detection in the context of mixed-type features and temporal data.

## 6.1 Contributions

The first algorithm investigated is the Dirichlet Process Mixture Model (DPMM), a Bayesian density estimation method trained through mean-field variational inference. This approach is a mixture model where each component is represented as a product of exponential-family distributions. The model parameters are learnt by optimizing a lower bound on the log-marginal likelihood, while the mixing proportions of the components are directed by a Dirichlet process. We used a Beta prior on the weights of the Dirichlet process and a Gamma prior on the scaling parameter driving the growth of the number of components. The derivation of the exponential-family representation for suitable likelihoods, conjugate priors and posteriors resulted in a fast and accurate modeling of mixed-type features, providing improved novelty detection performance when applied to datasets composed of numerical and categorical features. However, the use of exponential-family distributions induces a computation overhead.

As a result, this approach is not suitable for datasets composed exclusively of numerical data, where plain Gaussian distributions (DPGMM) provide comparable results.

We further performed a detailed comparative evaluation of state-of-the-art novelty detection methods on a wide range of real datasets. Each problem having its own characteristics, we observed that no method consistently outperformed the others, leaving room for future algorithms designed for specific use cases. A thorough analysis of the data along with a good understanding of the constraints inherent to the problem, e.g. scalability or interpretability, remains thus critical to choose a suitable method. Overall, good novelty detection abilities were observed for Isolation Forest, Robust Kernel Density Estimation and one-class SVM, although the two last methods show important computation times and memory consumptions when applied to large datasets. Simple alternatives with increased scalability are the Gaussian Mixture Model and Probabilistic PCA. Traditional outlier detection algorithms such as LOF and ABOD were strongly outperformed by the previous methods while showing a poor scalability.

As an alternative to the methods introduced above, we developed the Deep Gaussian Process autoencoder (DGP-AE), a probabilistic neural network using approximate Gaussian processes at each layer. The approximation is performed with random feature expansions which yields a tractable and scalable model inferred by stochastic variational inference. The inference only requires tensor products and is achieved through mini-batch learning. This makes our model suitable for distributed and GPU computing. The DGP-AE is flexible and can be applied to any type of data, including mixed-type features. Our model showed meaningful latent representations which suggests good dimensionality reduction abilities. Through experiments, we demonstrated that DGP-AE achieves competitive or better novelty detection performance than state-of-the-art and DNN-based novelty detection methods.

Motivated by industrial constraints in production environments, we eventually performed a review of state-of-the-art novelty detection methods in the context of discrete event sequences. Our study is based on a wide collection of datasets and compares the anomaly detection performance, the scalability and the interpretability of the selected methods. While LCS is a traditional metric for comparing sequences, we showed that the Levenshtein distance provided a similar accuracy for a reduced computation time. With good performance, scalability and interpretability, HMM is the recommended choice for intrusion detection. $k$-NN, $k$-MEDOIDS, $t$-STIDE and LSTM-AE are suitable choices for genomics applications, although distance-based approaches such as $k$-NN and $k$-MEDOIDS are limited to small datasets. Based on their performance and interpretability, HMM, $k$-NN, $k$-MEDOIDS and ISM are also appropriate for user behavior analysis.

## 6.2 Future work

This work suggests several directions for future work. We have observed numerous studies using the area under the ROC curve to compare supervised or unsupervised anomaly detection methods. While this metric is suitable for classification problems with a balanced class distribution, we remind the reader that it is not appropriate for anomaly detection. When tackling such problems, the area under the precision-recall curve, called the average precision, should prevail. Computation time is a strong constraint when selecting an algorithm. Distributing novelty detection methods would allow the use of larger datasets while addressing scalability issues [Otey et al., 2006], thus highlighting possible method-specific trade-offs between accuracy and computation time. In the case of DPMM, treating the truncation level on the number of components as a variational parameter could strongly improve the estimated density while reducing the training time. Extending DPMM to support mini-batch training would also increase the scalability of the method while allowing for distributed and GPU computing. The latent representation provided by DGP-AEs is probabilistic and yields uncertainty estimates. Training a density estimation algorithm on the latent variables to produce new inputs for the decoder would turn DGP-AEs into generative models. Few papers address the problem of novelty detection for images, which leaves room for future work. In line with this perspective, adding convolutional layers to DGP-AEs would make these methods suitable for applications involving images, allowing for end-to-end training of the model and the filters. The use of 1D convolutional layers paired with a product of softmax likelihoods could allow DGP-AEs to learn temporal patterns for discrete event sequences and identify anomalous samples. This scope could be extended to multivariate data, although while similarity metrics have been developed for multivariate time-series [Yang & Shahabi, 2004], we are not aware of such metric for multivariate event sequences. Machine learning algorithms designed for streaming data are actively researched, numerous network-based and sensor-based applications involving data streams [Pokrajac et al., 2007]. However, few novelty detection methods are both incremental and sufficiently scalable to tackle these problems. Unsupervised anomaly detection methods based on novel neural networks such as GANs are also being researched [Schlegl et al., 2017].

# 7

# Résumé

## INTRODUCTION

La détection de nouveautés est une tâche fondamentale et inhérente à de nombreux domaines de recherche. Celle-ci est utilisée pour le nettoyage des données [Liu et al., 2004], la détection d'incidents, le contrôle de dommages [Dereszynski & Dietterich, 2011, Worden et al., 2000], la détection de fraudes liées aux cartes de crédit [Hodge & Austin, 2004] et à la sécurité des réseaux [Pokrajac et al., 2007], ainsi que diverses applications médicales telles que la détection de tumeurs cérébrales [Prastawa et al., 2004], et de cancers du sein [Greensmith et al., 2006]. La détection de nouveautés consiste en l'identification de données de test significativement différentes du jeu de données utilisé pour l'apprentissage du modèle [Pimentel et al., 2014]. Ce problème est également connu sous le nom de "détection d'anomalies". Les difficultés rencontrées dans cette tâche sont causées par la rareté et le coût d'obtention des données labélisées permettant d'identifier des anomalies dans un jeu de données d'apprentissage. En outre, peu d'informations sont généralement disponibles sur la distribution de ces nouveautés. Les données d'apprentissage peuvent également être corrompues par des valeurs extrêmes. Ceci est susceptible d'affecter la capacité des méthodes de détection de nouveautés à caractériser avec précision la distribution des échantillons associés à un comportement normal du système étudié. Il existe par ailleurs de nombreuses applications, telles celles que nous étudions dans cette thèse, où le volume et l'hétérogénéité des données peuvent poser d'importantes contraintes de calcul afin de réagir rapidement aux anomalies identifiées et de développer des algorithmes flexibles pour la détection de nouveautés. À titre d'exemple, la société informatique pour l'aviation Amadeus fournit des plateformes de réservation de billets dont le trafic est de plusieurs millions de transactions par seconde, produisant plus de 3 millions de réservations par jour et des pétaoctets de données stockées. Cette société gère près de la moitié des réservations de vols dans le monde et subit des tentatives de fraude pouvant causer des pertes de revenus et des indemnisations. Détecter des anomalies dans de tels volumes de données est une tâche complexe pour un opérateur humain; ce secteur bénéficierait donc grandement d'une approche automatisée et évolutive. En raison du coût d'obtention de données labélisées et de la difficulté des méthodes supervisées à identifier des anomalies peu fréquentes [Japkowicz & Stephen, 2002], la détection de nouveautés est souvent abordée comme un problème d'apprentissage automatique non supervisé [Pimentel et al., 2014]. À noter que ce problème est également décrit comme semi-supervisé lorsque le jeu de données d'apprentissage est exempt d'anomalies [Chandola et al., 2012].

Nous considérons ici un problème d'apprentissage non supervisé dans lequel est donné en entrée un ensemble de vecteurs $X = [x_1, \ldots, x_n]^\top$. Détecter des nouveautés consiste en l'identification de nouveaux vecteurs de test $x_*$ qui diffèrent considérablement du jeu d'entraînement $X$. La détection de nouveautés est donc un problème de classification contenant une seule classe et visant à construire un modèle décrivant la distribution des échantillons normaux d'un jeu de données. Les méthodes d'apprentissage non supervisées permettent d'effectuer des prédictions sur les données de test $x_*$; étant donné un modèle ayant pour paramètres $\boldsymbol{\theta}$, les prédictions sont définies par $h(x_* | X, \boldsymbol{\theta})$. En supposant que la fonction $h(x_* | X, \boldsymbol{\theta})$ soit continue, celle-ci peut être considérée comme un score permettant de séparer les données nominales des anomalies. Ces scores permettent d'ordonner les vecteurs de test $x_*$, mettant en évidence les points qui diffèrent le plus des données d'entraînement $X$. Plus spécifiquement, il est possible de définir un seuil $\alpha$ et de considérer qu'un point de test $x_*$ est une nouveauté lorsque $h(x_* | X, \boldsymbol{\theta}) > \alpha$.

Après ce seuillage, la qualité d'un algorithme de détection de nouveautés peut être évaluée en se basant sur des mesures proposées dans la littérature pour les problèmes de classification binaires, du nom de *précision* et *rappel*. Dans cette thèse, nous évaluons les résultats des méthodes de détection de nouveautés en faisant varier $\alpha$ sur la plage de valeurs prise par $h(x_* | X, \boldsymbol{\theta})$ sur un ensemble de points de test. Lorsque nous modifions $\alpha$, nous obtenons donc un ensemble de valeurs de précisions et rappels formant une courbe *précision-rappel*. Nous pouvons ensuite calculer la surface sous cette courbe, nommée la *précision moyenne* (AP). En pratique, une valeur optimale pour $\alpha$ est sélectionnée afin d'obtenir un équilibre entre la précision dans les anomalies relevées et le nombre de faux positifs.

La détection de nouveautés a fait l'objet d'études théoriques approfondies [Pimentel et al., 2014,Hodge & Austin, 2004]. Des évaluations expérimentales de l'état de l'art ont également été réalisées [Emmott et al., 2016], étudiant également la malédiction de la dimensionnalité [Zimek et al., 2012]. Dans l'un des travaux les plus récents sur la détection de nouveautés [Pimentel et al., 2014], ces méthodes sont réparties dans les catégories suivantes. (i) Les approches probabilistes estiment la densité de probabilité de $X$, définie par les paramètres du modèle $\boldsymbol{\theta}$. Un score de nouveauté est ensuite obtenu via la fonction de vraisemblance $P(x_*|\boldsymbol{\theta})$, qui calcule la probabilité qu'un point de test soit généré par la distribution estimée au préalable. Ces approches sont génératives et permettent une compréhension simple des données par le biais de distributions paramétrées. (ii) Les méthodes basées sur la distance comparent des échantillons par paires en utilisant diverses métriques de similarité. Les échantillons ayant peu de voisins dans un rayon donné ou se trouvant à une grande distance d'un groupe de points reçoivent un score de nouveauté élevé. (iii) Les méthodes étudiant le domaine d'appartenance des données délimitent le domaine de la classe nominale à l'aide d'une frontière de décision. Le label attribué aux points de test est ensuite basé sur l'emplacement de ces points par rapport à cette limite. (iv) Les approches reposant sur la théorie de l'information mesurent l'augmentation de l'entropie causée par l'inclusion d'un point de test dans la classe nominale. Les méthodes d'isolation (v) sont une alternative et tentent d'isoler les valeurs peu fréquentes des autres échantillons. Ces techniques isolent donc les anomalies au lieu de construire un modèle des points nominaux. (vi) La majorité des réseaux de neurones non supervisés utilisés pour la détection de nouveautés sont des autoencodeurs. Ces réseaux apprennent une représentation compressée des données d'apprentissage en minimisant l'erreur obtenue en comparant les données en entrée et les vecteurs reconstruits en sortie. Les points de test présentant une erreur de reconstruction élevée sont classifiés comme anomalies.

Alors que la plupart des tâches de détection d'anomalies reposent sur des jeux de données numériques [Emmott et al., 2016,Breunig et al., 2000,Ramaswamy et al., 2000], des méthodes de détection de nouveautés ont été appliquées avec succès sur des données catégorielles [Hodge & Austin, 2004], des série chronologiques [Marchi et al., 2015,Kundzewicz & Robson, 2004] des séquences discrètes [Chandola et al., 2008,Warrender et al., 1999,Cohen, 1995] et des données mixtes [Domingues et al., 2018a,Domingues et al., 2018b]. Nous étudions et décrivons l'état de l'art consacré aux méthodes de détection de nouveautés dans le chapitre 2, incluant des algorithmes adaptés aux données numériques, catégorielles et aux séquences d'évènements. Le chapitre 3 détaille un algorithme probabiliste appelé Dirichlet Process Mixture Model (DPMM) que nous entraînons par inférence variationnelle. L'utilisation d'une mixture de distributions appartenant à la famille exponentielle permet d'appliquer ce modèle à des données de types mixtes. Nous effectuons également une évaluation expérimentale des méthodes de l'état de l'art sur des données numériques et catégorielles, incluant l'algorithme DPMM, et comparons les performances de ces algorithmes en nous basant sur leur capacité à détecter des nouveautés, leur scalabilité, leur robustesse et leur sensibilité à la malédiction de la dimensionnalité. Le chapitre 4 décrit un autoencodeur utilisant des processus Gaussiens (Deep Gaussian Process autoencoder). Nous proposons également une approche non paramétrique et probabiliste pour atténuer les problèmes liés au choix d'une architecture appropriée pour ce réseau de neurones, tout en tenant compte de l'incertitude des transformations effectuées par les autoencodeurs. Nous montrons notamment que cela peut être réalisé tout en entraînant le modèle sur de grands volumes de données. Le chapitre 5 étend finalement la comparaison des algorithmes de détection de nouveautés aux méthodes compatibles avec les séquences temporelles. Cette dernière étude détaille les performances des méthodes sur un large éventail de jeux de données appartenant à plusieurs domaines de recherche, tout en fournissant des informations sur la scalabilité et l'interprétabilité des algorithmes sélectionnés.

## Chapitre 3  Dirichlet Process Mixture Model pour la détection d'anomalies

### 3.1  Dirichlet Process Mixture Model

Ce chapitre présente le Dirichlet Process Mixture Model (DPMM), un modèle non paramétrique et probabiliste entraîné par inférence variationnelle [Jordan et al., 1999]. L'algorithme est une méthode non supervisée d'estimation de la densité et de partitionnement, dans laquelle le nombre de composants de la mixture augmente avec l'observation de nouvelles données. Le nombre de composants et les proportions assignées à chaque distribution sont estimés par inférence variationnelle. Dans l'algorithme DPMM, les observations sont générées par une distribution appartenant à la famille exponentielle, produisant un modèle flexible et précis. L'utilisation de fonctions de vraisemblance, de distributions a priori conjuguées et de distributions a posteriori sous forme de familles exponentielles permet à cette méthode d'être compatible avec les données de types mixtes. Ce travail regroupe la méthode d'inférence variationnelle présentée par Bishop dans [Bishop, 2006], l'utilisation d'une distribution a priori Beta sur le processus de Dirichlet responsable des poids de chaque groupe dans [Blei & Jordan, 2006] et l'application d'une distribution a priori Gamma sur le paramètre d'échelle du processus de Dirichlet proposé par [Escobar & West, 1995].

Le Dirichlet Process Mixture Model est défini par un ensemble de variables latentes et de paramètres $\boldsymbol{W}$. Étant donné un ensemble d'observations $\boldsymbol{X}$, nous considérons tout d'abord la densité jointe

$$p(\boldsymbol{X}, \boldsymbol{W}) = p(\boldsymbol{W})p(\boldsymbol{X}|\boldsymbol{W}). \tag{7.1}$$

Notre modèle bayésien génère les variables latentes à partir d'une distribution a priori $p(\boldsymbol{W})$ et vise à modéliser la fonction de vraisemblance $p(\boldsymbol{X}|\boldsymbol{W})$ qui est souvent insoluble. L'inférence de ce modèle consiste à estimer la distribution a posteriori $p(\boldsymbol{W}|\boldsymbol{X})$. Cette tâche est réalisée en approximant $p(\boldsymbol{W}|\boldsymbol{X})$ par la distribution variationnelle $q(\boldsymbol{W})$ en utilisant l'inférence variationnelle. Cette approximation est ici effectuée par la méthode de champ moyen.

Les proportions $\boldsymbol{\pi}$ de la mixture sont décrites par un processus de Dirichlet (DP). Intuitivement, ce processus est similaire au modèle du bâton cassé, où les poids $v_k$ assignés aux $K$ composants de la mixture sont échantillonnés à partir d'une distribution $\text{Beta}(1, w)$. Chaque composant est donc représenté par une distribution exprimée sous forme de famille exponentielle. Les paramètres des fonctions de vraisemblance sont échantillonnés à partir de la distribution a posteriori appartenant également à la famille exponentielle. Les paramètres de la distribution a posteriori sont estimés durant l'apprentissage du modèle. La Figure 7.1 résume les dépendances entre les variables présentes dans le modèle.

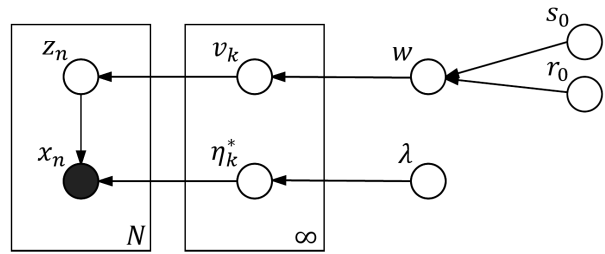

Figure 7.1: Modèle graphique du Dirichlet Process Mixture Model.

Afin d'entraîner ce modèle, nous optimisons une fonction représentant la limite inférieure de la vraisemblance marginale $p(\boldsymbol{X}|\boldsymbol{\theta})$, où $\boldsymbol{\theta}$ représente les paramètres de la distribution a priori. L'application du cadre de travail d'inférence variationnelle nous permet d'obtenir des équations formant un algorithme d'espérance-maximisation (EM). Des itérations successives sur les équations obtenues optimisent de manière déterministe les paramètres de la fonction a posteriori.

L'algorithme DPMM utilisant une mixture de composants, celui-ci peut donc être utilisé afin de répartir des données en plusieurs groupes, chaque groupe contenant des données similaires. Le nombre de groupes à utiliser dans cette tâche est estimé par l'algorithme en fonction des données d'apprentissage.

En guise de contribution supplémentaire, nous présentons dans les appendices A et B la représentation sous forme de famille exponentielle de fonctions de vraisemblance et de distributions a priori conjuguées et a posteriori. Un jeu de données composé à la fois de variables catégorielles et de données numériques sous forme de flottants et d'entiers peut donc être modélisé par DPMM via un produit de distributions catégorielles, de distributions Normales multivariées et de distributions de Poisson, le modèle capturant par ailleurs la corrélation entre ces caractéristiques et regroupant les données par similarité.

## 3.2 Évaluation de l'état de l'art

Nous évaluons l'algorithme DPMM sur un large éventail de tâches de détection de nouveauté. La complexité inhérente à ce domaine est induite par la contamination des données d'entraînement par des anomalies, ainsi que par d'importantes disparités dans la forme, la taille et la densité des groupes de données. La complexité de la détection de nouveautés et l'ancienneté des études adressant ce problème motivent une nouvelle étude expérimentale du domaine. Dans la deuxième partie de ce chapitre, nous étendons notre étude des méthodes de détection de nouveauté (Chapitre 2)en effectuant une comparaison expérimentale approfondie de nombreux algorithmes de l'état de l'art.

Cette étude utilise 12 jeux de données publiques et labélisés, la plupart étant recommandés pour la détection d'anomalies dans [Emmott et al., 2016], auxquels s'ajoutent 3 nouveaux jeux de données industriels générés par les systèmes de production d'Amadeus, entreprise majeure du secteur du voyage. Notre étude utilise davantage de données et de méthodes que les travaux précédents, et examine divers moyens d'utiliser les données catégorielles. À l'inverse, la plupart des études antérieures n'utilisent que des données numériques et identifient le plus souvent les anomalies présentent uniquement dans l'ensemble d'apprentissage, alors que notre étude teste la capacité de généralisation des méthodes en détectant des nouveautés dans les données de test inconnues du modèle. Les algorithmes paramétriques et non paramétriques sélectionnés appartiennent à diverses approches, incluant des algorithmes probabilistes, des méthodes basées sur les plus proches voisins, des réseaux de neurones, des méthodes basées sur la théorie de l'information et des méthodes d'isolation. Leurs performances sur les jeux de données labélisés sont évaluées par l'aire sous les courbes ROC et précision-rappel (PR), respectivement nommées ROC AUC et précision moyenne (AP). Les labels sont utilisés uniquement afin de mesurer les performances des méthodes, et ne sont donc pas accessibles par lesdites méthodes. Ces mesures sont reportées sur la Figure 7.2.

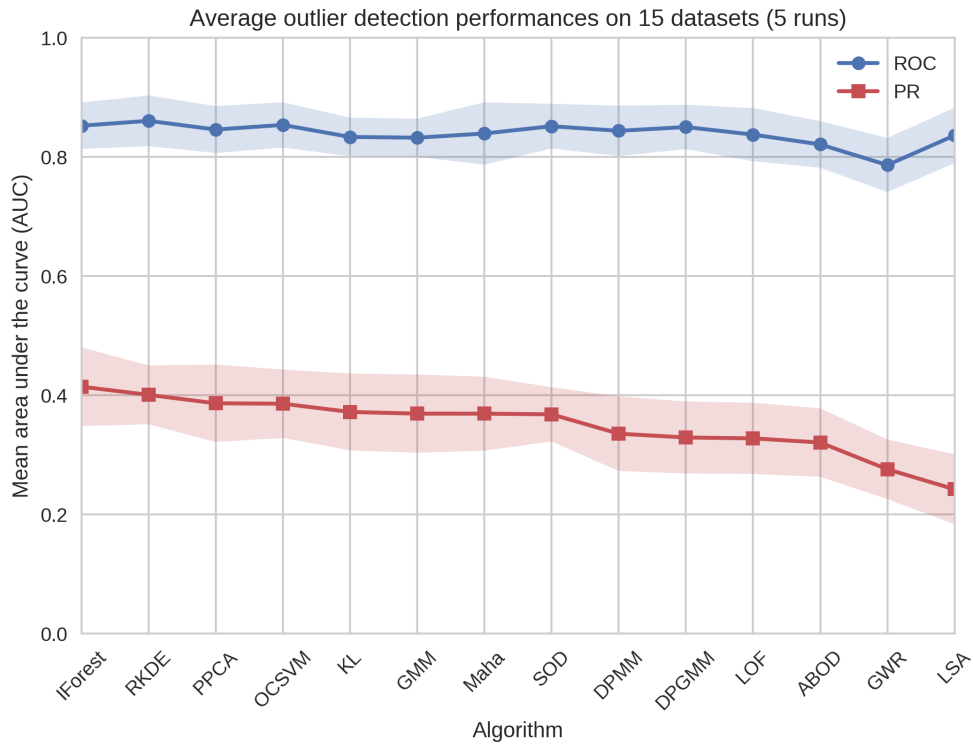

Figure 7.2: Aire moyenne sous les courbes ROC et précision-rappel par algorithme (triées par PR décroissante).

Afin de fournir un aperçu complet de ces méthodes, nous comparons également le temps de calcul requis par l'apprentissage du modèle et par les prédictions, la consommation de mémoire vive et la robustesse de chaque méthode. Ces mesures sont effectuées sur des jeux de données synthétiques avec un nombre croissant d'échantillons et de dimensions, et avec une augmentation de la proportion de bruit de fond. Les résultats nous permettent de comparer les algorithmes non seulement en fonction de leurs performances de détection d'anomalies, mais également en fonction de leur scalabilité, robustesse et adéquation aux problèmes de grandes dimensions. Nos résultats sont résumés dans le Tableau 7.1.

Table 7.1: Résistance à la malédiction de la dimensionnalité et scalabilité du temps de calcul et de la consommation mémoire pour des jeux de données comportant un nombre croissant d'échantillons et de dimensions.

| Algorithme | Temps apprentissage/prédiction | | Usage mémoire | | Robustesse | | |
| | ↗ Échantillons | ↘ Dimensions | ↗ Échantillons | ↘ Dimensions | ↗ Bruit | Haute dim. | Stabilité |
|---|---|---|---|---|---|---|---|
| GMM | Bas/Bas | Moyen/Moyen | Bas | Moyen | Haut | Moyen | Moyen |
| BGM | Bas/Bas | Moyen/Moyen | Bas | Moyen | Haut | Moyen | Haut |
| DPGMM | Moyen/Bas | Haut/Haut | Bas | Haut | Haut | Haut | Haut |
| RKDE | Haut/Haut | Haut/Haut | Haut | Bas | Haut | Haut | Haut |
| PPCA | Bas/Bas | Haut/Bas | Bas | Bas | Haut | Moyen | Moyen |
| LSA | Bas/Moyen | Bas/Bas | Moyen | Bas | Bas | Bas | Moyen |
| MAHA | Bas/Moyen | Moyen/Bas | Bas | Moyen | Moyen | Bas | Haut |
| LOF | Haut/Haut | Bas/Bas | Haut | Bas | Moyen | Haut | Haut |
| ABOD | Bas/Haut | Bas/Moyen | Bas | Bas | Moyen | Bas | Moyen |
| SOD | Haut/Haut | Bas/Moyen | Haut | Bas | Bas | Haut | Moyen |
| KL | Bas/Moyen | Bas/Moyen | Bas | Moyen | Haut | Moyen | Haut |
| GWR | Moyen/Moyen | Moyen/Bas | Bas | Bas | Bas | Haut | Moyen |
| OCSVM | Haut/Haut | Bas/Bas | Bas | Bas | Bas | Haut | Haut |
| IFOREST | Bas/Moyen | Bas/Bas | Moyen | Bas | Haut | Haut | Moyen |

Dans le contexte de la détection d'anomalies non supervisée, nous avons comparé la précision, la robustesse, le temps de calcul et la consommation mémoire de 14 algorithmes sur des jeux de données synthétiques et réels. Notre étude démontre que IFOREST présente de bonnes capacités de détection de nouveautés tout en offrant une excellente scalabilité sur les jeux de données volumineux, ainsi qu'une consommation mémoire acceptable pour des jeux de données inférieurs à un million d'échantillons. Les résultats suggèrent que cet algorithme est plus approprié que RKDE dans un environnement de production car ce dernier est beaucoup plus coûteux en temps de calcul et en mémoire. OCSVM est également un bon candidat, mais n'est pas non plus adapté aux grands ensembles de données.

Certains jeux de données suggérés dans [Emmott et al., 2016] sont obtenus par échantillonnage d'une petite proportion d'anomalies à partir de jeux de données de classification. Ceci donne lieu à des nuages denses d'anomalies qui permettent à des méthodes simples, telle la distance de Mahalanobis, de surpasser plusieurs algorithmes de l'état de l'art conçus pour l'estimation de densité. Si ces algorithmes simples disposent d'une bonne scalabilité, ils ne peuvent cependant capturer la complexité des jeux de données dans lesquels la classe nomi-nale ne suit pas une distribution gaussienne ou est répartie en plusieurs groupes. Dans ces cas, nos tests démontrent la supériorité des alternatives non paramétriques.

SOD a montré de bonnes performances dans la détection d'anomalies, y compris sur les jeux de données contenant de nombreuses dimensions, au prix d'une faible scalabilité. L'utilisation de familles exponentielles pour DPMM s'est révélée extrêmement coûteuse en temps de calcul sans pour autant améliorer de manière substantielle la détection de nouveautés effectuée par la même méthode se basant seulement des distributions gaussiennes (DPGMM). Néanmoins, l'utilisation de distributions catégorielles dans DPMM a permis de réduire le temps de calcul sur les ensembles de données de types mixtes, tout en améliorant les performances de détection d'anomalies sur ces mêmes jeux de données. LOF, ABOD, GWR, KL et LSA ont obtenu les plus basses performances, les trois premières méthodes ayant également une faible scalabilité. Nous avons par ailleurs évalué la densité modélisée par chaque méthode et mis en évidence un cas limite pour ABOD dans le cas d'ensembles de données composés de plusieurs groupes distincts.

Si cette étude couvre la plupart des algorithmes communément utilisés pour résoudre des problèmes de détection de nouveautés, certains algorithmes spécifiques peuvent être choisis dans le cas d'environnements contraignants. Par exemple, les capacités d'un algorithme à être distribué, entraîné sur un flux de données ou entraîné par mini-lots peuvent être des conditions préalables à la gestion de grands volumes de données. L'extension de méthodes existantes afin de supporter ces fonctionnalités est un important domaine de recherche, les méthodes DPMM, K-MEANS, SVM ou GMM étant déjà compatibles. D'autres perspectives de recherche sur la détection d'anomalies s'orientent également vers les méthodes d'apprentissage ensemblistes [Zimek et al., 2014] et la détection de valeurs extrêmes sur des jeux de données multi-vues [Iwata & Yamada, 2016]. DPMM pourrait être amélioré en apprenant par inférence variationnelle le seuil de troncature $K$ sur le nombre de composants utilisés dans la mixture. Détecter automatiquement la meilleure fonction de vraisemblance à utiliser pour chaque caractéristique offrirait également une flexibilité supplémentaire au modèle. Notre implémentation de cette méthode pourrait enfin être étendue pour prendre en charge l'apprentissage par mini-lots, ce qui permettrait d'entraîner cet algorithme de manière distribuée.

## Chapitre 4  Deep Gaussian Process pour la détection d'anomalies

Les réseaux de neurones profonds sont récemment devenus la méthode d'apprentissage privilégiée pour les problèmes supervisés, notamment en raison de leur importante capacité de représentation et de leur scalabilité sur de grands volumes de données [LeCun et al., 2015].

Ces méthodes ont permis d'atteindre d'excellentes performances dans de nombreux domaines d'applications tels que la vision par ordinateur [Krizhevsky et al., 2012], la reconnaissance vocale [Hinton et al., 2012] et le traitement du langage naturel [Collobert & Weston, 2008]. La question est donc de savoir si ces techniques peuvent également s'appliquer et fournir des résultats d'une telle qualité dans le cas de l'apprentissage non supervisé et plus spécifiquement pour de la détection de nouveautés. Les réseaux de neurones profonds appliqués à l'apprentissage non supervisé font l'objet d'importantes recherches [Kingma & Welling, 2014, Goodfellow et al., 2014], mais nous ignorons encore si ceux-ci peuvent concurrencer les méthodes de détection de nouveautés modernes. Nous n'avons pas connaissance d'études récentes sur les réseaux de neurones visant la détection de nouveautés. Le dernier article sur ce sujet date de 15 ans [Markou & Singh, 2003] et n'inclus donc pas les développements récents effectués dans ce domaine.

Les principaux challenges liés à l'utilisation de réseaux de neurones profonds pour des tâches d'apprentissage sont (i) la nécessité de spécifier une architecture adaptée au problème à résoudre et (ii) la nécessité de contrôler la généralisation du modèle. Diverses formes de régularisation ont été proposées afin d'atténuer le problème de surapprentissage et d'améliorer la généralisation, comme le *dropout* [Srivastava et al., 2014b, Gal & Ghahramani, 2016], mais des questions restent en suspens sur la manière et les principes généraux de conception de ces réseaux de neurones. Les processus gaussiens profonds (DGPs) sont des candidats idéaux pour adresser simultanément les problèmes (i) et (ii) ci-dessus. Les DGPs sont des modèles probabilistes non paramétriques profonds utilisant une composition de processus probabilistes qui permettent d'utiliser implicitement un nombre infini de neurones dans chaque couche [Damianou & Lawrence, 2013, Duvenaud et al., 2014]. De plus, leur nature probabiliste induit une forme de régularisation empêchant le surapprentissage et permettant de sélectionner le modèle de manière efficace [Neal, 1996]. Bien que les DGPs soient particulièrement attrayants pour résoudre les problèmes généraux adressés par les réseaux de neurones, l'apprentissage des ces modèles est difficile à résoudre. Récemment, plusieurs contributions ont été apportées afin de simplifier l'entraînement de ces modèles [Bui et al., 2016, Cutajar et al., 2017, Bradshaw et al., 2017], et ceux-ci peuvent actuellement concurrencer les réseaux de neurones profonds (DNNs) en termes de scalabilité et précision tout en fournissant une meilleure quantification de l'incertitude [Gal & Ghahramani, 2016, Cutajar et al., 2017, Gal et al., 2017].

Ce chapitre présente un algorithme non supervisé pour la détection de nouveautés basé sur les processus gaussiens profonds et utilisant une architecture autoencodeur. Le DGP autoencodeur proposé (DGP-AE) effectue une approximation des processus gaussiens de chaque couche du modèle en générant de nouvelles dimensions aléatoires et en entraînant le modèle obtenu par inférence variationnelle stochastique. Les principales caractéristiques de l'approche proposée sont les suivantes: (i) Les DGP-AES sont des modèles probabilistes non supervisés capables d'estimer des distributions extrêmement complexes et disposent d'une fonction de notation pour prédire des scores d'anomalies; (ii) Les DGP-AES peuvent modéliser tout type de données, y compris les cas comportant des caractéristiques de types mixtes, telles que des données continues, des entiers positifs et des variables catégorielles; (iii) L'entraînement du modèle ne nécessite pas de factorisations matricielles coûteuses en temps de calcul et potentiellement problématiques d'un point de vue algébrique, mais uniquement des produits de tenseurs; (iv) Les DGP-AES peuvent être entraînés à l'aide d'un apprentissage par mini-lots, et peuvent donc exploiter les infrastructures distribuées ainsi que les GPUs; (v) L'entraînement des DGP-AES utilise l'inférence variationnelle stochastique, et peut donc être implémenté aisément à l'aide d'outils de différenciation automatique, rendant cette méthode pratique et scalable pour la détection de nouveautés. Bien que nous laissions cela pour des travaux futurs, nous notons que DGP-AE peut être facilement adapté à différents types de représentations, par exemple avec des filtres de convolution pour les applications basées sur des images, ce qui permet un entraînement simultané du modèle et des filtres.

L'algorithme DGP-AE est comparé avec de nombreux réseaux de neurones concurrents proposés dans la littérature conçus pour adresser les problèmes non supervisés relatifs à d'importants volumes de données, comme les autoencodeurs variationnels (VAE) [Kingma & Welling, 2014], les processus gaussiens variationnels autoencodés ( VAE-DGP) [Dai et al., 2016] et l'estimateur de densité autorégressif (NADE) [Uria et al., 2016]. À travers une série d'expériences, dans lesquelles nous comparons ces méthodes avec l'état de l'art des méthodes de détection de nouveautés, comme Isolation Forest [Liu et al., 2008] et l'estimation de densité robuste par noyau (RKDE) [Kim & Scott, 2012], nous démontrons que les DGP-AES offrent une capacité de modélisation flexible et un algorithme d'apprentissage pratique, tout en atteignant des performances de pointe.

Les DGP-AES sont des réseaux de neurones utilisant une architecture dite d'autoencodeur. Un autoencodeur est un modèle dont les couches sont divisées en deux parties, celles appartenant à l'encodeur et celles appartenant au décodeur. La partie encodeur du réseau de neurones transforme chaque entrée x en un vecteur de variables latentes z, tandis que la partie décodeur tente de reconstruire le vecteur d'entrée x à partir des variables latentes z. Les variables latentes étant généralement de dimension inférieure aux vecteurs d'entrée, le réseau apprend donc une représentation compressée des données. L'entraînement de ce modèle est le plus souvent réalisé en minimisant l'erreur de reconstruction entre les vecteurs en entrée et ceux en sortie produits par le réseau.

Dans cette thèse, nous proposons de construire les fonctions de transformations de l'encodeur et du décodeur à l'aide de processus gaussiens (GP). En conséquence, nous souhaitons apprendre de manière conjointe une projection non linéaire probabiliste basée sur les DGPs (l'encodeur) et un modèle de variables latentes basé sur les DGPs (le décodeur). Les blocs de construction formant les couches des réseaux de neurones DGPs sont les processus gaussiens (GPs), qui sont des distributions a priori sur des fonctions; de manière formelle, un GP est un ensemble de variables aléatoires ayant pour propriété que tout sous-ensemble de ces variables est conjointement gaussien [Rasmussen & Williams, 2006]. La fonction de covariance d'un GP modélise la covariance entre les variables aléatoires sur différents vecteurs d'entrée, et il est possible de définir une fonction paramétrique pour leur moyenne.

Empiler plusieurs couches de GPs dans un réseau DGP signifie que la sortie d'un GP est donnée en entrée au GP constituant la couche suivante; cette construction donne lieu à une composition de processus stochastiques, similaire à une composition de fonctions. Afin d'illustrer nos propos, la Figure 7.3 représente un DGP-AE à deux couches.

Les performances de notre modèle sont évaluées en mesurant l'aire sous la courbe précision-rappel. Ces mesures sont effectuées sur 11 jeux de données labélisés pour la détection d'anomalies, et nos expériences comparent un total de 12 algorithmes, dont 10 réseaux de neurones et deux algorithmes provenant de l'état de l'art de la détection de nouveautés. Ces mesures sont reportées dans le Tableau 4.3.

Afin de comparer la convergence et la scalabilité de notre réseau de neurones avec l'état de l'art, nous reportons également dans la Figure 7.4 la convergence du log de la fonction de vraisemblance moyenne (MLL) ainsi que l'aire sous la courbe précision-rappel.

Nos expériences montrent que les DGP-AEs atteignent en moyenne les meilleures performances dans le cadre de la détection de nouveautés. L'utilisation d'une fonction de vraisemblance *softmax* pour les variables catégorielles et l'augmentation du nombre de couches de ces réseaux de neurones nous permet en outre d'améliorer les performances des DGP-AEs.

L'étude de la convergence montre un apprentissage rapide de la fonction vraisemblance pour les DGP-AEs, en particulier lors de l'utilisation d'un noyau ARC. Ceci démontre l'efficacité de la régularisation des DGPs et leur capacité à généraliser lors de l'apprentissage de modèles complexes. Des expériences supplémentaires effectuées sur les variables latentes montrent par ailleurs que les DGP-AEs ont une excellente capacité à réduire le nombre de dimensions d'un jeu de données tout en conservant les différences entre échantillons (Figure 4.5).

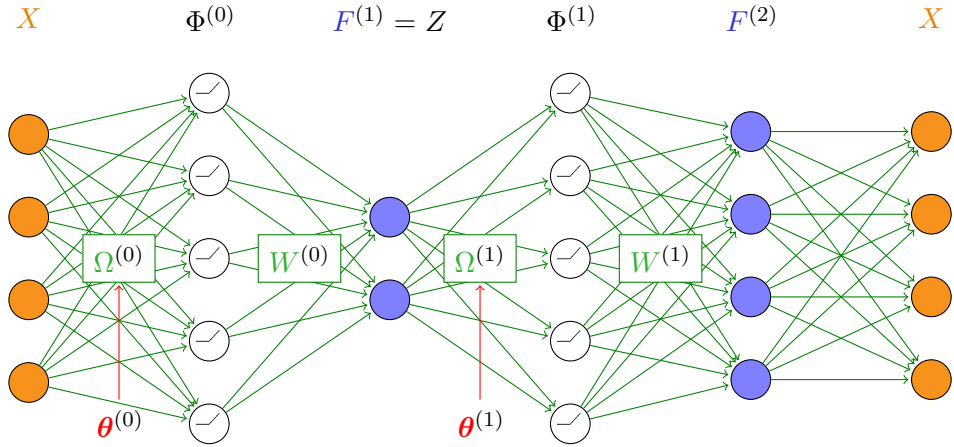

Figure 7.3: Architecture d'un autoencodeur DGP composé de deux couches GPs. Les processus gaussiens sont approximés par un ensemble de deux couches, la première $\Phi^{(l)}$ effectue une extension aléatoire des dimensions, suivie par une transformation linéaire qui résulte en la couche $F^{(l)}$. Les paramètres des fonctions de covariance sont $\theta^{(l)} = \left((\sigma^2)^{(l)}, \Lambda^{(l)}\right)$, et les distributions a priori sur les poids sont $p\left(\Omega_{\cdot j}^{(l)}\right) = N\left(0, \left(\Lambda^{(l)}\right)^{-1}\right)$ et $p\left(W_{\cdot i}^{(l)}\right) = N(0, I)$. La couche $Z$ représente les variables latentes.

Ce chapitre présente donc un nouveau réseau de neurones probabiliste pour la détection non supervisée d'anomalies. Le modèle DGP-AE proposé est un autoencodeur reposant sur des processus gaussiens pour représenter les transformations inhérentes à l'encodeur et au décodeur. L'inférence de ce modèle est scalable et effectuée par approximation des DGPs via une extension aléatoire des dimensions. L'entraînement du modèle obtenu par inférence variationnelle stochastique permet l'exploitation des infrastructures distribuées et des GPUs. Le DGP-AE est capable de modéliser des données de manière flexible et supporte les jeux de données contenant des dimensions de types mixtes, cette capacité étant activement étudiée dans la littérature récente [Vergari et al., 2018]. De plus, le modèle dispose d'un entraînement robuste et d'une facilité d'implémentation via des outils de différenciation automatique, puisque contrairement à la plupart des modèles basés sur des GPs [Dai et al., 2016], notre modèle n'utilise que des produits de tenseurs et aucune factorisation matricielle. À travers une série d'expériences, nous avons enfin démontré que les DGP-AEs obtenaient des résultats compétitifs par rapport aux méthodes de détection de nouveautés les plus récentes et aux méthodes de détection de nouveautés basées sur des réseaux de neurones.

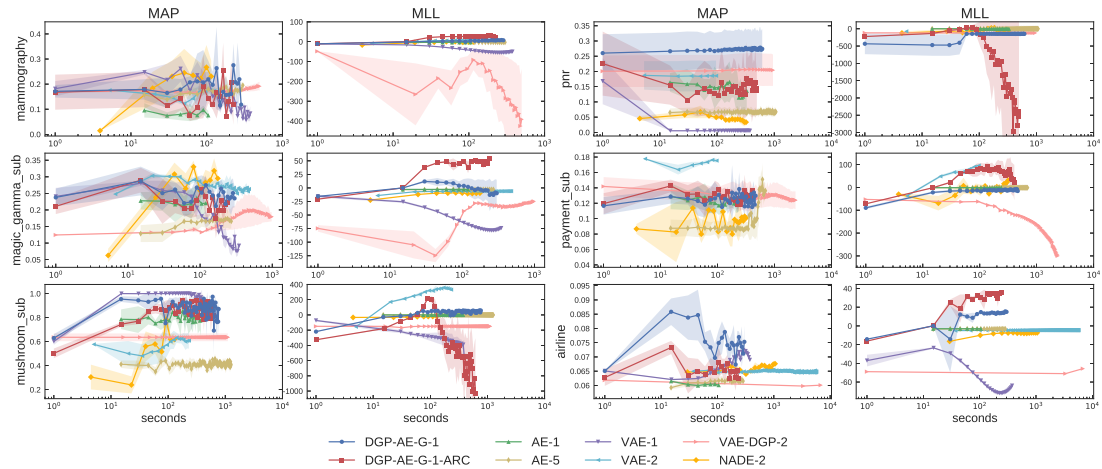

Figure 7.4: Évolution de la précision moyenne (map) et du log de la fonction de vraisemblance (mll) au cours du temps pour les réseaux de neurones sélectionnés. Les mesures sont effectuées sur des données de test lors d'une validation croisée (3 *folds*). Pour les deux métriques, des résultats élevés indiquent de bonnes performances.

Ce chapitre étudie le problème de la détection des anomalies appliqué aux données temporelles, en particulier dans le cadre de séquences discrètes d'événements ordonnés dans le temps. Ce problème peut être divisé en deux catégories. Le premier est la *détection de points de rupture*, où les ensembles de données sont de longues séquences dans lesquelles nous cherchons à localiser des sous-séquences anormales et contiguës, dénotant un changement soudain de comportement. Les cas d'utilisation relatifs à ce problème sont la lecture de capteurs [Kundzewicz & Robson, 2004] et la *first story detection* [Petrović et al., 2010]. Une deuxième catégorie considère les jeux de données comme des ensembles de séquences et cible la détection de séquences anormales par rapport aux échantillons nominaux. Notre étude se concentre sur ce dernier domaine, qui englobe des cas d'utilisation tels que l'identification de protéines pour la génomique [Chandola et al., 2008, Sun et al., 2006], la détection de fraudes et d'intrusions [Maxion & Townsend, 2002, Warrender et al., 1999, Chandola et al., 2008] et l'analyse de comportements utilisateur (uba) [Sculley & Brodley, 2006].

Bien qu'il s'agisse d'un sujet d'intérêt dans la littérature, la plupart des études analysant ce problème se concentrent sur ses aspects théoriques [Gupta et al., 2014, Chandola et al., 2012], et ne fournissent donc pas d'évaluation ni de comparaison expérimentale des algorithmes pouvant résoudre ce problème. Chandola et al. présentent une comparaison des méthodes de détection de nouveautés pour les données séquentielles, mais leur article utilise une métrique artisanale qui n'est pas validée par la littérature pour mesurer les performances, et n'inclus pas les algorithmes récemment publiés dans le domaine. Notre travail prolonge les études précédentes en y ajoutant les contributions suivantes: (i) comparaison des performances de détection de nouveautés de 12 algorithmes, en incluant de nouveaux réseaux de neurones, sur 81 jeux de données contenant des séquences discrètes issues de divers domaines de recherche; (ii) évaluation de la robustesse des méthodes sélectionnées à l'aide de jeux de données contaminés par des valeurs extrêmes, contrairement aux études précédentes qui se limitent à des données d'entraînement exemptes d'anomalies; (iii) évaluation de la scalabilité de chaque algorithme, en reportant le temps de calcul requis par l'apprentissage et la prédiction, la consommation de mémoire et la capacité de détection d'anomalies sur des jeux de données synthétiques comportant un nombre croissant d'échantillons, d'évènements dans chaque séquence et d'anomalies; (iv) étude de l'interprétabilité des différentes approches dans le but de fournir des explications et de motiver les prédictions du modèle entraîné. À notre connaissance, cette étude est la première à évaluer les méthodes de détection de nouveautés pour les séquences discrètes avec autant de jeux de données et d'algorithmes. Ce travail est également le premier à évaluer la scalabilité des méthodes sélectionnées, critère de sélection important pour les environnements soumis à de fortes contraintes sur les temps de réponse, et pour les systèmes limités en ressources tels les systèmes intégrés. Les 81 jeux de données utilisés sont liés à la génomique, à la détection d'intrusions et à l'analyse du comportement des utilisateurs. Ces jeux de données sont divisés en 9 catégories et couvrent un total de 68 832 séquences (Table 5.2).

Une fois nos expériences réalisées, nous avons pu comparer les performances des algorithmes sur la détection de nouveautés. Cependant, poussant plus loin notre analyse des résultats sur ces jeux de données, nous avons souhaité construire un modèle d'inférence du comportement de chaque méthode en fonction des caractéristiques des données d'entraînement. Nous avons donc appris un méta-modèle interprétable en nous basant sur les caractéristiques des jeux de données reportés dans le Tableau 5.2. Les performances des méthodes et les caractéristiques des jeux de données sont fournies en entrée, et nous construisons un arbre de décision par algorithme afin de prédire les performances de la méthode sélectionnée. Nous obtenons des modèles de classification binaires qui prédisent si un algorithme figurera dans les 25% plus performants ou dans les 25% moins performants. Ces arbres exposent les forces et les faiblesses des méthodes étudiées et mettent en évidence les caractéristiques des données impactant les performances des méthodes. De manière plus visuelle, le résultat de cette analyse est résumé dans la Figure 7.5. Les filtres sélectionnés sont obtenus en extrayant les règles ayant une profondeur inférieure à 4 dans les arbres de décisions.

Table 7.2: Interprétabilité et scalabilité des méthodes.

| Algorithme | Temps apprentissage/prédiction | | Usage mémoire | | |
|---|---|---|---|---|---|
| | ↗ Échantillons | ↗ Longueur | ↗ Échantillons | ↗ Longueur | Interprétabilité |
| HMM | Moyen/Bas | Bas/Bas | Bas | Bas | Haute |
| $k$-NN-LCS | Haut/Haut | Moyen/Haut | Haut | Bas | Haute |
| $k$-NN-LEV | Haut/Haut | Moyen/Haut | Haut | Bas | Moyenne |
| LOF-LCS | Haut/Haut | Moyen/Haut | Haut | Bas | Moyenne |
| LOF-LEV | Haut/Haut | Moyen/Haut | Haut | Bas | Moyenne |
| $k$-MEDOIDS-LCS | Haut/Bas | Moyen/Moyen | Haut | Bas | Haute |
| $k$-MEDOIDS-LEV | Haut/Bas | Moyen/Moyen | Haut | Bas | Moyenne |
| $t$-STIDE | Bas/Bas | Bas/Bas | Bas | Bas | Haute |
| RIPPER | Haut/Bas | Haut/Moyen | Haut | Haut | Moyenne |
| ISM | Bas/Bas | Moyen/Bas | Moyen | Moyen | Haute |
| SEQ2SEQ | Bas/Moyen | Haut/Haut | Bas | Haut | Basse |
| LSTM-AE | Bas/Bas | Bas/Bas | Bas | Moyen | Basse |

Nous étudions également la scalabilité et l'interprétabilité de ces méthodes et reportons le résumé de nos analyses dans le Tableau 7.2.

Nous avons mesuré les performances et la scalabilité de l'état de l'art des méthodes de détection de nouveautés, utilisant de nombreux jeux de données synthétiques et réels. La métrique standard utilisée dans la littérature pour comparer les séquences d'événements est LCS. Bien que LCS produise un résultat plus transparent que la distance de Levenshtein, nos expériences montrent que LCS ne permet pas d'identifier de meilleures anomalies et requiert davantage de temps de calcul. Les résultats suggèrent que $k$-NN, $k$-MEDOIDS, $t$-STIDE et LSTM-AE sont des choix appropriés pour identifier des anomalies dans la génomique, et que HMM et RIPPER sont des algorithmes efficaces pour détecter les intrusions. HMM est un bon candidat pour la plupart des applications de détection de nouveautés et dispose d'une bonne scalabilité et interprétabilité. HMM est donc approprié pour l'analyse de comportements utilisateurs, tout comme $k$-NN, $k$-MEDOIDS et ISM qui produisent des modèles suffisamment transparents. La rapidité de prédiction observée pour HMM, $t$-STIDE et ISM montre que ces algorithmes sont capables de résister à une charge importante inhérente aux environnements de production. Un manque de scalabilité a été noté pour RIPPER et les méthodes basées sur des matrices de distances, telles $k$-NN, $k$-MEDOIDS et LOF. Le recours à des approches alternatives dans le cadre d'importants volumes de données est donc recommandé. Les réseaux de neurones LSTM montrent un manque d'interprétabilité, suggérant des axes de recherche futurs motivés par l'utilisation massive de ces méthodes.

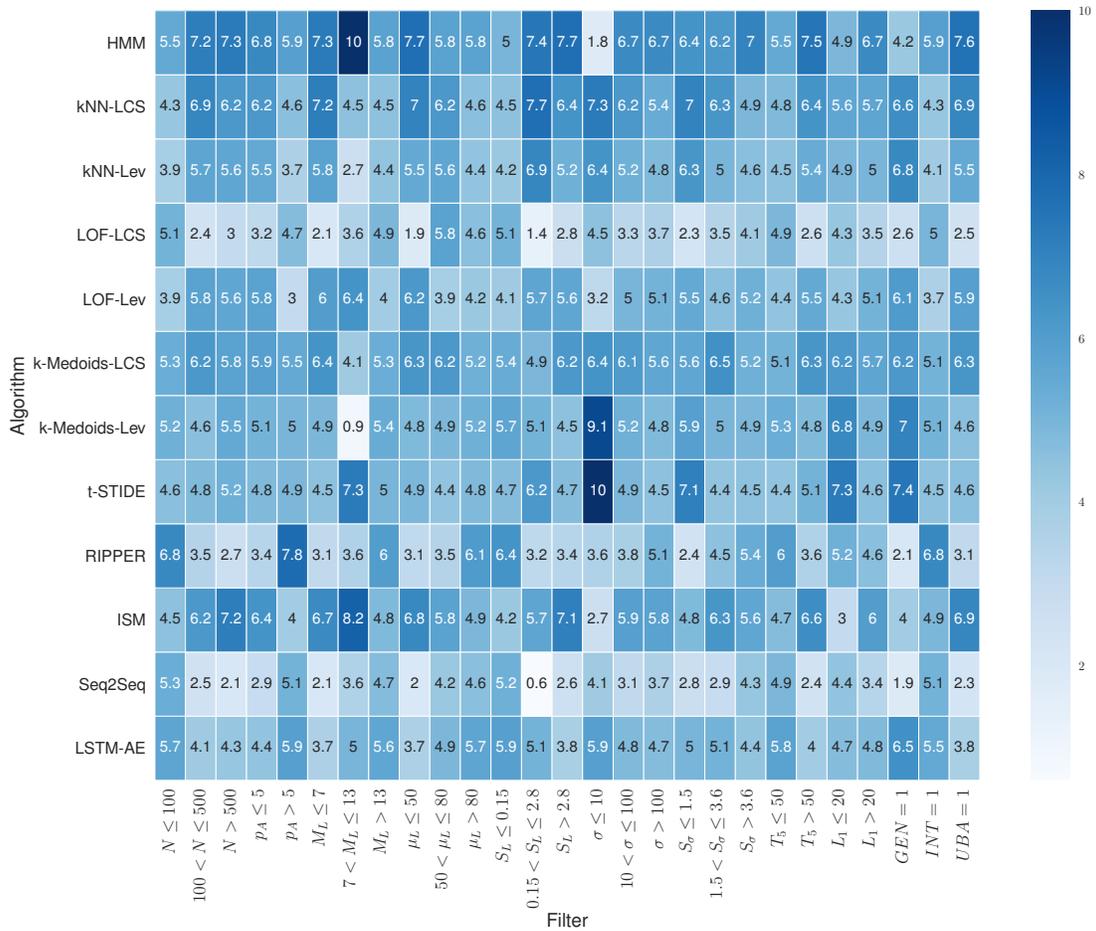

Figure 7.5: Capacité de détection d'anomalies par algorithme et par caractéristique. Un score proche de 10 indique que la méthode est la plus à même de détecter des anomalies sur ce type de données. Un score proche de 0 signifie que la méthode est parmi les moins efficaces pour cet ensemble de données. $N$ est le nombre d'échantillons; $p_A$ la proportion d'anomalies; $M_L$, $\mu_L$ et $S_L$ sont le minimum, la moyenne et l'entropie sur la longueur des séquences; $\sigma$ et $S_\sigma$ sont le nombre d'évènements et l'entropie sur cette distribution; $T_5$ est la proportion des données représentée par les 5% d'évènements les plus fréquents; $L_1$ est la proportion d'évènements rares représentant 1% des données; GEN, INT et UBA correspondent aux jeux de données de génomique, intrusion et comportements utilisateurs.

## Perspectives

Cette thèse suggère plusieurs axes de recherches pouvant donner lieu à des travaux futurs. Nous avons constaté que de nombreuses études font usage de l'aire sous la courbe ROC pour comparer des méthodes de détection d'anomalies supervisées ou non supervisées. Bien que cette métrique soit appropriée pour les problèmes de classification comportant une distribution de classes équilibrée, nous rappelons que celle-ci ne doit pas être utilisée pour la détection d'anomalies. Dans ce cas spécifique, l'aire sous la courbe précision-rappel, appelée précision moyenne, prévaut. Le temps de calcul est une caractéristique importante lors de la sélection d'un algorithme. La distribution de méthodes de détection de nouveautés permettrait d'utiliser des ensembles de données plus volumineux tout en résolvant les problèmes de scalabilité [Otey et al., 2006], mettant ainsi en évidence d'éventuels compromis, propres à chaque méthode, entre précision et temps de calcul. Dans le cas de DPMM, le seuil de troncature sur le nombre de composants utilisés dans la mixture pourrait être remplacé par une variable variationnelle, ce qui améliorerait le processus d'estimation de densité et pourrait réduire le temps de calcul. L'entraînement par mini-lots représente également un axe d'amélioration pour l'algorithme DPMM, cette fonctionnalité améliorant la scalabilité de la méthode tout en rendant possible l'utilisation du calcul distribué et des GPUs. Les variables latentes produites par les DGP-AEs sont probabilistes, ce qui induit une incertitude dans les transformations. Entraîner un algorithme d'estimation de densité sur ces variables latentes permettrait de générer de nouvelles entrées pour le décodeur, transformant les DGP-AEs en algorithmes génératifs. Peu de travaux appliquent la détection de nouveautés aux images, laissant la porte ouverte à des travaux futurs. L'ajout de couches de convolution dans les DGP-AEs rendrait ces méthodes appropriées pour la classification d'images, ce qui permettrait l'inférence du modèle et des filtres. Associer des couches de convolution de dimension 1 à un produit de vraisemblances de type *softmax* permettrait aux DGP-AEs d'apprendre des modèles temporels orientés vers les séquences d'événements discrets, permettant l'identification de séquences anormales. Dans le cas de séquences mul-tivariées, des métriques de similarité ont été développées pour les séries temporelles [Yang & Shahabi, 2004]. Cependant, nous n'avons pas connaissance d'une telle métrique pour les séquences d'événements discrets. Les algorithmes d'apprentissage automatique conçus pour être entraînés sur des flux de données continus sont en cours d'investigation, ceux-ci répondant aux contraintes des nombreuses applications réseaux ou reposant sur des capteurs qui impliquent des flux de données [Pokrajac et al., 2007]. Cependant, peu de méthodes de détection de nouveautés sont à la fois incrémentales et de scalabilité suffisante pour résoudre ces problèmes. Des méthodes de détection d'anomalies non supervisées basées sur de nouveaux réseaux de neurones, tels que les GANs, font également l'objet de recherches [Schlegl et al., 2017].

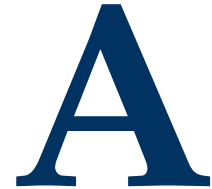

# Derivation of exponential-family distributions

## A.1 Multivariate normal

Let $d$ be the dimension of the space. In the following equations, we take advantage of the property $\text{tr}(a^T \cdot b) = vec(a) \cdot vec(b)$ with $a$ and $b$ vectors. We thus assume a vectorization of the matrices at the second-to-last step of the Normal and Normal-Wishart derivations. This allows us to use the trace property: $\text{tr}(a^T B c) = \text{tr}(c a^T B)$.

$$N(\boldsymbol{x}|\boldsymbol{\mu}, \boldsymbol{\Sigma}) = (2\pi)^{-\frac{d}{2}} |\boldsymbol{\Sigma}|^{-\frac{1}{2}} e^{-\frac{1}{2}(\boldsymbol{x}-\boldsymbol{\mu})^T \boldsymbol{\Sigma}^{-1}(\boldsymbol{x}-\boldsymbol{\mu})}$$

$$= \frac{1}{\sqrt{2\pi}^d} \exp(-\frac{1}{2} \ln |\boldsymbol{\Sigma}| - \frac{1}{2}(\boldsymbol{x}-\boldsymbol{\mu})^T \boldsymbol{\Sigma}^{-1}(\boldsymbol{x}-\boldsymbol{\mu}))$$

$$= \frac{1}{\sqrt{2\pi}^d} \exp(\mathrm{tr}\left(\boldsymbol{\Sigma}^{-1}\boldsymbol{\mu}\boldsymbol{x}^T\right) - \frac{1}{2}\mathrm{tr}\left(\boldsymbol{\Sigma}^{-1}\boldsymbol{x}\boldsymbol{x}^T\right) - \frac{1}{2}\boldsymbol{\mu}^T \boldsymbol{\Sigma}^{-1}\boldsymbol{\mu} - \frac{1}{2}\ln |\boldsymbol{\Sigma}|)$$

$$= \frac{1}{\sqrt{2\pi}^d} \exp\left( \begin{pmatrix} \boldsymbol{\Sigma}^{-1}\boldsymbol{\mu} \\ -\frac{1}{2}\boldsymbol{\Sigma}^{-1} \end{pmatrix}^T \cdot \begin{pmatrix} \boldsymbol{x} \\ \boldsymbol{x}\boldsymbol{x}^T \end{pmatrix} - (\frac{1}{2}\boldsymbol{\mu}^T \boldsymbol{\Sigma}^{-1}\boldsymbol{\mu} + \frac{1}{2}\ln |\boldsymbol{\Sigma}|) \right) \tag{A.1}$$

$$= h(\boldsymbol{x}) \exp(\eta(\boldsymbol{\theta}) \cdot T(\boldsymbol{x}) - A(\boldsymbol{\theta}))$$

This gives us the following parameter and inverse parameter mappings:

$$\begin{cases} \boldsymbol{\eta_1} = \boldsymbol{\Sigma}^{-1}\boldsymbol{\mu} \\ \boldsymbol{\eta_2} = -\frac{1}{2}\boldsymbol{\Sigma}^{-1} \end{cases} \qquad\qquad \begin{cases} \boldsymbol{\mu} = -\frac{1}{2}\boldsymbol{\eta_2}^{-1}\boldsymbol{\eta_1} \\ \boldsymbol{\Sigma} = -\frac{1}{2}\boldsymbol{\eta_2}^{-1} \end{cases}$$

## A.2 Wishart

$$W(\boldsymbol{x}|\boldsymbol{V}, n) = \frac{|\boldsymbol{x}|^{\frac{n-d-1}{2}} e^{\frac{-\mathrm{tr}(\boldsymbol{V}^{-1}\boldsymbol{x})}{2}}}{2^{\frac{nd}{2}}|\boldsymbol{V}|^{\frac{n}{2}}\Gamma_d(\frac{n}{2})}$$

$$= \exp(-\frac{1}{2}\mathrm{tr}(\boldsymbol{V}^{-1}\boldsymbol{x}) + \frac{n-d-1}{2}\ln|\boldsymbol{x}| - \frac{nd}{2}\ln 2 - \frac{n}{2}\ln|\boldsymbol{V}| - \ln\Gamma_d(\frac{n}{2}))$$

$$= \exp\left( \begin{pmatrix} -\frac{1}{2}\boldsymbol{V}^{-1} \\ \frac{n-d-1}{2} \end{pmatrix}^T \cdot \begin{pmatrix} \boldsymbol{x} \\ \ln|\boldsymbol{x}| \end{pmatrix} - (\frac{n}{2}(d\ln 2 + \ln|\boldsymbol{V}|) + \ln\Gamma_d(\frac{n}{2})) \right)$$

$$= h(\boldsymbol{x}) \exp(\eta(\boldsymbol{\theta}) \cdot T(\boldsymbol{x}) - A(\boldsymbol{\theta}))$$

$$\tag{A.2}$$

Where $\Gamma_d$ is the multivariate gamma function. The parameter mappings are:

$$\begin{cases} \boldsymbol{\eta_1} = -\frac{1}{2}\boldsymbol{V}^{-1} \\ \boldsymbol{\eta_2} = \frac{n-d-1}{2} \end{cases} \qquad\qquad \begin{cases} \boldsymbol{V} = -\frac{1}{2}\boldsymbol{\eta_1}^{-1} \\ n = 2\eta_2 + d + 1 \end{cases}$$

### A.3 Normal-Wishart

The following derivation assumes a preliminary vectorization of the matrices as explained in Section A.1.

$$\boldsymbol{\Lambda}|\boldsymbol{V}, n \sim W(\boldsymbol{\Lambda}|\boldsymbol{V}, n) \tag{A.3}$$

$$\boldsymbol{\mu}|\boldsymbol{\mu}_0, \lambda, \boldsymbol{\Lambda} \sim N(\boldsymbol{\mu}|\boldsymbol{\mu}_0, (\lambda \boldsymbol{\Lambda})^{-1}) \tag{A.4}$$

$$(\boldsymbol{\mu}, \boldsymbol{\Lambda}) \sim NW(\boldsymbol{\mu}_0, \lambda, \boldsymbol{V}, n) \tag{A.5}$$

$$
\begin{aligned}
NW(\boldsymbol{x}, \boldsymbol{\Lambda}|\boldsymbol{\mu}_0, \lambda, \boldsymbol{V}, n) &= \frac{|\boldsymbol{\Lambda}|^{\frac{n-d-1}{2}} e^{\frac{-\operatorname{tr}(\boldsymbol{V}^{-1}\boldsymbol{\Lambda})}{2}}}{2^{\frac{nd}{2}} |\boldsymbol{V}|^{\frac{n}{2}} \Gamma_d(\frac{n}{2})} (2\pi)^{-\frac{d}{2}} |(\lambda\boldsymbol{\Lambda})^{-1}|^{-\frac{1}{2}} e^{-\frac{1}{2}(\boldsymbol{x}-\boldsymbol{\mu}_0)^T \lambda\boldsymbol{\Lambda}(\boldsymbol{x}-\boldsymbol{\mu}_0)} \\
&= \frac{1}{\sqrt{2\pi}^d} \exp(\operatorname{tr}\left(\lambda\boldsymbol{\Lambda}\boldsymbol{\mu}_0\boldsymbol{x}^T\right) - \frac{1}{2}\operatorname{tr}\left(\lambda\boldsymbol{\Lambda}\boldsymbol{x}\boldsymbol{x}^T\right) - \frac{1}{2}\boldsymbol{\mu}_0^T\lambda\boldsymbol{\Lambda}\boldsymbol{\mu}_0 \\
&\quad - \frac{1}{2}\ln|(\lambda\boldsymbol{\Lambda})^{-1}| - \frac{1}{2}\operatorname{tr}(\boldsymbol{V}^{-1}\boldsymbol{\Lambda}) + \frac{n-d-1}{2}\ln|\boldsymbol{\Lambda}| - \frac{nd}{2}\ln 2 \\
&\quad - \frac{n}{2}\ln|\boldsymbol{V}| - \ln\Gamma_d(\frac{n}{2})) \\
&= \frac{1}{\sqrt{2\pi}^d} \exp(\frac{n-d-1}{2}\ln|\boldsymbol{\Lambda}| + \frac{d}{2}\ln\lambda + \frac{1}{2}\ln|\boldsymbol{\Lambda}| - \frac{1}{2}\boldsymbol{\mu}_0^T\lambda\boldsymbol{\Lambda}\boldsymbol{\mu}_0 \\
&\quad - \frac{1}{2}\operatorname{tr}(\boldsymbol{V}^{-1}\boldsymbol{\Lambda}) + \operatorname{tr}\left(\lambda\boldsymbol{\Lambda}\boldsymbol{\mu}_0\boldsymbol{x}^T\right) - \frac{1}{2}\operatorname{tr}\left(\lambda\boldsymbol{\Lambda}\boldsymbol{x}\boldsymbol{x}^T\right) - \frac{nd}{2}\ln 2 \\
&\quad - \frac{n}{2}\ln|\boldsymbol{V}| - \ln\Gamma_d(\frac{n}{2})) \\
&= \frac{1}{\sqrt{2\pi}^d} \exp\left( \begin{pmatrix} \frac{n-d}{2} \\ -\frac{1}{2}(\boldsymbol{\mu}_0\boldsymbol{\mu}_0^T\lambda + \boldsymbol{V}^{-1}) \\ \boldsymbol{\mu}_0\lambda \\ -\frac{1}{2}\lambda \end{pmatrix}^T \cdot \begin{pmatrix} \ln|\boldsymbol{\Lambda}| \\ \boldsymbol{\Lambda} \\ \boldsymbol{x}^T\boldsymbol{\Lambda} \\ \boldsymbol{\Lambda}\boldsymbol{x}\boldsymbol{x}^T \end{pmatrix} - (-\frac{d}{2}\ln\lambda \right. \\
&\quad \left. + \frac{nd}{2}\ln 2 + \frac{n}{2}\ln|\boldsymbol{V}| + \ln\Gamma_d(\frac{n}{2})) \right) \\
&= h(\boldsymbol{x}) \exp(\eta(\boldsymbol{\theta}) \cdot T(\boldsymbol{x}) - A(\boldsymbol{\theta}))
\end{aligned} \tag{A.6}
$$

Which results in the following parameter mappings:

$$\begin{cases} \eta_1 = \frac{n-d}{2} \\ \boldsymbol{\eta_2} = -\frac{1}{2}(\boldsymbol{\mu}_0 \boldsymbol{\mu}_0^T \lambda + \boldsymbol{V}^{-1}) \\ \boldsymbol{\eta_3} = \boldsymbol{\mu}_0 \lambda \\ \eta_4 = -\frac{1}{2}\lambda \end{cases} \qquad \begin{cases} \boldsymbol{\mu}_0 = -\frac{\boldsymbol{\eta_3}}{2\eta_4} \\ \lambda = -2\eta_4 \\ \boldsymbol{V} = \left(-2\boldsymbol{\eta_2} + \frac{\boldsymbol{\eta_3}\boldsymbol{\eta_3}^T}{2\eta_4}\right)^{-1} \\ n = 2\eta_1 + d \end{cases}$$

## A.4 Conjugate prior of the Beta distribution

The hyperparameters of this prior are $\lambda_0$, $x_0$ and $y_0$. However, its normalization factor does not have a closed form which limits the use of the Beta distribution.

$$\begin{aligned} \pi(\alpha, \beta | \lambda_0, x_0, y_0) &\propto \left(\frac{\Gamma(\alpha+\beta)}{\Gamma(\alpha)\Gamma(\beta)}\right)^{\lambda_0} x_0^\alpha y_0^\beta \\ &\propto \exp\left(\lambda_0 \ln\left(\frac{\Gamma(\alpha+\beta)}{\Gamma(\alpha)\Gamma(\beta)}\right) + \alpha \ln x_0 + \beta \ln y_0\right) \\ &\propto \exp\left(\begin{pmatrix} \lambda_0 \\ \ln x_0 \\ \ln y_0 \end{pmatrix}^T \cdot \begin{pmatrix} \ln\left(\frac{\Gamma(\alpha+\beta)}{\Gamma(\alpha)\Gamma(\beta)}\right) \\ \alpha \\ \beta \end{pmatrix}\right) \\ &= h(\boldsymbol{x})\exp(\eta(\boldsymbol{\theta})\cdot T(\boldsymbol{x}) - A(\boldsymbol{\theta})) \end{aligned} \qquad \text{(A.7)}$$

Where the parameter mappings are

$$\begin{cases} \eta_1 = \lambda_0 \\ \eta_2 = \ln x_0 \\ \eta_3 = \ln y_0 \end{cases} \qquad \begin{cases} \lambda_0 = \eta_1 \\ x_0 = e_2^\eta \\ y_0 = e_3^\eta \end{cases}$$

## A.5 Conjugate prior of the Gamma distribution

The hyperparameters of this prior are $p$, $q$, $r$ and $s$.

$$
\begin{aligned}
f(\alpha, \beta | p, q, r, s) &\propto \frac{p^{\alpha-1} e^{-\beta q}}{\Gamma(\alpha)^r \beta^{-\alpha s}} \\
&\propto p^{\alpha-1} e^{-\beta q} \Gamma(\alpha)^{-r} \beta^{\alpha s} \\
&\propto \exp\left((\alpha - 1)\ln p - \beta q - r \ln \Gamma(\alpha) + \alpha s \ln \beta\right) \\
&\propto \exp\left(\begin{pmatrix} r \\ s \\ \ln p \\ -q \end{pmatrix}^T \cdot \begin{pmatrix} \ln \Gamma(\alpha) \\ \alpha \ln \beta \\ \alpha \\ \beta \end{pmatrix} - \ln p\right) \\
&= h(\boldsymbol{x}) \exp(\eta(\boldsymbol{\theta}) \cdot T(\boldsymbol{x}) - A(\boldsymbol{\theta}))
\end{aligned}
\tag{A.8}
$$

The parameter mappings are

$$
\begin{cases}
\eta_1 = r \\
\eta_2 = s \\
\eta_3 = \ln p \\
\eta_4 = -q
\end{cases}
\qquad\qquad
\begin{cases}
p = e^{\eta_3} \\
q = -\eta_4 \\
r = \eta_1 \\
s = \eta_2
\end{cases}
$$

# B

# Derivation of exponential-family conjugate priors and posteriors

## B.1  Beta - Conjugate prior of Binomial likelihood

$$
\begin{aligned}
p(\eta^*|\boldsymbol{\lambda}) &= h(\eta^*)\exp(\lambda_1^T\eta^* + \lambda_2(-a(\eta^*)) - a(\boldsymbol{\lambda})) \\
&= \exp\left(\lambda_1\ln\frac{p}{1-p} + \lambda_2 n\ln(1-p) - a(\boldsymbol{\lambda})\right) \\
&= \left(\frac{p}{1-p}\right)_1^\lambda (1-p)^{n\lambda_2}e^{-a(\boldsymbol{\lambda})} \\
&= p^{(\lambda_1+1)-1}(1-p)^{(n\lambda_2-\lambda_1+1)-1}e^{-a(\boldsymbol{\lambda})} \\
&= \frac{p^{\alpha-1}(1-p)^{\beta-1}}{B(\alpha,\beta)}
\end{aligned}
\tag{B.1}
$$

We recognize a Beta distribution with parameters

$$
\begin{cases}
\lambda_1 = \alpha - 1 \\
\lambda_2 = \frac{\beta+\alpha-2}{n}
\end{cases}
\qquad\qquad
\begin{cases}
\alpha = \lambda_1 + 1 \\
\beta = n\lambda_2 - \lambda_1 + 1
\end{cases}
$$

The expectation of terms of the sufficient statistics for the posterior are given thereafter,

where $\psi$ is the digamma function.

$$
\begin{aligned}
\mathbb{E}[\boldsymbol{\eta^*}] &= \frac{\partial a(\tau_1, \cdots)}{\partial \tau_1} \\
&= \frac{\partial}{\partial \tau_1} \left( \ln \Gamma(\tau_1 + 1) + \ln \Gamma(\beta) - \ln \Gamma(\tau_1 + \beta + 1) \right) \\
&= \psi(\tau_1 + 1) - \psi(\tau_1 + \beta + 1) \\
&= \psi(\alpha) - \psi(\alpha + \beta)
\end{aligned}
\tag{B.2}
$$

$$
\begin{aligned}
\mathbb{E}[-a(\boldsymbol{\eta^*})] &= \frac{\partial a(\cdots, \tau_2)}{\partial \tau_2} \\
&= \frac{\partial}{\partial \tau_2} \left( \ln \Gamma(\alpha) + \ln \Gamma(n\tau_2 - \tau_1 + 1) - \ln \Gamma(\alpha + n\tau_2 - \tau_1 + 1) \right) \\
&= n\psi(n\tau_2 - \tau_1 + 1) - n\psi(\alpha + n\tau_2 - \tau_1 + 1) \\
&= n\psi(\beta) - n\psi(\alpha + \beta)
\end{aligned}
\tag{B.3}
$$

## B.2 Dirichlet - Conjugate prior of Multinomial likelihood

$$
\begin{aligned}
p(\boldsymbol{\eta^*}|\boldsymbol{\lambda}) &= h(\boldsymbol{\eta^*}) \exp(\boldsymbol{\lambda_1}^T \boldsymbol{\eta^*} + \lambda_2(-a(\boldsymbol{\eta^*})) - a(\boldsymbol{\lambda})) \\
&= \exp\left( \boldsymbol{\lambda_1}^T \begin{pmatrix} \ln p_1 \\ \vdots \\ \ln p_m \end{pmatrix} - a(\boldsymbol{\lambda}) \right) \\
&= \prod_{i=1}^m p_i^{(\lambda_{1i}+1)-1} e^{-a(\boldsymbol{\lambda})} \\
&= \frac{1}{B(\alpha)} \prod_{i=1}^m p_i^{\alpha_i - 1}
\end{aligned}
\tag{B.4}
$$

We recognize a Dirichlet distribution with parameters

$$
\begin{cases}
\boldsymbol{\lambda_1} = \begin{pmatrix} \lambda_{11} \\ \vdots \\ \lambda_{1m} \end{pmatrix} = \begin{pmatrix} \alpha_1 - 1 \\ \vdots \\ \alpha_m - 1 \end{pmatrix} \\
\lambda_2 = 0
\end{cases}
\qquad
\begin{cases}
\boldsymbol{\alpha} = \begin{pmatrix} \lambda_{11} + 1 \\ \vdots \\ \lambda_{1m} + 1 \end{pmatrix}
\end{cases}
$$

The expectations for the posterior are

$$
\begin{aligned}
\mathbb{E}[\boldsymbol{\eta^*}] &= \frac{\partial a(\boldsymbol{\tau_1}, \cdots)}{\partial \boldsymbol{\tau_1}} \\
&= \frac{\partial}{\partial \boldsymbol{\tau_1}} \left( \sum_{i=1}^{m} \ln \Gamma(\tau_{1i} + 1) - \ln \Gamma \left( \sum_{i=1}^{m} (\tau_{1i} + 1) \right) \right) \\
&= \begin{pmatrix} \psi(\tau_{11} + 1) - \psi \left( \sum_{i=1}^{m} \tau_{1i} + 1 \right) \\ \vdots \\ \psi(\tau_{1m} + 1) - \psi \left( \sum_{i=1}^{m} \tau_{1i} + 1 \right) \end{pmatrix} \\
&= \begin{pmatrix} \psi(\alpha_1) - \psi \left( \sum_{i=1}^{m} \alpha_i \right) \\ \vdots \\ \psi(\alpha_m) - \psi \left( \sum_{i=1}^{m} \alpha_i \right) \end{pmatrix}
\end{aligned}
\tag{B.5}
$$

$$
\mathbb{E}[-a(\boldsymbol{\eta^*})] = 0 \tag{B.6}
$$

## B.3  Gamma - Conjugate prior of Poisson likelihood

$$
\begin{aligned}
p(\eta^*|\lambda) &= h(\eta^*) \exp(\lambda_1 \eta^* + \lambda_2(-a(\eta^*)) - a(\lambda)) \\
&= \exp(\lambda_1 \ln \lambda_0 - \lambda_2 \lambda_0 - \ln \Gamma(\alpha) + \alpha \ln(\beta)) \\
&= \frac{\beta^\alpha}{\Gamma(\alpha)} \lambda_0^{\lambda_1} e^{-\lambda_2 \lambda_0} \\
&= \frac{\beta^\alpha}{\Gamma(\alpha)} \lambda_0^{\alpha - 1} e^{-\beta \lambda_0}
\end{aligned}
\tag{B.7}
$$

We recognize a Gamma distribution where $\lambda_0$ is the parameter of the Poisson distribution.

$$\begin{cases} \lambda_1 = \alpha - 1 \\ \lambda_2 = \beta \end{cases} \qquad\qquad \begin{cases} \alpha = \lambda_1 + 1 \\ \beta = \lambda_2 \end{cases}$$

The expectations are

$$\begin{aligned}
\mathbb{E}[\eta^*] &= \frac{\partial a(\tau_1, \cdots)}{\partial \tau_1} \\
&= \frac{\partial}{\partial \tau_1}(\ln \Gamma(\tau_1 + 1) - (\tau_1 + 1)\ln \beta) \\
&= \psi(\tau_1 + 1) - \ln \beta \\
&= \psi(\alpha) - \ln \beta
\end{aligned} \tag{B.8}$$

$$\begin{aligned}
\mathbb{E}[-a(\eta^*)] &= \frac{\partial a(\cdots, \tau_2)}{\partial \tau_2} \\
&= \frac{\partial}{\partial \tau_2}(\ln \Gamma(\alpha) - \alpha \ln(\tau_2)) \\
&= \frac{\alpha}{\tau_2} \\
&= \frac{\alpha}{\beta}
\end{aligned} \tag{B.9}$$

$$p(\boldsymbol{\eta^*}|\boldsymbol{\lambda}) = h(\boldsymbol{\eta^*}) \exp(\boldsymbol{\lambda_1^T}\boldsymbol{\eta^*} + \lambda_2(-a(\boldsymbol{\eta^*})) - a(\boldsymbol{\lambda}))$$

$$= (2\pi)^{-\frac{d}{2}} \exp\left(\boldsymbol{\lambda_1^T}\begin{pmatrix}\boldsymbol{\Sigma^{-1}\mu} \\ -\frac{1}{2}\boldsymbol{\Sigma^{-1}}\end{pmatrix} + \lambda_2\left(-\frac{1}{2}\boldsymbol{\mu^T\Sigma^{-1}\mu} - \frac{1}{2}\ln|\boldsymbol{\Sigma}|\right) - a(\boldsymbol{\lambda})\right)$$

$$= (2\pi)^{-\frac{d}{2}} \exp\left(\boldsymbol{\lambda_{11}^T}\boldsymbol{\Lambda\mu} - \frac{1}{2}\boldsymbol{\lambda_{12}^T}\boldsymbol{\Lambda} - \frac{1}{2}\lambda_2\boldsymbol{\mu^T\Lambda\mu} - \frac{1}{2}\lambda_2\ln|\boldsymbol{\Lambda^{-1}}| - a(\boldsymbol{\lambda})\right)$$

$$= (2\pi)^{-\frac{d}{2}} \exp\left((\boldsymbol{\mu_0}\lambda_0)^T\boldsymbol{\Lambda\mu} - \frac{1}{2}(\boldsymbol{\mu_0\mu_0^T}\lambda_0 + \boldsymbol{V^{-1}})\boldsymbol{\Lambda} - \frac{1}{2}\lambda_0\boldsymbol{\mu^T\Lambda\mu}\right.$$

$$\left. + \frac{1}{2}((n-d-1)+1)\ln|\boldsymbol{\Lambda}| - a(\boldsymbol{\lambda})\right)$$

$$= (2\pi)^{-\frac{d}{2}} \exp\left(\mathrm{tr}(\lambda_0\boldsymbol{\Lambda\mu\mu_0^T}) - \frac{1}{2}\boldsymbol{\mu_0^T}\lambda_0\boldsymbol{\Lambda\mu_0} - \frac{1}{2}\mathrm{tr}(\lambda_0\boldsymbol{\Lambda\mu\mu^T}) - \frac{1}{2}\mathrm{tr}(\boldsymbol{V^{-1}\Lambda})\right.$$

$$\left. + \frac{n-d-1}{2}\ln|\boldsymbol{\Lambda}| + \frac{1}{2}\ln|\boldsymbol{\Lambda}| - \left(-\frac{d}{2}\ln\lambda_0 + \frac{nd}{2}\ln 2 + \frac{n}{2}\ln|\boldsymbol{V}| + \ln\Gamma_d(\frac{n}{2})\right)\right)$$

$$= \frac{|\boldsymbol{\Lambda}|^{\frac{n-d-1}{2}}e^{\frac{-\mathrm{tr}(\boldsymbol{V^{-1}\Lambda})}{2}}}{2^{\frac{nd}{2}}|\boldsymbol{V}|^{\frac{n}{2}}\Gamma_d(\frac{n}{2})}(2\pi)^{-\frac{d}{2}}|\lambda_0\boldsymbol{\Lambda}|^{\frac{1}{2}}e^{-\frac{1}{2}(\boldsymbol{\mu-\mu_0})^T\lambda_0\boldsymbol{\Lambda}(\boldsymbol{\mu-\mu_0})}$$

$$(B.10)$$

We recognize a $NW(\boldsymbol{\mu}, \boldsymbol{\Lambda}|\boldsymbol{\mu}_0, \lambda_0, \boldsymbol{V}, n)$ distribution with the following parameters. The previous derivation used transformations such as $|\lambda A| = \lambda^d|A|$ or $|A^{-1}| = |A|^{-1}$ and assumes that matrices are vectorized.

$$\begin{cases}\boldsymbol{\lambda_{11}} = \boldsymbol{\mu_0}\lambda_0 \\ \boldsymbol{\lambda_{12}} = (\boldsymbol{\mu_0\mu_0^T}\lambda_0 + \boldsymbol{V^{-1}})^T \\ \lambda_2 = \lambda_0 \\ \lambda_2 = n - d\end{cases} \qquad \begin{cases}\boldsymbol{\mu_0} = \frac{\boldsymbol{\lambda_{11}}}{\lambda_2} \\ \lambda_0 = \lambda_2 \\ \boldsymbol{V} = \left(\boldsymbol{\lambda_{12}} - \frac{\boldsymbol{\lambda_{11}\lambda_{11}^T}}{\lambda_2}\right)^{-T} \\ n = \lambda_2 + d\end{cases}$$

The mappings induce the constraint $\lambda_0 = n - d$. We used the notation $\boldsymbol{B^{-T}} = (\boldsymbol{B^{-1}})^T$.

The expectations for the posterior are given below, where $\mathbb{E}[\boldsymbol{\eta^*}]$ contains a vector and a matrix and $\mathbb{E}[-a(\boldsymbol{\eta^*})]$ is a scalar. $\boldsymbol{\tau}$ is the natural parameter of the posterior, corresponding to $\boldsymbol{\lambda}$ for the prior. We also used the inverse parameter mapping previously defined.

$$\mathbb{E}[\boldsymbol{\eta}^*] = \begin{pmatrix} \frac{\partial a(\boldsymbol{\tau_{11}}, \cdots)}{\partial \boldsymbol{\tau_{11}}} \\ \frac{\partial a(\cdots, \boldsymbol{\tau_{12}}, \cdots)}{\partial \boldsymbol{\tau_{12}}} \end{pmatrix} \tag{B.11}$$

$$
\begin{aligned}
\frac{\partial a(\boldsymbol{\tau_{11}}, \cdots)}{\partial \boldsymbol{\tau_{11}}} &= \frac{\partial}{\partial \boldsymbol{\tau_{11}}} \Bigg( -\frac{d}{2} \ln \tau_2 + \frac{(\tau_2 + d)d}{2} \ln 2 + \frac{\tau_2 + d}{2} \ln \left| \left( \boldsymbol{\tau_{12}} - \frac{\boldsymbol{\tau_{11}} \boldsymbol{\tau_{11}}^T}{\tau_2} \right)^{-T} \right| \\
&\quad + \ln \Gamma_d \left( \frac{\tau_2 + d}{2} \right) \Bigg) \\
&= \frac{\partial}{\partial \boldsymbol{\tau_{11}}} \left( -\frac{\tau_2 + d}{2} \ln \left| \boldsymbol{\tau_{12}} - \frac{\boldsymbol{\tau_{11}} \boldsymbol{\tau_{11}}^T}{\tau_2} \right| \right) \\
&= \frac{\partial}{\partial \boldsymbol{\tau_{11}}} \left( -\frac{\tau_2 + d}{2} \ln \left( \left( 1 - \frac{\boldsymbol{\tau_{11}}^T \boldsymbol{\tau_{12}}^{-1} \boldsymbol{\tau_{11}}}{\tau_2} \right) |\boldsymbol{\tau_{12}}| \right) \right) \\
&= \frac{\partial}{\partial \boldsymbol{\tau_{11}}} \left( -\frac{\tau_2 + d}{2} \ln (\tau_2 - \boldsymbol{\tau_{11}}^T \boldsymbol{\tau_{12}}^{-1} \boldsymbol{\tau_{11}}) \right) \\
&= \frac{(\tau_2 + d)(\boldsymbol{\tau_{12}}^{-1} + \boldsymbol{\tau_{12}}^{-T}) \boldsymbol{\tau_{11}}}{2\tau_2 - 2\boldsymbol{\tau_{11}}^T \boldsymbol{\tau_{12}}^{-1} \boldsymbol{\tau_{11}}}
\end{aligned}
$$

$$\tag{B.12}$$

Where we used $|\boldsymbol{B} - \boldsymbol{x}\boldsymbol{x}^T| = (1 - \boldsymbol{x}^T \boldsymbol{B}^{-1} \boldsymbol{x})|\boldsymbol{B}|$ and $\frac{\partial \boldsymbol{x}^T \boldsymbol{B} \boldsymbol{x}}{\partial \boldsymbol{x}} = (\boldsymbol{B} + \boldsymbol{B}^T)\boldsymbol{x}$.

$$\frac{\partial a(\cdots, \boldsymbol{\tau_{12}}, \cdots)}{\partial \boldsymbol{\tau_{12}}} = \frac{\partial}{\partial \boldsymbol{\tau_{12}}} \left( -\frac{d}{2} \ln \tau_2 + \frac{(\tau_2 + d)d}{2} \ln 2 + \frac{\tau_2 + d}{2} \ln \left| \left( \boldsymbol{\tau_{12}} - \frac{\boldsymbol{\tau_{11} \tau_{11}^T}}{\tau_2} \right)^{-T} \right| \right.$$

$$\left. + \ln \Gamma_d \left( \frac{\tau_2 + d}{2} \right) \right)$$

$$= \frac{\partial}{\partial \boldsymbol{\tau_{12}}} \left( -\frac{\tau_2 + d}{2} \ln \left| \boldsymbol{\tau_{12}} - \frac{\boldsymbol{\tau_{11} \tau_{11}^T}}{\tau_2} \right| \right)$$

$$= \frac{\partial}{\partial \boldsymbol{\tau_{12}}} \left( -\frac{\tau_2 + d}{2} \left( \ln \left( \tau_2 - \boldsymbol{\tau_{11}^T \tau_{12}^{-1} \tau_{11}} \right) + \ln |\boldsymbol{\tau_{12}}| \right) \right)$$

$$= -\frac{\tau_2 + d}{2} \left( \frac{\boldsymbol{\tau_{12}^{-T} \tau_{11} \tau_{11}^T \tau_{12}^{-T}}}{\tau_2 - \boldsymbol{\tau_{11}^T \tau_{12}^{-1} \tau_{11}}} + \boldsymbol{\tau_{12}^{-T}} \right)$$

$$\text{(B.13)}$$

Using $\frac{\partial a^T \boldsymbol{X}^{-1} b}{\partial \boldsymbol{X}} = -\boldsymbol{X}^{-T} a b^T \boldsymbol{X}^{-T}$ and $\frac{\partial |\boldsymbol{X}|}{\partial \boldsymbol{X}} = |\boldsymbol{X}| \boldsymbol{X}^{-T}$.

$$\mathbb{E}[-a(\boldsymbol{\eta}^*)] = \frac{\partial a(\cdots, \tau_2)}{\partial \tau_2}$$

$$= \frac{\partial}{\partial \tau_2}\left( -\frac{d}{2}\ln\tau_2 + \frac{(\tau_2+d)d}{2}\ln 2 + \frac{\tau_2+d}{2}\ln\left|\left(\boldsymbol{\tau_{12}} - \frac{\boldsymbol{\tau_{11}\tau_{11}^T}}{\tau_2}\right)^{-T}\right| \right.$$

$$\left. + \ln\Gamma_d\left(\frac{\tau_2+d}{2}\right)\right)$$

$$= -\frac{d}{2\tau_2} + \frac{d}{2}\ln 2 + \frac{\partial}{\partial \tau_2}\left( -\frac{\tau_2+d}{2}\left(\ln\left(\frac{\tau_2 - \boldsymbol{\tau_{11}^T\tau_{12}^{-1}\tau_{11}}}{\tau_2}\right) + \ln|\boldsymbol{\tau_{12}}|\right)\right.$$

$$\left. + \left(\frac{d(d-1)}{4}\ln\pi + \sum_{i=1}^{d}\ln\Gamma\left(\frac{\tau_2+d}{2} + \frac{1-i}{2}\right)\right)\right)$$

$$= -\frac{d}{2\tau_2} + \frac{d}{2}\ln 2 + \frac{1}{2}\sum_{i=1}^{d}\psi\left(\frac{\tau_2+d+1-i}{2}\right) + \frac{\partial}{\partial \tau_2}\left( -\frac{\tau_2}{2}\ln\left(\tau_2 - \boldsymbol{\tau_{11}^T\tau_{12}^{-1}\tau_{11}}\right)\right.$$

$$\left. + \frac{\tau_2}{2}\ln\tau_2 - \frac{\tau_2}{2}\ln|\boldsymbol{\tau_{12}}| - \frac{d}{2}\ln\left(\tau_2 - \boldsymbol{\tau_{11}^T\tau_{12}^{-1}\tau_{11}}\right) + \frac{d}{2}\ln\tau_2\right)$$

$$= \frac{d}{2}\ln 2 + \frac{1}{2}\sum_{i=1}^{d}\psi\left(\frac{\tau_2+d+1-i}{2}\right) - \frac{1}{2}\ln(\tau_2 - \boldsymbol{\tau_{11}^T\tau_{12}^{-1}\tau_{11}})$$

$$- \frac{\tau_2}{2\tau_2 - 2\boldsymbol{\tau_{11}^T\tau_{12}^{-1}\tau_{11}}} - \frac{1}{2}\ln|\boldsymbol{\tau_{12}}| + \frac{1}{2}\ln\tau_2 + \frac{1}{2} - \frac{d}{2\tau_2 - 2\boldsymbol{\tau_{11}^T\tau_{12}^{-1}\tau_{11}}}$$

$$= \frac{1}{2}\left(1 - \frac{d+\tau_2}{\tau_2 - \boldsymbol{\tau_{11}^T\tau_{12}^{-1}\tau_{11}}} + d\ln 2 - \ln|\boldsymbol{\tau_{12}}| + \ln\tau_2 - \ln(\tau_2 - \boldsymbol{\tau_{11}^T\tau_{12}^{-1}\tau_{11}})\right.$$

$$\left. + \sum_{i=1}^{d}\psi\left(\frac{\tau_2+d+1-i}{2}\right)\right)$$

<div align="right">(B.14)</div>

## B.5 Conjugate prior of Gamma likelihood

$$p(\boldsymbol{\eta^*}|\boldsymbol{\lambda}) = h(\boldsymbol{\eta^*})\exp(\boldsymbol{\lambda_1}^T\boldsymbol{\eta^*} + \lambda_2(-a(\boldsymbol{\eta^*})) - a(\boldsymbol{\lambda}))$$

$$= \exp\left(\boldsymbol{\lambda_1}^T\begin{pmatrix}\alpha - 1 \\ -\beta\end{pmatrix} + \lambda_2(-\ln\Gamma(\alpha) + \alpha\ln\Gamma(\beta)) - a(\boldsymbol{\lambda})\right)$$

$$= \exp\left(\lambda_{11}(\alpha - 1) - \lambda_{12}\beta - \lambda_2\ln\Gamma(\alpha) + \lambda_2\alpha\ln\beta - a(\boldsymbol{\lambda})\right)$$

$$= \frac{(\ln\lambda_{11})^{\alpha-1}e^{-\lambda_{12}\beta}}{\Gamma(\alpha)^{\lambda_2}\beta^{\alpha\lambda_2}}e^{-a(\boldsymbol{\lambda})}$$

$$\propto \frac{p^{\alpha-1}e^{-\beta q}}{\Gamma(\alpha)^r\beta^{-\alpha s}}$$

(B.15)

We recognize the corresponding conjugate prior with the following parameters, where $p, q, r, s > 0$ and $f(\alpha, \beta | p, q, r, s) \propto \frac{p^{\alpha-1}e^{-\beta q}}{\Gamma(\alpha)^r\beta^{-\alpha s}}$ if $\alpha, \beta > 0$, 0 otherwise.

$$\begin{cases}\lambda_{11} = e^p \\ \lambda_{12} = q \\ \lambda_2 = r \\ \lambda_2 = -s\end{cases} \qquad\qquad \begin{cases}p = \ln\lambda_{11} \\ q = \lambda_{12} \\ r = \lambda_2 \\ s = -\lambda_2\end{cases}$$

These mappings result in the constraint $r = -s$. The expectation of the sufficient statistic terms cannot be computed for the corresponding posterior since we cannot obtain the analytical form of the normalization factor.

## B.6 Conjugate prior of Beta likelihood

$$p(\boldsymbol{\eta}^*|\boldsymbol{\lambda}) = h(\boldsymbol{\eta}^*) \exp(\boldsymbol{\lambda_1}^T \boldsymbol{\eta}^* + \lambda_2(-a(\boldsymbol{\eta}^*)) - a(\boldsymbol{\lambda}))$$

$$= \exp\left(\boldsymbol{\lambda_1}^T \begin{pmatrix} \alpha - 1 \\ \beta - 1 \end{pmatrix} + \lambda_2(-\ln\Gamma(\alpha) - \ln\Gamma(\beta) + \ln\Gamma(\alpha+\beta)) - a(\boldsymbol{\lambda})\right)$$

$$= \exp\left(\lambda_{11}(\alpha-1) + \lambda_{12}(\beta-1) + \lambda_2 \ln\frac{\Gamma(\alpha+\beta)}{\Gamma(\alpha)\Gamma(\beta)} - a(\boldsymbol{\lambda})\right)$$

$$= \left(\frac{\Gamma(\alpha+\beta)}{\Gamma(\alpha)\Gamma(\beta)}\right)^{\lambda_2} (\ln\lambda_{11})^{\alpha-1}(\ln\lambda_{12})^{\beta-1} e^{-a(\boldsymbol{\lambda})}$$

$$= \left(\frac{\Gamma(\alpha+\beta)}{\Gamma(\alpha)\Gamma(\beta)}\right)^{\lambda_2} (\ln\lambda_{11})^{\alpha}(\ln\lambda_{12})^{\beta} e^{-a(\boldsymbol{\lambda})-\lambda_{11}-\lambda_{12}}$$

$$\propto \left(\frac{\Gamma(\alpha+\beta)}{\Gamma(\alpha)\Gamma(\beta)}\right)^{\lambda_0} x_0^{\alpha} y_0^{\beta}$$

(B.16)

We recognize the corresponding conjugate prior $\pi(\alpha,\beta|\lambda_0,x_0,y_0)$ with based on the following parameter mappings.

$$\begin{cases} \lambda_{11} = e^{x_0} \\ \lambda_{12} = e^{y_0} \\ \lambda_2 = \lambda_0 \end{cases} \qquad \begin{cases} \lambda_0 = \lambda_2 \\ x_0 = \ln\lambda_{11} \\ y_0 = \ln\lambda_{12} \end{cases}$$

As previously, the expectations for the posterior cannot be computed due to the missing analytical form for the normalization factor.